\documentclass[10pt,journal,compsoc]{IEEEtran}

\ifCLASSOPTIONcompsoc
  \usepackage[nocompress]{cite}
\else
  \usepackage{cite}
\fi

\usepackage{ragged2e} 
\usepackage[permil]{overpic}
\usepackage{arydshln}  
\usepackage{amsmath}
\usepackage{amssymb}
\usepackage{amsfonts}
\usepackage{booktabs} 
\usepackage{multirow}
\usepackage{algorithm}
\usepackage{algorithmic}
\usepackage{xcolor}
\usepackage{subcaption}
\usepackage{mathtools}
\usepackage{amsthm}

\theoremstyle{plain}
\newtheorem{definition}{Definition}
\newtheorem{theorem}{Theorem}
\newtheorem{lemma}{Lemma}

\newtheorem{corollary}{Corollary}
\newtheorem{proposition}{Proposition}
\usepackage{hyperref}   
\usepackage{cleveref}
\crefname{figure}{Fig.}{Figs.}
\crefname{table}{Table}{Tables}
\crefname{equation}{Eq.}{Eqs.}

\newcommand\ie{i.e.,~}
\newcommand\et{\textit{et al.~}}

\newcommand{\dam}[0]{\textsc{Dam}}
\newcommand{\sam}[0]{\textsc{Sam}}
\newcommand{\mam}[0]{\textsc{Mam}}
\newcommand{\cc}[0]{\textsc{Cc}}
\newcommand{\cn}[0]{\textsc{Cn}}
\newcommand{\mamcn}[0]{\textsc{MamCn}}
\newcommand{\mamcc}[0]{\textsc{MamCc}}
\newcommand{\samcn}[0]{\textsc{SamCn}}
\newcommand{\samcc}[0]{\textsc{SamCc}}
\newcommand{\damcn}[0]{\textsc{DamCn}}
\newcommand{\damcc}[0]{\textsc{DamCc}}
\newcommand{\mamfl}[0]{\textsc{MamFl}}
\newcommand{\mamifl}[0]{\textsc{MamIfl}}

\newcommand{\miplma}[0]{\textsc{MiplMa}}
\newcommand{\elimipl}[0]{\textsc{EliMipl}}
\newcommand{\demipl}[0]{\textsc{DeMipl}}
\newcommand{\miplgp}[0]{\textsc{MiplGp}}
\newcommand{\fastmipl}[0]{\textsc{FastMipl}}
\newcommand{\promipl}[0]{\textsc{ProMipl}}
\newcommand{\proden}[0]{\textsc{Proden}}

\newcommand{\lws}[0]{\textsc{Lws}}

\newcommand{\pop}[0]{\textsc{Pop}}

\begin{document}

\title{Calibratable Disambiguation Loss for Multi-Instance Partial-Label Learning}

\author{
Wei Tang, Yin-Fang Yang, Weijia~Zhang, and~Min-Ling~Zhang,~\IEEEmembership{Senior~Member,~IEEE}
\IEEEcompsocitemizethanks{\IEEEcompsocthanksitem Wei Tang, Yin-Fang Yang, and Min-Ling Zhang are with the School of Computer Science and Engineering, Southeast University, Nanjing 210096, China, and the Key Laboratory of Computer Network and Information Integration (Southeast University), MoE, China.\protect\\
E-mail: \{tangw, yangyf, zhangml\}@seu.edu.cn.
\IEEEcompsocthanksitem Weijia~Zhang is with the School of Computer and Information Sciences, The University of Newcastle, Callaghan, NSW {\rm 2308}, Australia. \protect\\
E-mail: weijia.zhang@newcastle.edu.au.
\IEEEcompsocthanksitem Corresponding author: Min-Ling~Zhang.
}
}

\IEEEtitleabstractindextext{%
\begin{abstract}
\justifying
Multi-instance partial-label learning (MIPL) is a weakly supervised framework that extends the principles of multi-instance learning (MIL) and partial-label learning (PLL) to address the challenges of inexact supervision in both instance and label spaces. However, existing MIPL approaches often suffer from poor calibration, undermining classifier reliability. In this work, we propose a plug-and-play calibratable disambiguation loss (CDL) for classification and calibration, which modulates a disambiguation objective by a top-vs-competitor prediction margin. The competitor is instantiated either as the second strongest candidate label or as the strongest non-candidate label, yielding two variants that respectively emphasize candidate-level separation and candidate-vs-non-candidate suppression. Theoretically, we analyze CDL as a margin-modulated momentum-based disambiguation loss (MDL) objective, derive a lower-bound and a pseudo-label confidence-alignment bound for calibration, and show through gradient and momentum analyses how margin shaping affects weight updates. Experimental results on benchmark and real-world MIPL datasets, together with representative PLL adaptation, confirm that our CDL significantly improves both classification accuracy and expected calibration error.
\end{abstract}

}

\maketitle

\IEEEdisplaynontitleabstractindextext

\IEEEpeerreviewmaketitle

\IEEEraisesectionheading{\section{Introduction}\label{sec:intro}}
\IEEEPARstart{W}{eakly} supervised learning enables the construction of predictive models with limited supervision. According to \cite{zhou2018brief}, weak supervision can be classified into three types: inexact, inaccurate, and incomplete. Inexact supervision arises from a coarse alignment between instances and labels, a common and challenging issue in real-world applications. Two prominent frameworks that address inexact supervision from different perspectives are \emph{multi-instance learning (MIL)} \cite{wang2018revisiting,Weijia22,zhang2022dtfd} and \emph{partial-label learning (PLL)} \cite{ZhangF0L0QS22,HeFLLY22,GongYBL23}. MIL addresses inexactness in the instance space, where the positive instances within a bag are unidentified \cite{hajj24a,zhang2024data}, while PLL focuses on inexactness in the label space, where the true label remains hidden within a candidate label set \cite{LiJLWO23,WuWZ24,wang2025causalpll}.

\begin{figure}[!t]
\setlength{\abovecaptionskip}{0.cm} 
    \centering
	\begin{overpic}[width=8cm]{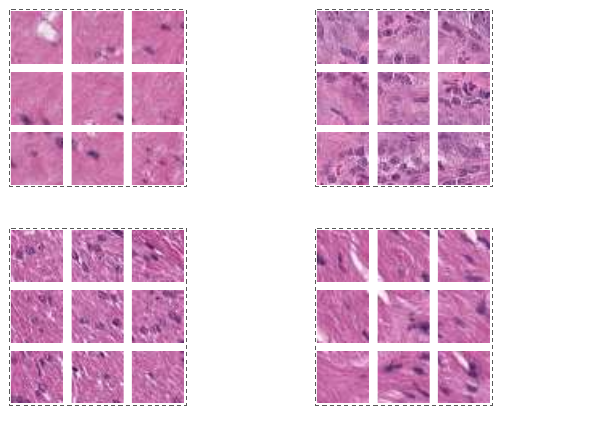}
        \put(145, 380){\small (a)}
        \put(145, 10){\small (b)}
        \put(660, 380){\small (c)}
	    \put(660, 10){\small (d)}
        \put(335, 600){\textcolor{red}{MUS}}
        \put(335, 500){STR}
        \put(850, 630){LYM}
        \put(850, 550){STR}
        \put(850, 470){\textcolor{red}{TUM}}
        \put(335, 260){\textcolor{red}{MUS}}
        \put(335, 180){STR}
        \put(335, 100){TUM}
        \put(850, 260){MUS}
        \put(850, 180){\textcolor{red}{STR}}
        \put(850, 100){TUM}  
    \end{overpic}
    \caption{Pathology image classification with crowd-sourced candidate label sets is a MIPL scenario \cite{tang2023demipl}, where true labels are highlighted in red and false-positive labels in black. The labels include LYM (lymphocytes), MUS (smooth muscle), STR (cancer-associated stroma), and TUM (colorectal adenocarcinoma epithelium).}
 \label{fig:mipl_illu}
\end{figure}

Recent advancements have introduced \emph{multi-instance partial-label learning (MIPL)} to jointly model these two sources of inexactness \cite{tang2023miplgp}. In MIPL, each training example is a multi-instance bag associated with a candidate label set. During training, both the positive instances in the instance space and the true label in the label space remain unknown. Fig.~\ref{fig:mipl_illu} shows a pathology image classification scenario. Whole-slide or high-resolution pathology images are often divided into patches for computational feasibility \cite{campanella2019}, while candidate labels can be collected from crowd-sourced annotators to reduce expert annotation cost \cite{GroteSFWF19,tang2023demipl}. Such data naturally contain inexactness in both instance and label spaces, making MIPL a suitable formulation.

Existing MIPL approaches can be categorized into two paradigms: the instance-space paradigm and the embedded-space paradigm. The instance-space paradigm generates a predicted label for a multi-instance bag by aggregating the prediction probabilities of its constituent instances \cite{tang2023miplgp}. In contrast, the embedded-space paradigm classifies multi-instance bags directly by aggregating them into a single feature vector \cite{tang2023demipl}. The latter paradigm often exhibits superior classification performance owing to its ability to capture global feature representations. Label disambiguation is central to MIPL, which involves identifying the true label from the candidate set and significantly influences classification performance \cite{tang2023demipl, LvXF0GS20}. However, existing MIPL objectives are primarily designed for disambiguation. They do not explicitly ensure that the predicted confidence reflects the empirical probability of being correct, which limits their reliability in decision-sensitive applications. \looseness=-1

\begin{figure}[!t]
\setlength{\abovecaptionskip}{0.cm} 
    \centering
	\begin{overpic}[width=8.7cm]{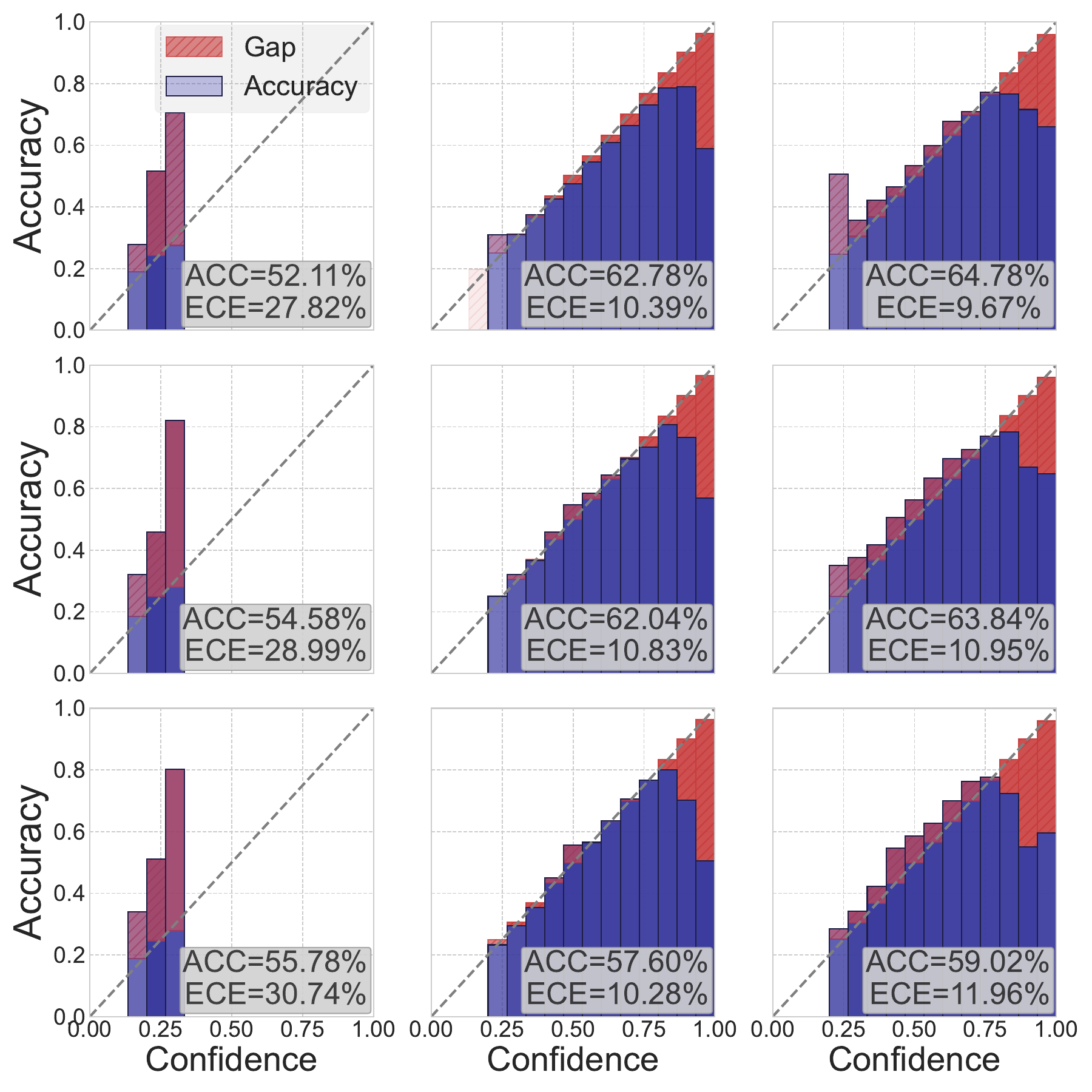}
        \put(90, 940){\small (a)}
    	\put(405, 940){\small (d)}
    	\put(715, 940){\small (g)}
    	\put(90, 625){\small (b)}
    	\put(405, 625){\small (e)}
    	\put(715, 625){\small (h)}
        \put(90, 315){\small (c)}
    	\put(405, 315){\small (f)}
    	\put(715, 315){\small (i)}
    \end{overpic}
    \caption{
    Reliability diagrams of (a) {\demipl} \cite{tang2023demipl}, (b) {\elimipl} \cite{tang2024elimipl}, (c) {\miplma} \cite{tang2024miplma}, (d) {\damcc}, (e) {\samcc}, (f) {\mamcc}, (g) {\damcn}, (h) {\samcn}, and (i) {\mamcn} on the C-KMeans test set. The diagrams display mean accuracy (ACC) and expected calibration error (ECE) from ten runs, with (d)-(i) representing our methods. The bar color intensity reflects the number of samples assigned to the corresponding confidence intervals.
}
 \label{fig:ece_intro}
\end{figure}

This limitation is evident in Fig. \ref{fig:ece_intro} (a), (b), and (c), the predicted confidences of {\demipl} \cite{tang2023demipl}, {\elimipl} \cite{tang2024elimipl}, and {\miplma} \cite{tang2024miplma} tend to cluster around $0.25$, leading to poor expected calibration error (ECE). In pathology image classification, such unreliable confidence estimates are undesirable because confidence scores may guide triage, human review, diagnosis, and treatment planning. Moreover, calibration losses designed for fully supervised data assume known true labels. Naively adapting focal-type losses to MIPL can interfere with pseudo-label evolution and lead to under-confident or over-confident predictions. Therefore, MIPL requires a training objective that improves calibration without weakening label disambiguation.

To this end, we propose a plug-and-play \emph{Calibratable Disambiguation Loss} (CDL). CDL modulates a momentum-based disambiguation objective by a top-vs-competitor margin, where the competitor is instantiated either as the second-highest candidate-label probability or as the highest non-candidate-label probability. The two variants, namely CDL-CC and CDL-CN, respectively encourage candidate-level separation and candidate-vs-non-candidate suppression. Low-margin bags retain strong disambiguation signals, whereas well-separated bags receive less incentive for excessive probability sharpening. CDL can be seamlessly integrated into existing embedded-space MIPL frameworks. As shown in Fig.~\ref{fig:ece_intro} (d)--(i), the resulting variants improve both classification accuracy and ECE.

Our key contributions are as follows: 1) To the best of our knowledge,  this work is the first to identify model calibration issues in MIPL and show that existing disambiguation-centered methods can produce poorly calibrated confidence estimates. 2) We propose CDL, a plug-and-play top-vs-competitor margin-modulated disambiguation loss with two complementary instantiations. 3) We theoretically analyze CDL as a margin-modulated MDL objective, establish its loss-level and calibration-related properties, and explain its margin-shaping effect through gradient and momentum analyses. 4) Extensive experiments on benchmark and real-world MIPL datasets, together with representative PLL adaptation, demonstrate that CDL consistently improves both classification accuracy and calibration performance.

\section{Related Work}\label{sec:related_work}
\subsection{Multi-Instance Learning}
Multi-instance learning (MIL), initially developed for drug activity prediction \cite{dietterich1997}, has since been applied across diverse fields, including text classification \cite{ZhouSL09, amores2013multiple, carbonneau2018multiple}, object detection \cite{YuanWFLXJY21}, and video anomaly detection \cite{LvYSL0Z23}. Recent advancements in MIL often incorporate attention mechanisms to synthesize features from multiple instances within a bag into a cohesive representation for classification. For example, Ilse \et \cite{IlseTW18} introduced both plain and gated attention mechanisms, which significantly improved the performance of binary MIL tasks. An extension of this paradigm, the loss-based attention mechanism \cite{Shi20} effectively addresses multi-class classification challenges. The success of attention-based MIL approaches has led to their widespread adoption in applications such as pathology image classification \cite{cui2023bayesmil, FourkiotiVB24} and time series classification \cite{early2024inherently}. These approaches utilize a weighted aggregation of attention scores and instances to generate a comprehensive bag-level feature. 

Despite the significant advancements, existing MIL approaches remain ineffective for MIPL scenarios because they cannot directly handle ambiguous candidate label sets \cite{tang2023miplgp}.

\subsection{Partial-Label Learning}
Partial-label learning (PLL) has diverse applications across numerous real-world domains, such as face naming \cite{cour2011learning, lyu2019gm, YaoDC0W020}, object classification \cite{liu2012conditional,he2025partialclip}, bioinformatics \cite{briggs2012rank}, and facial age estimation \cite{WangZ20, wangdb2022}. Recent developments have led to the emergence of several deep learning-based PLL methods. For example, Lv \et \cite{LvXF0GS20} utilized linear classifiers and multi-layer perceptrons to generate feature representations from instances, employing progressive disambiguation techniques to identify the true labels. Building on this foundation, Feng \et \cite{FengL0X0G0S20} explored the process of generating PLL data and introduced two theoretically robust algorithms for PLL. Similarly, Wen \et \cite{WenCHL0L21} proposed a weighted loss function for disambiguation that provides a versatile approach applicable to various algorithms. Furthermore, Xu \et \cite{XuLLQG23} applied progressive purification of candidate labels to train classifiers within the instance-dependent PLL framework. Recently, Wang \et \cite{wang2025realistic} established the first PLL benchmark and introduced theoretically justified model selection criteria for PLL.

However, the inherent limitations of these PLL approaches in handling inexact supervision in the instance space impede their effectiveness in MIPL scenarios \cite{tang2023miplgp}.

\subsection{Multi-Instance Partial-Label Learning}
MIPL extends MIL and PLL to handle dual inexact supervision, where both instances within a bag and labels in the candidate set are ambiguous. Tang \et \cite{tang2023miplgp} first formalized MIPL and proposed {\miplgp}, an instance-space method that assigns pseudo-negative classes, propagates bag-level candidate labels to instances, and performs Dirichlet-based disambiguation with Gaussian process regression. By transforming discrete candidate labels into continuous ones, {\miplgp} enables uncertainty-aware label disambiguation. However, its bag-level prediction relies on the maximum instance-level response, limiting its ability to learn global bag representations.
To address this limitation, {\demipl} \cite{tang2023demipl} adopts an embedded-space paradigm, where bags are encoded by attention-based aggregation and true labels are identified through a momentum strategy. Although {\demipl} captures global bag information, it insufficiently constrains the full label space and may assign high probabilities to non-candidate labels. {\elimipl} \cite{tang2024elimipl} and {\miplma} \cite{tang2024miplma} mitigate this issue by explicitly suppressing non-candidate labels, yielding improved performance over earlier MIPL methods. However, these approaches remain primarily discriminative and do not characterize the data-generation mechanism. {\promipl} \cite{yang2024promipl} addresses this issue with a probabilistic generative model that infers latent ground-truth labels from the assumed generation process, but its reliance on a specific bag prior may restrict cross-dataset generalization. More recently, {\fastmipl} \cite{yang2025fastmipl} introduced a mixed-effects formulation to model instance-bag dependencies with improved efficiency, while its generalized linear structure may limit expressivity on complex data. Wang \et \cite{wang2024learning} further explored MIPL through latent structural learning and neuro-symbolic integration, extending its application scope but relying on relatively strong assumptions.

Despite these advances, existing MIPL studies mainly emphasize label disambiguation, whereas model calibration remains insufficiently explored.

\subsection{Model Calibration}
Model calibration is crucial for ensuring that predicted probabilities accurately reflect true likelihoods of the respective classes \cite{GuoPSW17}. Calibration methods can be broadly categorized into training-time and post-hoc approaches. Training-time calibration technologies incorporate regularization techniques during training to address overconfidence in deep neural networks. For example, label smoothing is a prominent technique that reduces overfitting and enhances calibration by mitigating excessively confident predictions \cite{MullerKH19, WangFZ21}. Implicit regularization methods, such as mixup training and focal loss, originally designed to improve generalization, have also proven effective for calibration \cite{ThulasidasanCBB19, ZhangDK022, MukhotiKSGTD20, WangFZ21, TaoD023}.
The post-hoc calibration methods adjust model outputs after training to better align predicted probabilities with true likelihoods. Platt scaling refines binary classifier outputs using learned parameters \cite{Platt1999, TomaniGECB21}, whereas temperature scaling, an extension of Platt scaling, rescales logits in multi-class tasks using a learned temperature parameter \cite{GuoPSW17}.

While the above calibration techniques provide substantial benefits for supervised learning tasks, using them in MIPL is challenging due to the absence of true labels.

\section{Preliminaries}
\subsection{Notations}
We formalize a MIPL training dataset as $\mathcal{D} = \{(\boldsymbol{X}_i, \mathcal{S}_i) \mid 1 \le i \le m\}$, where $\mathcal{D}$ consists of $m$ multi-instance bags, each associated with a corresponding candidate label set. The instance space is represented by $\mathcal{X} = \mathbb{R}^d$, and the label space is denoted as $\mathcal{Y} = \{1, 2, \ldots, k\}$, which includes $k$ distinct class labels. Specifically, the $i$-th multi-instance bag $\boldsymbol{X}_i = \{\boldsymbol{x}_{i,1}, \boldsymbol{x}_{i,2}, \ldots, \boldsymbol{x}_{i,n_i}\}$ contains $n_i$ instances within a $d$-dimensional space. The candidate label set $\mathcal{S}_i$ and the non-candidate label set $\bar{\mathcal{S}}_i$ are subsets of $\mathcal{Y}$, satisfying the constraints $\mathcal{S}_i \cup \bar{\mathcal{S}}_i = \mathcal{Y}$ and $\mathcal{S}_i \cap \bar{\mathcal{S}}_i = \varnothing$.

Notably, each bag contains at least one instance associated with the true label, referred to as a positive instance. In contrast, negative instances may correspond to background or irrelevant content, but they should not be associated with any false-positive labels in the candidate label set. For example, in pathology image classification \cite{tang2023demipl}, positive instances represent specific cell types, such as lymphocytes or colorectal adenocarcinoma epithelium, whereas negative instances include background regions or non-cellular areas.

\begin{figure}[!t]
    \centering
    \begin{overpic}[width=9cm]{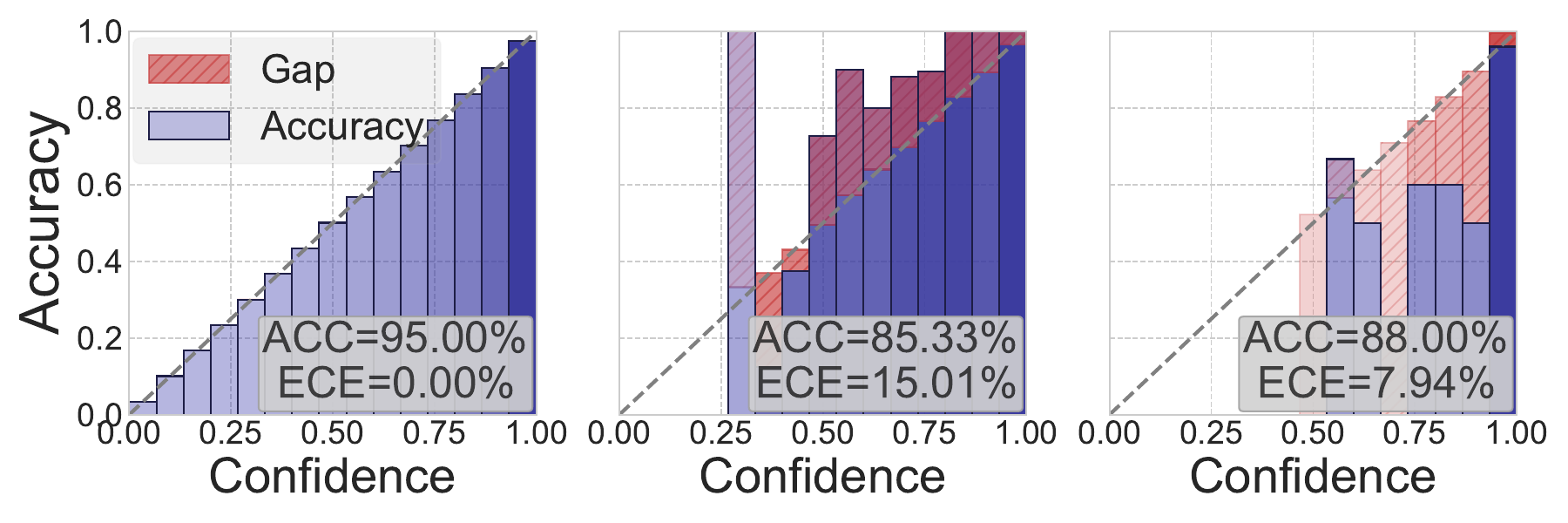}
        \put(30, -20){\small  (a) Perfect calibration}
    	\put(480, -20){\small  (b) FL}
        \put(790, -20){\small  (c) IFL}
    \end{overpic}
    \caption{Reliability diagrams of {\sam} \cite{tang2024elimipl} with FL or IFL on the FMNIST-{\scriptsize{MIPL}} dataset with one false positive label ($r=1$).}
 \label{fig:ece_elimipl_fl}
\end{figure}

\begin{figure*}[!t]
    \setlength{\abovecaptionskip}{0.1cm} 
    \centering
    \begin{overpic}[width=180mm]{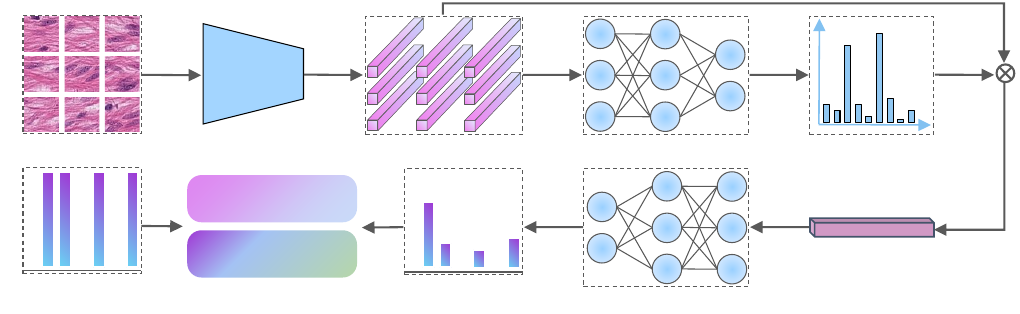}   
        \put(0, 160) {\small multi-instance bag $\boldsymbol{X}_i$}
        \put(230, 230) {\small $\Psi(\cdot)$}
        \put(185, 160) {\small feature extractor}
        \put(335, 160) {\small instance-level features $\boldsymbol{H}_i$}
        \put(562, 160) {\small aggregation mechanism}
        \put(782, 160) {\small attention scores $\boldsymbol{A}_{i}$}
        \put(782, 58) {\small bag-level feature $\boldsymbol{z}_i$}
        \put(0, 5) {\small candidate label set $\mathcal{S}_i$}
        \put(205, 119) {\small disambiguation}
        \put(192, 96) {\small (\textit{existing approaches})}
        \put(193, 63) {\small \textbf{\textcolor{black}{disambiguation \&}}}
        \put(198, 40) {\small \textbf{\textcolor{black}{calibration}} \textcolor{black}{(\textit{ours})}}
        \put(28, 22) {\small $1$}
        \put(43, 22) {\small $2$}
        \put(60, 22) {\small $3$}
        \put(77, 22) {\small $4$}
        \put(94, 22) {\small $5$}
        \put(110, 22) {\small $6$}
        \put(126, 22) {\small $7$}
        \put(404, 22) {\small $1$}
        \put(417, 22) {\small $2$}
        \put(433, 22) {\small $3$}
        \put(450, 22) {\small $4$}
        \put(467, 22) {\small $5$}
        \put(484, 22) {\small $6$}
        \put(499, 22) {\small $7$}
        \put(394, 5) {probabilities $\hat{\boldsymbol{p}}_i$}
        \put(620, 8) {classifier}
    \end{overpic}
    \caption{Framework of MIPL approaches within the embedded-space paradigm.}
    \label{fig:framework}
\end{figure*}

\subsection{Calibration}
A well-calibrated model ensures that the confidence scores of the predictions reflect the true probabilities of correctness \cite{GuoPSW17, WangLZHZ23, TaoD023, wang24cm}, which can be formally expressed as:
\begin{equation}
    \mathbb{P}(\hat{y} = y \mid \hat{p} = p) = p, \quad \forall p \in [0,1],
\end{equation}
where $\hat{y}$ and $\hat{p}$ denote the predicted label and confidence score, respectively, while $y$ represents the true label. The confidence score $p$ is the model's estimated probability that the prediction $\hat{y}$ is correct.

The expected calibration error (ECE) quantifies calibration quality from finite samples by partitioning the predicted confidences into $R$ bins $\{B_r\}_{r=1}^{R}$ and computing:
\begin{equation}
\label{eq:ece}
    \text{ECE} = \sum_{r=1}^R \frac{|B_r|}{m} \left| A_r - P_r \right|,
\end{equation}
where $A_r = \frac{1}{|B_r|} \sum_{i \in B_r} \mathbb{I}(\hat{y}_i = y_i)$ and $P_r = \frac{1}{|B_r|} \sum_{i \in B_r} \hat{p}_i$ represent the accuracy and average confidence in bin $r$, respectively. This metric assesses how well-predicted confidences align with actual correctness, typically employing $R=15$ bins. A model is well-calibrated if its predicted confidences closely reflect empirical accuracies. Fig. \ref{fig:ece_elimipl_fl} (a) illustrates perfect calibration using generated data, where, in each bin, predicted confidences precisely match accuracies, resulting in perfect calibration, \ie $\text{ECE} = 0$.

Focal loss (FL) \cite{LinGGHD20} was introduced to mitigate class imbalance in object detection by reducing the loss weight for predicted samples with high confidence. Mukhoti \textit{et al.} \cite{MukhotiKSGTD20} demonstrated that replacing cross-entropy loss with FL enhances model calibration. Moreover, Wang \textit{et al.} \cite{WangFZ21} proposed inverse focal loss (IFL), which improves calibration performance further. The formulations of FL and IFL are:
\begin{equation}
\label{eq:fl}
    \mathcal{L}_{\text{FL}} = -(1 - \hat{p}_{i,y_i})^\gamma \log \hat{p}_{i,y_i},
\end{equation}
\begin{equation}
\label{eq:ifl}
    \mathcal{L}_{\text{IFL}} = -(1 + \hat{p}_{i,y_i})^\gamma \log \hat{p}_{i,y_i},
\end{equation}
where $y_i$ denotes the true label of the $i$-th sample.

In standard supervised learning, focal losses significantly improve calibration. However, applying FL and IFL in MIPL is challenging due to the absence of true labels, as illustrated in Fig. \ref{fig:ece_elimipl_fl} (b) and (c).

\subsection{The Pitfalls of Naive Focal Losses in MIPL}
\label{sec:pilot}
One naive approach for extending the focal losses to MIPL is to select the candidate label with the highest predicted probability as a surrogate of the true label. However, this naive implementation does not create well-calibrated models and also significantly reduces the classification accuracy in MIPL.
To compensate for this, we propose to employ all candidate labels with different weights. Specifically, we utilize the Scaled additive Attention Mechanism ({\sam}) in {\elimipl} \cite{tang2024elimipl} to derive the holistic features of multi-instance bags, coupled with FL or IFL as the loss function. Since FL and IFL were originally designed for fully supervised settings, we refine FL and IFL for MIPL as follows: 
\begin{equation}
\label{eq:fl_mipl}
    \mathcal{L}_{\text{FL}}^{\text{MIPL}} = - \sum_{c \in \mathcal{S}_i}  w_{i, c}^{(t)} (1 - \hat{p}_{i, c}^{(t)})^\gamma  \log (\hat{p}_{i, c}^{(t)}),
\end{equation}
\vspace{-2mm}
\begin{equation}
\label{eq:ifl_mipl}
    \mathcal{L}_{\text{IFL}}^{\text{MIPL}} = - \sum_{c \in \mathcal{S}_i}  w_{i, c}^{(t)} (1 + \hat{p}_{i, c}^{(t)})^\gamma  \log (\hat{p}_{i, c}^{(t)}),
\end{equation}
where $\mathcal{S}_i$ is the candidate label set of the $i$-th multi-instance bag and $\hat{p}_{i,c}$ is the predicted probabilities of the $c$-th label. $w_{i, c}^{(t)}$ denotes the weights of the $c$-th class at the $t$-th epoch. 

Experimental results in Fig. \ref{fig:ece_elimipl_fl} reveal that the average classification accuracy using FL and IFL for calibration is lower compared to that of {\elimipl} ($90.27\%$). Specifically, FL suffers from under-confidence, where classification accuracy significantly exceeds predicted confidence, while IFL exhibits over-confidence, where predicted confidence significantly surpasses classification accuracy. The results indicate that while FL and IFL can be adapted for MIPL, their effectiveness is constrained, resulting in reduced classification accuracy and notable under-confidence and over-confidence issues. Thus, despite their strong performance in standard supervised learning, applying FL and IFL in MIPL requires further refinement and optimization.

\section{The Proposed Approach}
\label{sec:methodology}
The framework of the MIPL approaches based on the embedded-space paradigm is detailed in Fig. \ref{fig:framework}. First, a feature extractor $\Psi(\cdot)$ generates instance-level feature representations $\boldsymbol{H}_i$ from the $i$-th multi-instance bag $\boldsymbol{X}_i$. Second, an aggregation mechanism combines $\boldsymbol{H}_i$ into a unified feature vector $\boldsymbol{z}_i$. Subsequently, a classifier estimates the probability distribution $\hat{\boldsymbol{p}}_i$ for each multi-instance bag. Unlike the existing MIPL methods that focus solely on disambiguation, our approach integrates an additional emphasis on model calibration, thereby providing a more comprehensive solution.

\subsection{Aggregation Mechanisms in MIPL}
\label{subsec:aggregation}
In MIPL, aggregation mechanisms play a crucial role in integrating information from multi-instance bags.
Given a multi-instance bag $\boldsymbol{X}_i = \{\boldsymbol{x}_{i,1}, \boldsymbol{x}_{i,2}, \ldots, \boldsymbol{x}_{i,n_i}\}$, its instance-level feature representations $\boldsymbol{H}_i = \Psi(\boldsymbol{X}_i) = \{\boldsymbol{h}_{i,1}, \boldsymbol{h}_{i,2}, \ldots, \boldsymbol{h}_{i,n_i}\}$ are learned via a feature extractor $\Psi(\cdot)$. The significance of the $j$-th instance within the $i$-th multi-instance bag is modeled as follows:
\begin{equation}
\label{eq:contribution}
    \xi(h_{i,j}) = \boldsymbol{W}^\top (\text{tanh}(\boldsymbol{W}_t^\top \boldsymbol{h}_{i,j}) \odot \text{sigm}(\boldsymbol{W}_s^\top \boldsymbol{h}_{i,j})),
\end{equation}
where $\boldsymbol{W}^\top$, $\boldsymbol{W}_t^\top$, and $\boldsymbol{W}_s^\top$ are learnable weight matrices, including bias terms. Here, $\text{tanh}(\cdot)$ and $\text{sigm}(\cdot)$ denote the hyperbolic tangent and sigmoid functions, respectively, and $\odot$ represents element-wise multiplication.

The central aspect of aggregation mechanisms is the computation of attention scores. Existing aggregation mechanisms in MIPL can be categorized into three types: Disambiguation Attention Mechanism ({\dam}) in {\demipl} \cite{tang2023demipl}, Scaled additive Attention Mechanism ({\sam}) in {\elimipl} \cite{tang2024elimipl}, and Margin-aware Attention Mechanism ({\mam}) in {\miplma} \cite{tang2024miplma} and {\promipl} \cite{yang2024promipl}. While all three mechanisms employ attention mechanisms for aggregation, they differ in how attention scores are computed.

\begin{enumerate}
    \item {\dam} determines the attention scores by applying a nonlinear transformation to $\xi(h_{i,j})$:
    \begin{equation}
    \label{eq:demipl_att}
        a_{i,j} = \frac{1 / \{ 1 + \exp \{- \xi(h_{i,j})\} \}} {\sum_{j^\prime = 1}^{n_i} 1 / \{ 1 + \exp \{- \xi(h_{i,j^\prime})\} \}}.
    \end{equation}
    
    \item {\sam} refines attention scores by exponentiating $\xi(h_{i,j})$ and introducing a scaling factor $l$:
    \begin{equation}
    \label{eq:elimipl_att}
        a_{i,j} = \frac{\exp \{ \xi(h_{i,j}) / \sqrt{l} \}} {\sum_{j^\prime = 1}^{n_i} \exp \{\xi(h_{i,j^\prime}) / \sqrt{l} \}}.
    \end{equation}
    
    \item {\mam} dynamically adjusts attention scores during training by varying the temperature parameter $\tau^{(t)}$:
    \begin{equation}
    \label{eq:miplma_att}
        \breve{a}_{i,j} = \frac{\exp \{ \xi(h_{i,j}) / \tau^{(t)} \}} {\sum_{j^\prime = 1}^{n_i} \exp \{\xi(h_{i,j^\prime}) / \tau^{(t)} \}}.
    \end{equation}
    At the first epoch, the temperature parameter is initialized as a predefined constant and is annealed at the $t$-th epoch according to:
    \begin{equation}
    \label{eq:tau}
        \tau^{(t)} = \max \{ \tau_m, \tau^{(t-1)} * 0.95\},
    \end{equation}
    where $\tau_m$ denotes the minimum temperature parameter, and $\tau^{(0)}$ is the predefined initial value. To stabilize training, a normalization operation is applied to the attention scores $\breve{a}_{i,j}$:
    \begin{equation}
    \label{eq:normalization}
        a_{i,j} = \frac{\breve{a}_{i,j} - \bar{a}_{i}} {\sqrt{\sum_{j^\prime=1}^{n_i} (\breve{a}_{i,j^\prime} - \bar{a}_{i})^2 / (n_i - 1) } },
    \end{equation}
    where $\bar{a}_{i} = \frac{1}{n_i} \sum_{j^\prime=1}^{n_i} \breve{a}_{i,j^\prime}$ represents the mean attention score for the $i$-th multi-instance bag.
\end{enumerate}

After computing attention scores using one of the three mechanisms, the bag-level feature $\boldsymbol{z}_i$ is obtained as a weighted sum of the scores and instance-level features:
\begin{equation}
\label{eq:aggregation}
	\boldsymbol{z}_i =  \sum \limits_{j=1}^{n_i} a_{i,j} \boldsymbol{h}_{i,j}.
\end{equation}
In summary, all three MIPL attention mechanisms have the same workflow but differ in their computation of the attention scores.

\subsection{Calibratable Disambiguation Loss}
\label{subsec:cdl}
As illustrated in Fig. \ref{fig:ece_elimipl_fl}, applying  FL and IFL to MIPL can lead to under-confidence and over-confidence in the predicted probabilities. To tackle these problems, we introduce a novel calibratable disambiguation loss (CDL) specifically designed to balance prediction probabilities.
\begin{definition}[Calibratable Disambiguation Loss]
\label{def:cdl}
For the $i$-th training bag, let $\mathcal{S}_i$ denote its candidate label set. At epoch $t$, we define the current top-ranked candidate label and its corresponding predicted probability as $u_i^{(t)} = \arg\max_{c\in S_i} \hat p_{i,c}^{(t)}$ and $q_i^{(t)} = \hat p_{i,u_i^{(t)}}^{(t)} = \max_{c\in S_i} \hat p_{i,c}^{(t)}$.

The Calibratable Disambiguation Loss (CDL) for multi-instance partial-label learning is defined as:
\begin{equation}
\label{eq:cdl}
    \mathcal{L}_{\text{CDL}} = - \sum_{c \in \mathcal{S}_i} w_{i, c}^{(t)} (1 - q_i^{(t)} + \Phi(\hat{\boldsymbol{p}}_{i}^{(t)}))^\gamma \log (\hat{p}_{i, c}^{(t)}),
\end{equation}
where $w_{i, c}^{(t)}$ denotes the pseudo-label weight for the $c$-th class at the $t$-th epoch. $\gamma$ denotes the exponential factor, and $\Phi(\hat{\boldsymbol{p}}_{i}^{(t)})$ represents the predictive probability of a competitor.
\end{definition}

The modulation $(1-q_i^{(t)}+\Phi(\hat{\mathbf p}_i))^\gamma$ can be written as $(1-\beta_i)^\gamma$, where $\beta_i = q_i^{(t)} - \Phi(\hat{\boldsymbol{p}}_{i}^{(t)})$ is a top-vs-competitor margin and is computed from the current predictive distribution.
Therefore, $\Phi(\hat{\mathbf p}^{(t)}_i)$ makes CDL margin-aware: it couples the loss weight to the separability between the top candidate and its competitor, rather than to an absolute confidence value alone. This encourages the model to increase $q_i^{(t)}$ while suppressing $\Phi(\hat{\boldsymbol{p}}_{i}^{(t)})$, thereby reshaping the predictive distribution toward better-separated and more reliable probabilities.  \looseness=-1

For non-candidate labels $\bar{c} \in \bar{\mathcal{S}}_i$, the pseudo-label weight $w_{i, \bar{c}}^{(t)}$ is maintained at zero for all epochs $t \in \{1, 2, \dots, T\}$, where $T$ denotes the total number of training epochs. Conversely, for candidate labels $c \in \mathcal{S}_i$, the weight is initialized as $w_{i,c}^{(1)} = \frac{1}{\left|\mathcal{S}_i\right|}$ with $|\cdot|$ representing the set cardinality.
For epochs $t \in \{2,3,\dots, T\}$, the weight is updated as follows: 
\begin{equation}
\label{eq:update_w}
	w_{i,c}^{(t)} = \alpha^{(t)} w_{i,c}^{(t-1)}  + (1-\alpha^{(t)}) \frac{\hat{p}_{i, c}^{(t)}}{\sum_{c \in \mathcal{S}_i} \hat{p}_{i, c}^{(t)}},
\end{equation}
where the parameter $\alpha^{(t)} = \frac{T - t}{T}$ regulates the update rate.

CDL can be instantiated in various ways by adjusting the transformation function $\Phi(\hat{\boldsymbol{p}}_{i}^{(t)})$. Specifically, we explore two distinct implementations: one where $\Phi(\hat{\boldsymbol{p}}_{i}^{(t)})$ is the second highest predicted probability among the candidate labels, \ie $\Phi(\hat{\boldsymbol{p}}_{i}^{(t)}) = \max_{c^\prime \in \mathcal S_i\setminus\{u_i\}}  \hat{p}_{i, c^\prime}^{(t)}$, and the corresponding loss function is defined as:
\begin{equation}
\label{eq:cc}
	\mathcal{L}_{\text{CDL-CC}} = - \sum_{c \in \mathcal{S}_i}  w_{i, c}^{(t)} (1 - q_i^{(t)} + \max_{c^\prime \in \mathcal S_i\setminus\{u_i\}}  \hat{p}_{i, c^\prime}^{(t)} )^\gamma  \log (\hat{p}_{i, c}^{(t)}),
\end{equation}
and another in which $\Phi(\hat{\boldsymbol{p}}_{i}^{(t)})$ is the maximum predicted probability among non-candidate labels, \ie $\Phi(\hat{\boldsymbol{p}}_{i}^{(t)}) = \max_{\bar{c} \in \mathcal{\bar{S}}_i}  \hat{p}_{i, \bar{c}}^{(t)}$ is formulated as follows:
\begin{equation}
\label{eq:cn}
\mathcal{L}_{\text{CDL-CN}} = - \sum_{c \in \mathcal{S}_i}  w_{i, c}^{(t)} (1 - q_i^{(t)} + \max_{\bar{c} \in \mathcal{\bar{S}}_i}  \hat{p}_{i, \bar{c}}^{(t)} )^\gamma   \log (\hat{p}_{i, c}^{(t)}).
\end{equation}

The two CDL instantiations adjust the learning dynamics through the top-vs-competitor margin $\beta_i$. When the prediction is ambiguous and the top candidate is not well separated from its competitor, $\beta_i$ is small and the modulation $(1-\beta_i)^\gamma$ stays close to one. In this case, the loss is not down-weighted and the model keeps a strong disambiguation signal, which helps alleviate under-confidence by continuing to sharpen the distribution only after sufficient evidence is accumulated.
When the prediction becomes well separated, $\beta_i$ is large and $(1-\beta_i)^\gamma$ decreases, which reduces the incentive to further increase already confident probabilities and limits excessive sharpening, thereby mitigating over-confidence. Moreover, CDL is plug-and-play, enabling seamless integration with existing MIPL approaches. We propose a systematic integration of the two CDL instantiations with three established attention mechanisms: Disambiguation Attention Mechanism ({\dam}) \cite{tang2023demipl}, Scaled additive Attention Mechanism ({\sam}) \cite{tang2024elimipl}, and Margin-aware Attention Mechanism ({\mam}) \cite{tang2024miplma}. This integration yields six variants, \ie {\damcc}, {\damcn}, {\samcc}, {\samcn}, {\mamcc}, and {\mamcn}. \looseness=-1

\begin{algorithm}[!t]	
\caption{Pseudo-Code of CDL}
\label{alg:algorithm}
\begin{flushleft}
\textbf{Inputs}: \\
$\mathcal{D}$: MIPL training set $\{(\boldsymbol{X}_i,\mathcal{S}_i) \mid 1 \le i \le m \}$, where $\boldsymbol{X}_i = \{\boldsymbol{x}_{i,1}, \boldsymbol{x}_{i,2}, \ldots, \boldsymbol{x}_{i,n_i}\}$, $\boldsymbol{x}_{i,j} \in \mathcal{X}$, $\mathcal{X} = \mathbb{R}^d$, $\mathcal{S}_i \subset \mathcal{Y}$, and $\mathcal{Y} = \{1, 2, \ldots, k\}$ \\
$T$: Total number of training epochs \\
$\boldsymbol{X}_{*}$: Unseen multi-instance bag with $n_*$ instances \\
\textbf{Outputs}: \\
$\boldsymbol{Y}_{*}$: Predicted label for $\boldsymbol{X}_{*} = \{\boldsymbol{x}_{i,1}, \boldsymbol{x}_{i,2}, \ldots, \boldsymbol{x}_{i,n_*}\}$ \\
\textbf{Process}: 
\end{flushleft}
\begin{algorithmic}[1]  
\STATE Initialize the weights $w_{i,c}^{(1)} = \frac{1}{\left|\mathcal{S}_i\right|}$ for the candidate labels
\FOR{$t=1$ to $T$}
    \STATE Shuffle the training set $\mathcal{D}$ into $B$ mini-batches
    \FOR{$b=1$ to $B$}
        \STATE Learn instance-level features using $\Psi(\cdot)$
        \STATE Compute attention scores based on Eq. (\ref{eq:demipl_att}) for {\dam}, Eq. (\ref{eq:elimipl_att}) for {\sam}, or Eqs. (\ref{eq:miplma_att}, \ref{eq:tau}, \ref{eq:normalization}) for {\mam}
        \STATE Aggregate instance-level features into bag-level features according to Eq. (\ref{eq:aggregation})
        \STATE Classify the multi-instance bag and obtain predicted probabilities
        \STATE Update the weights $w_{i,c}^{(t)}$ using Eq. (\ref{eq:update_w})
        \STATE Compute the calibratable disambiguation loss according to Eq. (\ref{eq:cc}) or Eq. (\ref{eq:cn}) 
        \STATE Update model parameters with the SGD optimizer
    \ENDFOR  
\ENDFOR
\STATE Learn instance-level features of the unknown multi-instance bag $\boldsymbol{X}_{*}$ in the test set
\STATE Compute attention scores and aggregate instance-level features into a single vector $\boldsymbol{z}_*$
\STATE Return $Y_{*} = \underset{c \in \mathcal{Y}}{\arg \max}~\hat{p}_{*,c}$
\end{algorithmic}
\end{algorithm}

\subsection{Pseudo-Code of Calibratable Disambiguation Loss}
\label{sec:pseudo_code}
Based on the provided dataset and parameters, Algorithm \ref{alg:algorithm} presents the pseudo-code of our CDL. Initially, the algorithm uniformly initializes the weights of the candidate label set (Step $1$). During each epoch, the training data is partitioned into multiple mini-batches (Step $3$). The models then extract instance-level features and compute attention scores (Steps $5$-$6$). These features are aggregated into bag-level representations for each mini-batch (Step $7$). Subsequently, the model classifies each multi-instance bag and updates the weights of the candidate label set (Steps $8$-$9$). Finally, the calibratable disambiguation loss is computed, and the model parameters are updated (Steps $10$-$11$).

For unseen multi-instance bags, the process starts with extracting instance-level features using $\Psi(\cdot)$ (Step $14$). Next, these features are aggregated into a bag-level representation using the attention mechanism {\dam}, {\sam}, or {\mam} (Step $15$). Finally, the predicted label is obtained by selecting the category with the highest prediction probability (Step $16$).

\section{Theoretical Analysis}
\label{sec:theoretical}
This section theoretically analyzes the proposed calibratable disambiguation loss (CDL). The analysis is organized around three complementary questions. First, from the loss perspective, we rewrite CDL as a margin-modulated version of the momentum-based disambiguation loss (MDL) and show that CDL preserves an MDL-like fitting signal on ambiguous low-margin bags. Second, from the calibration perspective, we relate population top-label calibration to pseudo-label confidence alignment and then show that the CDL risk controls this alignment under a non-degenerate modulation condition. Third, from the optimization perspective, we show that CDL is not merely a fixed reweighting of MDL: its gradient contains an additional margin-shaping term, whose logit-level effect is to increase the active top-vs-competitor margin, and whose influence is carried into later pseudo-label weights through the momentum update. 
For notational simplicity, we omit the training-epoch superscript and the explicit dependence on model parameters unless needed.

\subsection{Notation and Margin-Modulated Form of CDL}
\label{subsec:theory_notation}
Let $(\boldsymbol X_i,\mathcal S_i)$ denote the $i$-th MIPL example, where $\boldsymbol X_i$ is a multi-instance bag and $\mathcal S_i\subseteq\mathcal Y=\{1,2,\ldots,k\}$ is its candidate label set. Let $Y_i\in\mathcal Y$ be the latent ground-truth label. The bag-level predictive distribution, latent posterior, and pseudo-label weight vector are denoted by
\begin{equation}
\label{eq:theory_indexed_notation}
\begin{aligned}
\hat{\boldsymbol p}_i &= \big(\hat p_{i,1},\hat p_{i,2},\ldots,\hat p_{i,k}\big)\in\Delta^{k-1},\\
\boldsymbol\eta_i &= \big(\eta_{i,1},\eta_{i,2},\ldots,\eta_{i,k}\big)\in\Delta^{k-1},\\
\boldsymbol w_i &= \big(w_{i,1},w_{i,2},\ldots,w_{i,k}\big)\in\Delta^{k-1},
\end{aligned}
\end{equation}
where
\begin{equation}
\label{eq:theory_indexed_posterior_support}
\eta_{i,c}=\mathbb P(Y_i=c\mid \boldsymbol X_i,\mathcal S_i),
\qquad
\eta_{i,c}=w_{i,c}=0\ \text{for all }c\notin\mathcal S_i.
\end{equation}

For each bag, define the top candidate label and its probability as follows:
\begin{equation}
\label{eq:theory_top_candidate}
u_i=\arg\max_{c\in\mathcal S_i}\hat p_{i,c},
\qquad
q_i=\hat p_{i,u_i}=\max_{c\in\mathcal S_i}\hat p_{i,c}.
\end{equation}
For CDL-CC, the competitor is the second strongest candidate label, \ie $\phi_{i,\mathrm{CC}} = \max_{c^\prime \in\mathcal S_i\setminus\{u_i\}}\hat p_{i,c^\prime}$. For CDL-CN, the competitor is the strongest non-candidate label, \ie $\phi_{i,\mathrm{CN}} = \max_{\bar c\in\bar{\mathcal S}_i}\hat p_{i,\bar c}$.
For either CDL instantiation, write $\phi_i=\Phi(\hat{\boldsymbol p}_i)$, and define the top-vs-competitor margin and the CDL modulation factor by  \looseness=-1
\begin{equation}
\label{eq:theory_margin_modulation}
\beta_i=q_i - \phi_i,
\qquad
\lambda_i=\big(1-\beta_i\big)^\gamma.
\end{equation}
The per-bag MDL objective induced by the current pseudo-label weights is
\begin{equation}
\label{eq:theory_mdl_loss}
\ell_i^{\mathrm{MDL}}
=
-\sum_{c\in\mathcal S_i}w_{i,c}\log \hat p_{i,c}.
\end{equation}
With the notation in Eq. \eqref{eq:theory_margin_modulation}, the per-bag CDL objective can be written as a margin-modulated MDL objective:
\begin{equation}
\label{eq:theory_cdl_as_modulated_mdl}
\ell_i^{\mathrm{CDL}}
=
\lambda_i\ell_i^{\mathrm{MDL}}
=
\big(1-\beta_i\big)^\gamma
\left(
-\sum_{c\in\mathcal S_i}w_{i,c}\log \hat p_{i,c}
\right).
\end{equation}
Because $\boldsymbol w_i$ is supported on $\mathcal S_i$, the MDL objective admits the following KL-entropy decomposition:
\begin{equation}
\label{eq:theory_mdl_kl_entropy}
\begin{aligned}
\ell_i^{\mathrm{MDL}}
&=
\mathrm{KL}\!\left(\boldsymbol w_i\middle\|\hat{\boldsymbol p}_i\right)
+
\mathbb H\!\left(\boldsymbol w_i\right),
\\
\mathbb H\!\left(\boldsymbol w_i\right)
&=
-\sum_{c\in\mathcal S_i} w_{i,c}\log w_{i,c}.
\end{aligned}
\end{equation}
For a training set of $m$ bags, define the empirical MDL and CDL objectives by
\begin{equation}
\label{eq:theory_empirical_losses}
\mathcal L_{\mathrm{MDL}}
=
\frac{1}{m}\sum_{i=1}^{m}\ell_i^{\mathrm{MDL}},
\qquad
\mathcal L_{\mathrm{CDL}}
=
\frac{1}{m}\sum_{i=1}^{m}\ell_i^{\mathrm{CDL}}.
\end{equation}

\subsection{Linear Lower Bound of CDL}
\label{subsec:theory_lower_bound}
The first result formalizes the loss-level connection between CDL and MDL. It shows that the modulation does not remove the MDL fitting signal on low-margin bags, which is important for preserving disambiguation while allowing the modulation to down-weight already separated predictions.  

\begin{theorem}[Linear lower bound of CDL]
\label{thm:cdl_linear_lower_bound}
Let $\gamma\ge 1$. For the $i$-th training bag, assume that $\hat{\boldsymbol p}_i\in\Delta^{k-1}$, $\hat p_{i,c}>0$ for all $c\in\mathcal Y$, and $\boldsymbol w_i\in\Delta^{k-1}$ with $w_{i,c}=0$ for all $c\notin\mathcal S_i$. Let $u_i=\arg\max_{c\in\mathcal S_i}\hat p_{i,c}$, $q_i=\hat p_{i,u_i}=\max_{c\in\mathcal S_i}\hat p_{i,c}$. Let
\begin{equation}
\label{eq:app_thm1_mdl_cdl_def}
\ell_i^{\mathrm{MDL}}=-\sum_{c\in\mathcal S_i}w_{i,c}\log \hat p_{i,c},\qquad \ell_i^{\mathrm{CDL}}=\lambda_i\ell_i^{\mathrm{MDL}} .
\end{equation}
Then, for each training bag,
\begin{equation}
\label{eq:app_thm1_per_bag_bound}
\ell_i^{\mathrm{CDL}}\ge (1-\gamma\beta_i)\ell_i^{\mathrm{MDL}}=(1-\gamma\beta_i)\left[\mathrm{KL}\!\left(\boldsymbol w_i\middle\|\hat{\boldsymbol p}_i\right)+\mathbb H(\boldsymbol w_i)\right].
\end{equation}
Consequently,
\begin{equation}
\label{eq:thm1_empirical_weighted_bound}
\mathcal L_{\mathrm{CDL}}\ge \frac{1}{m}\sum_{i=1}^{m}(1-\gamma\beta_i)\ell_i^{\mathrm{MDL}} .
\end{equation}
Moreover, if $\beta_{\max}=\max_{1\le i\le m}\beta_i$, then
\begin{equation}
\label{eq:app_thm1_dataset_bound}
\mathcal L_{\mathrm{CDL}}\ge (1-\gamma\beta_{\max})\mathcal L_{\mathrm{MDL}}.
\end{equation}
\end{theorem}

Theorem~\ref{thm:cdl_linear_lower_bound} shows that CDL retains an MDL-like disambiguation when the margin $\beta_i$ is small. When $\beta_i>0$ is large, the modulation $\lambda_i=(1-\beta_i)^\gamma$ reduces the loss contribution of already well-separated bags. For CDL-CN, the case $\phi_{i,\mathrm{CN}}>q_i$ gives $\beta_i<0$, and CDL intentionally amplifies the loss $\phi_{i,\mathrm{CN}}>q_i \Longrightarrow \lambda_i = \big(1+\phi_{i,\mathrm{CN}}-q_i\big)^\gamma > 1$.
Therefore, CDL-CC mainly suppresses separated candidate-level predictions, while CDL-CN can additionally penalize bags whose non-candidate probabilities dominate the top candidate probability. Having established that CDL remains tied to the MDL fitting objective, we next connect this pseudo-label fitting view to the population top-label calibration. \looseness=-1

\subsection{Population Top-Label Calibration and Pseudo-Label Approximation}
\label{subsec:theory_calibration_target}
We use a generic MIPL draw $(\boldsymbol X_i, \mathcal S_i,Y_i)$ to keep the population analysis consistent with the empirical notation. The top-label prediction and its confidence score are defined over the full label space by
\begin{equation}
\label{eq:theory_indexed_top_confidence}
\hat y_i = \arg\max_{c\in\mathcal Y}\hat p_{i,c},
\qquad
C_i = \hat p_{i,\hat y_i} = \max_{c\in\mathcal Y}\hat p_{i,c}.
\end{equation}
Here $\hat y_i$ is used for calibration and may differ from the top candidate label used by CDL. The population top-label calibration error is
\begin{equation}
\label{eq:theory_indexed_population_calibration}
E_{\mathrm{cal}} = \mathbb E \left[\left| \mathbb E\!\left[\mathbb I\{Y_i=\hat y_i\}\mid C_i\right]-C_i \right| \right].
\end{equation}
The empirical ECE in Eq.~\eqref{eq:ece} is a binned finite-sample approximation of Eq.~\eqref{eq:theory_indexed_population_calibration}. To connect calibration with pseudo-label learning, define
\begin{equation}
\label{eq:theory_indexed_confidence_errors}
\begin{aligned}
E_{\mathrm{conf}} &= \mathbb E \left[ \left|\eta_{i,\hat y_i}-C_i\right| \right],
\\
E_{\mathrm{pconf}} &= \mathbb E \left[ \left|w_{i,\hat y_i}-C_i\right| \right],
\\
\delta_w^{\mathrm{TV}} &= \mathbb E \left[ d_{\mathrm{TV}}(\boldsymbol\eta_i,\boldsymbol w_i) \right].
\end{aligned}
\end{equation}
The total variation distance is
\begin{equation}
\label{eq:theory_indexed_tv_definition}
d_{\mathrm{TV}}(\boldsymbol\eta_i,\boldsymbol w_i) = \frac{1}{2}\|\boldsymbol\eta_i-\boldsymbol w_i\|_1.
\end{equation}

\begin{proposition}[Calibration is controlled by pseudo-label confidence error]
\label{prop:calibration_confidence_pseudo}
Fix the current training state. Assume that $\hat{\boldsymbol p}_i$ and $\boldsymbol w_i$ are measurable with respect to $\mathcal G_i=\sigma(\boldsymbol X_i, \mathcal S_i)$. Then
\begin{equation}
\label{eq:prop_indexed_main_bound}
E_{\mathrm{cal}} \le E_{\mathrm{conf}} \le E_{\mathrm{pconf}}+\delta_w^{\mathrm{TV}}.
\end{equation}
\end{proposition}

Proposition~\ref{prop:calibration_confidence_pseudo} provides an upper-bound decomposition of the calibration error. Specifically, $E_{\mathrm{cal}}$ is controlled by the pseudo-label confidence error $E_{\mathrm{pconf}}$, which reflects the discrepancy between the model confidence and the assigned pseudo-label weight, and the posterior approximation error $\delta_w^{\mathrm{TV}}$, which measures the quality of the pseudo-label weights as approximations to the true posterior distribution. Thus, this result identifies the optimizable component that CDL can influence and separates it from the quality of the pseudo-label approximation. The following theorem then links this optimizable component directly to the CDL risk. \looseness=-1

\subsection{Confidence Alignment Bound for CDL}
\label{subsec:theory_confidence_bound}
We next show that CDL controls $E_{\mathrm{pconf}}$ under a non-degenerate modulation condition. For the same generic bag $\boldsymbol X_i$, the corresponding MDL and CDL losses are shown as Eqs. \eqref{eq:theory_mdl_loss} and \eqref{eq:theory_cdl_as_modulated_mdl}. 
Define the population risks and the expected pseudo-label entropy as
\begin{equation}
\label{eq:theory_indexed_population_risks}
R_{\mathrm{MDL}}^w=\mathbb E\left[\ell_i^{\mathrm{MDL}}\right], ~~
R_{\mathrm{CDL}}^w=\mathbb E\left[\ell_i^{\mathrm{CDL}}\right], ~~ 
\mathcal H_w=\mathbb E\left[\mathbb H(\boldsymbol w_i)\right],
\end{equation}
where
\begin{equation}
\label{eq:theory_indexed_entropy}\mathbb H(\boldsymbol w_i)=-\sum_{c\in\mathcal S_i}w_{i,c}\log w_{i,c}.
\end{equation}

\begin{theorem}[Pseudo-label confidence alignment bound]
\label{thm:pseudo_label_confidence_alignment}
Assume that $\hat{\boldsymbol p}_i$ is induced by finite logits, so that $\hat p_{i,c}>0$ for all $c\in\mathcal Y$. Assume further that the relevant competitor set in CDL is nonempty and that the CDL modulation satisfies $\lambda_i\ge\lambda_0>0$. Then
\begin{equation}
\label{eq:thm_indexed_pconf_bound}
E_{\mathrm{pconf}}\le\sqrt{2\left(\lambda_0^{-1}R_{\mathrm{CDL}}^w-\mathcal H_w\right)}.
\end{equation}
Consequently,
\begin{equation}
\label{eq:thm_indexed_calibration_bound}
E_{\mathrm{cal}}\le E_{\mathrm{conf}}\le\sqrt{2\left(\lambda_0^{-1}R_{\mathrm{CDL}}^w-\mathcal H_w\right)}+\delta_w^{\mathrm{TV}}.
\end{equation}
\end{theorem}

Theorem~\ref{thm:pseudo_label_confidence_alignment} is conditional on the current pseudo-label weights. If $\boldsymbol w_i$ is close to the latent posterior $\boldsymbol\eta_i$, then $\delta_w^{\mathrm{TV}}$ is small and controlling the CDL risk improves confidence alignment. If the pseudo-label weights are inaccurate, the bound remains valid but can be loose. The assumption $\lambda_i\ge\lambda_0>0$ is a non-degeneracy condition preventing the CDL modulation from vanishing. The result explains why reducing CDL can improve the optimizable calibration component, while Proposition~\ref{prop:calibration_confidence_pseudo} clarifies that the final calibration bound also depends on the posterior quality of the pseudo-label weights. We next move from this risk-level view to the local optimization dynamics induced by the margin-dependent modulation.

\subsection{Optimization-Level Margin Shaping and Momentum Dynamics}
\label{subsec:theory_optimization_dynamics}

Although Eq. \eqref{eq:theory_cdl_as_modulated_mdl} writes CDL as a margin-modulated MDL objective, the modulation factor depends on the current prediction and hence on the model parameters. Therefore, CDL is not merely a fixed sample reweighting of MDL. This subsection makes this distinction explicit at the optimization level. We first decompose the CDL gradient into an MDL fitting term and an additional margin-shaping term, then interpret the latter in the softmax-logit space, and finally show how prediction-level margins are recursively transferred to future pseudo-label margins through the momentum update. Throughout this subsection, $\nabla$ denotes the gradient with respect to the model parameters.

\begin{proposition}[Gradient decomposition of CDL on differentiable regions]
\label{prop:gradient_decomposition}
Fix a training bag $(\boldsymbol X_i,\mathcal S_i)$, and let $\theta$ denote the model parameters. Suppose that there is an open parameter region $\mathcal U$ on which each $\hat p_{i,c}(\theta)$ is differentiable and strictly positive. Assume that the top candidate label $u_i=\arg\max_{c\in\mathcal S_i}\hat p_{i,c}(\theta)$ is uniquely attained and remains unchanged on $\mathcal U$. For CDL-CC, let $\mathcal C_i=\mathcal S_i\setminus\{u_i\}$; for CDL-CN, let $\mathcal C_i=\bar{\mathcal S}_i$. Assume that $\mathcal C_i\neq\varnothing$ and that the competitor $v_i=\arg\max_{c\in\mathcal C_i}\hat p_{i,c}(\theta)$ is also uniquely attained and remains unchanged on $\mathcal U$. Define
\begin{equation}
\label{eq:theory_active_qr_beta_lambda_1}
q_i(\theta)=\hat p_{i,u_i}(\theta),\qquad \phi_i(\theta)=\hat p_{i,v_i}(\theta),
\end{equation}
\begin{equation}
\label{eq:theory_active_qr_beta_lambda_2}
\beta_i(\theta)=q_i(\theta)-\phi_i(\theta),\qquad \lambda_i(\theta)=\big(1-\beta_i(\theta)\big)^\gamma.
\end{equation}
During the current gradient computation, regard the pseudo-label weights $\boldsymbol w_i$ as fixed, and write
\begin{equation}
\label{eq:theory_mdl_cdl_theta}
\ell_i^{\mathrm{MDL}}(\theta)=-\sum_{c\in\mathcal S_i}w_{i,c}\log \hat p_{i,c}(\theta),~~~
\ell_i^{\mathrm{CDL}}(\theta)=\lambda_i(\theta)\ell_i^{\mathrm{MDL}}(\theta).
\end{equation}
Then, for every $\theta\in\mathcal U$,
\begin{equation}
\label{eq:theory_gradient_decomposition}
\begin{aligned}
\nabla_{\theta}\ell_i^{\mathrm{CDL}}(\theta)
&= \lambda_i(\theta)\nabla_{\theta}\ell_i^{\mathrm{MDL}}(\theta) \\
&\quad-\gamma\big(1-\beta_i(\theta)\big)^{\gamma-1}\ell_i^{\mathrm{MDL}}(\theta)\nabla_{\theta}\beta_i(\theta).
\end{aligned}
\end{equation}
\end{proposition}

Proposition~\ref{prop:gradient_decomposition} is the basic differential identity behind the optimization-level effect of CDL. The first term in \eqref{eq:theory_gradient_decomposition} is the MDL fitting term scaled by the current modulation factor, whereas the second term is absent from MDL. In a gradient-descent step, this second term contributes the update component
\begin{equation}
\label{eq:theory_margin_update_component}
\delta_{\beta}
=
\eta\gamma\big(1-\beta_i(\theta)\big)^{\gamma-1}
\ell_i^{\mathrm{MDL}}(\theta)\nabla_{\theta}\beta_i(\theta),
\end{equation}
where $\eta>0$ is the step size. Therefore, whenever $\nabla_{\theta}\beta_i(\theta)\neq 0$, a first-order Taylor expansion gives
\begin{equation}
\label{eq:theory_margin_first_order_increase}
\beta_i(\theta+\delta_{\beta})-\beta_i(\theta)
=
\eta\gamma\big(1-\beta_i(\theta)\big)^{\gamma-1}
\ell_i^{\mathrm{MDL}}(\theta)
\|\nabla_{\theta}\beta_i(\theta)\|^2
+o(\eta).
\end{equation}
Therefore, the additional CDL component explicitly promotes an increase of the active top-vs-competitor margin. At exact ties, the max-based margin is not differentiable; the differentiable statement applies on each region with a fixed active top candidate and competitor. The following corollary translates this abstract parameter-space margin term into a concrete logit-space effect.

\begin{corollary}[Logit-space effect of margin shaping]
\label{cor:logit_margin_direction}
Fix a training bag $i$. Let $s_{i,c}\in\mathbb R$ be the logit of class $c$, and let
\begin{equation}
\label{eq:theory_cor3_softmax}
\hat p_{i,c}
=
\frac{\exp(s_{i,c})}{\sum_{a\in\mathcal Y}\exp(s_{i,a})},
\qquad c\in\mathcal Y .
\end{equation}
Assume that, in a neighborhood of the current logits, the top candidate $u$ and the competitor $v$ are unique, distinct, and fixed. Then $\beta_i=\hat p_{i,u}-\hat p_{i,v}$ is differentiable in this neighborhood, and
\begin{align}
\label{eq:theory_cor3_top}
\frac{\partial \beta_i}{\partial s_{i,u}}
&=\hat p_{i,u}(1-\hat p_{i,u}+\hat p_{i,v})>0,\\
\label{eq:theory_cor3_competitor}
\frac{\partial \beta_i}{\partial s_{i,v}}
&=-\hat p_{i,v}(1+\hat p_{i,u}-\hat p_{i,v})<0,\\
\label{eq:theory_cor3_other}
\frac{\partial \beta_i}{\partial s_{i,c}}
&=\hat p_{i,c}(\hat p_{i,v}-\hat p_{i,u}),~~~~ c\notin\{u,v\}.
\end{align}
Thus a positive step along $\nabla_{s_i}\beta_i$ locally increases the top-candidate logit and decreases the active-competitor logit.
\end{corollary}

Corollary~\ref{cor:logit_margin_direction} provides the logit-space interpretation of the margin-shaping term in Proposition~\ref{prop:gradient_decomposition}. It shows that increasing the active margin locally raises the top-candidate logit and suppresses the active-competitor logit. The preceding two results describe the current optimization step. However, the pseudo-label weights are updated across epochs by a momentum rule in MIPL. The following lemma records how candidate-level prediction margins are transferred into future pseudo-label margins. It is stated for two candidate labels $u,v\in\mathcal S_i$, because the pseudo-label weight is supported on the candidate set. For CDL-CN, the non-candidate competitor affects this recursion indirectly through the candidate-normalized predictions. \looseness=-1

\begin{lemma}[Momentum recursion for candidate pseudo-label margins]
\label{lem:pseudo_label_margin}
Fix a training bag $(\boldsymbol X_i,\mathcal S_i)$ and two candidate labels $u,v\in\mathcal S_i$. For $t=2,\ldots,T$, define
\begin{equation}
\label{eq:theory_margin_dynamic_def_compact}
\tilde p_{i,c}^{(t)}
=
\frac{\hat p_{i,c}^{(t)}}{\sum_{a\in\mathcal S_i}\hat p_{i,a}^{(t)}},
~~~
M_{i,uv}^{(t)}=w_{i,u}^{(t)}-w_{i,v}^{(t)},
~~~
\Delta_{i,uv}^{(t)}=\tilde p_{i,u}^{(t)}-\tilde p_{i,v}^{(t)} .
\end{equation}
Assume that the pseudo-label weights are updated by
\begin{equation}
\label{eq:theory_weight_dynamic_update}
w_{i,c}^{(t)}
=
\alpha^{(t)}w_{i,c}^{(t-1)}
+
\big(1-\alpha^{(t)}\big)\tilde p_{i,c}^{(t)},
~~~
c\in\mathcal S_i .
\end{equation}
Then, for every $t=2,\ldots,T$,
\begin{equation}
\label{eq:theory_margin_dynamic_one_step}
M_{i,uv}^{(t)}
=
\alpha^{(t)}M_{i,uv}^{(t-1)}
+
\big(1-\alpha^{(t)}\big)\Delta_{i,uv}^{(t)} .
\end{equation}
Equivalently, with $A_{a:b}=\prod_{\tau=a}^{b}\alpha^{(\tau)}$ and the empty product defined as one,
\begin{equation}
\label{eq:theory_margin_dynamic_closed_form_compact}
M_{i,uv}^{(t)}
=
A_{2:t}M_{i,uv}^{(1)}
+
\sum_{s=2}^{t}
\big(1-\alpha^{(s)}\big)A_{s+1:t}\Delta_{i,uv}^{(s)} .
\end{equation}
Moreover,
\begin{equation}
\label{eq:theory_delta_raw_prediction_margin}
\Delta_{i,uv}^{(s)}
=
\frac{\hat p_{i,u}^{(s)}-\hat p_{i,v}^{(s)}}
{\sum_{a\in\mathcal S_i}\hat p_{i,a}^{(s)}} .
\end{equation}
Thus past candidate prediction margins enter the current pseudo-label margin through the momentum coefficients. If $\alpha^{(s)}\in[0,1]$ for all $s$, then $M_{i,uv}^{(t)}$ is a convex combination of $M_{i,uv}^{(1)}$ and $\Delta_{i,uv}^{(2)},\ldots,\Delta_{i,uv}^{(t)}$.
\end{lemma}

Lemma~\ref{lem:pseudo_label_margin} complements the local gradient analysis by showing that the margins shaped at the prediction level are not transient. They enter the subsequent pseudo-label weights through the momentum rule. This completes the optimization-level explanation of CDL.

Taken together, the above results provide a coherent account of CDL from three complementary perspectives. At the loss level, the margin-modulated form in Eq. \eqref{eq:theory_cdl_as_modulated_mdl} and Theorem~\ref{thm:cdl_linear_lower_bound} relate CDL to MDL and clarify why low-margin bags still receive an MDL-like disambiguation signal. At the calibration level, Proposition~\ref{prop:calibration_confidence_pseudo} separates the calibration error into a pseudo-label confidence-alignment term and a pseudo-label approximation term, while Theorem~\ref{thm:pseudo_label_confidence_alignment} connects the former to the CDL risk. At the optimization level, Proposition~\ref{prop:gradient_decomposition}, Corollary~\ref{cor:logit_margin_direction}, and Lemma~\ref{lem:pseudo_label_margin} explain how the margin-dependent term reshapes the active top-vs-competitor margin and how this effect is propagated to subsequent pseudo-label updates.

These theoretical implications motivate the empirical evaluation in the next section. In particular, the analysis suggests that CDL should preserve the disambiguation ability of MDL, mitigate excessive confidence growth through margin modulation, and transmit margin information across epochs through the momentum update. Therefore, we evaluate CDL on benchmark and real-world MIPL datasets in terms of both classification accuracy and calibration quality.

\begin{table*}[!t] 
 \caption{Characteristics of the benchmark and real-world MIPL datasets. The C-R34-9 is introduced in this work as a new MIPL dataset. ``n/a'' is abbreviated for not applicable as the instance-level labels of the CRC-{\scriptsize{MIPL}} dataset are unknown.}
\label{tab:datasets}
\centering 
\footnotesize
\begin{tabular}{lccccccccccc}
\hline \hline
	\multicolumn{1}{l}{Dataset}				& \multicolumn{1}{c}{\#bag}		& \multicolumn{1}{c}{\#ins} 		& \multicolumn{1}{c}{max. \#ins}		& \multicolumn{1}{c}{min. \#ins}			& \multicolumn{1}{c}{avg. \#ins}	 	& \multicolumn{1}{c}{max. \#pos}		& \multicolumn{1}{c}{min. \#pos}			& \multicolumn{1}{c}{avg. \#pos}		& \multicolumn{1}{c}{\#dim}		& \multicolumn{1}{c}{\#C} 		& \multicolumn{1}{c}{avg. \#CLs}	\\
	\hline
MNIST-{\scriptsize{MIPL}} 			& \multicolumn{1}{c}{500}			& \multicolumn{1}{c}{20664}		& \multicolumn{1}{c}{48}				& \multicolumn{1}{c}{35}				& \multicolumn{1}{c}{41.33}		& \multicolumn{1}{c}{9.1\%}				& \multicolumn{1}{c}{7.0\%}				& \multicolumn{1}{c}{8.0\%}		& \multicolumn{1}{c}{784}			& \multicolumn{1}{c}{5}			& \multicolumn{1}{c}{{2, 3, 4}}		\\
FMNIST-{\scriptsize{MIPL}} 			& \multicolumn{1}{c}{500}			& \multicolumn{1}{c}{20810}		& \multicolumn{1}{c}{48}				& \multicolumn{1}{c}{36}				& \multicolumn{1}{c}{41.62} 		& \multicolumn{1}{c}{9.1\%}				& \multicolumn{1}{c}{7.0\%}				& \multicolumn{1}{c}{8.0\%}	&  \multicolumn{1}{c}{784}			& \multicolumn{1}{c}{5}			& \multicolumn{1}{c}{{2, 3, 4}}		\\
Birdsong-{\scriptsize{MIPL}} 			& \multicolumn{1}{c}{1300}		& \multicolumn{1}{c}{48425}		& \multicolumn{1}{c}{76}				& \multicolumn{1}{c}{25}				& \multicolumn{1}{c}{37.25}		& \multicolumn{1}{c}{10.0\%}				& \multicolumn{1}{c}{6.9\%}				& \multicolumn{1}{c}{8.3\%}		& \multicolumn{1}{c}{38}			& \multicolumn{1}{c}{13}			& \multicolumn{1}{c}{{2, 3, 4}}		\\
SIVAL-{\scriptsize{MIPL}} 				& \multicolumn{1}{c}{1500}		& \multicolumn{1}{c}{47414}		& \multicolumn{1}{c}{32}				& \multicolumn{1}{c}{31}				& \multicolumn{1}{c}{31.61} 		& \multicolumn{1}{c}{90.62\%}				& \multicolumn{1}{c}{3.12\%}				& \multicolumn{1}{c}{25.6\%}		& \multicolumn{1}{c}{30}			& \multicolumn{1}{c}{25}			& \multicolumn{1}{c}{{2, 3, 4}}		\\
	  \hdashline 
C-Row				& \multicolumn{1}{c}{7000}		& \multicolumn{1}{c}{56000}		& \multicolumn{1}{c}{8}				& \multicolumn{1}{c}{8}				& \multicolumn{1}{c}{8}			& \multicolumn{1}{c}{n/a}				& \multicolumn{1}{c}{n/a}				& \multicolumn{1}{c}{n/a}  	& \multicolumn{1}{c}{9}			& \multicolumn{1}{c}{7}			& \multicolumn{1}{c}{2.08}		\\
C-SBN				& \multicolumn{1}{c}{7000}		& \multicolumn{1}{c}{63000}		& \multicolumn{1}{c}{9}				& \multicolumn{1}{c}{9}				& \multicolumn{1}{c}{9}			& \multicolumn{1}{c}{n/a}				& \multicolumn{1}{c}{n/a}				& \multicolumn{1}{c}{n/a} 	& \multicolumn{1}{c}{15}			& \multicolumn{1}{c}{7}			& \multicolumn{1}{c}{2.08}		\\
C-KMeans	& \multicolumn{1}{c}{7000}		& \multicolumn{1}{c}{30178}		& \multicolumn{1}{c}{6}				& \multicolumn{1}{c}{3}				& \multicolumn{1}{c}{4.311}		& \multicolumn{1}{c}{n/a}				& \multicolumn{1}{c}{n/a}				& \multicolumn{1}{c}{n/a} 	& \multicolumn{1}{c}{6}			& \multicolumn{1}{c}{7}			& \multicolumn{1}{c}{2.08}			\\
C-SIFT				& \multicolumn{1}{c}{7000}		& \multicolumn{1}{c}{175000}		& \multicolumn{1}{c}{25}				& \multicolumn{1}{c}{25}				& \multicolumn{1}{c}{25}			& \multicolumn{1}{c}{n/a}				& \multicolumn{1}{c}{n/a}				& \multicolumn{1}{c}{n/a} 	&\multicolumn{1}{c}{128}			& \multicolumn{1}{c}{7}			& \multicolumn{1}{c}{2.08}		\\
C-R34-9	& \multicolumn{1}{c}{7000}		& \multicolumn{1}{c}{63000}		& \multicolumn{1}{c}{9}				& \multicolumn{1}{c}{9}				& \multicolumn{1}{c}{9}			& \multicolumn{1}{c}{n/a}				& \multicolumn{1}{c}{n/a}				& \multicolumn{1}{c}{n/a} 	&\multicolumn{1}{c}{1000}			& \multicolumn{1}{c}{7}			& \multicolumn{1}{c}{2.08}		\\
C-R34-16	& \multicolumn{1}{c}{7000}		& \multicolumn{1}{c}{112000}		& \multicolumn{1}{c}{16}				& \multicolumn{1}{c}{16}				& \multicolumn{1}{c}{16}			& \multicolumn{1}{c}{n/a}				& \multicolumn{1}{c}{n/a}				& \multicolumn{1}{c}{n/a} 	&\multicolumn{1}{c}{1000}			& \multicolumn{1}{c}{7}			& \multicolumn{1}{c}{2.08}		\\
C-R34-25 	& \multicolumn{1}{c}{7000}		& \multicolumn{1}{c}{175000}		& \multicolumn{1}{c}{25}				& \multicolumn{1}{c}{25}				& \multicolumn{1}{c}{25}			& \multicolumn{1}{c}{n/a}				& \multicolumn{1}{c}{n/a}				& \multicolumn{1}{c}{n/a} 	&\multicolumn{1}{c}{1000}			& \multicolumn{1}{c}{7}			& \multicolumn{1}{c}{2.08}		\\
	  \hline \hline
  \end{tabular}
\end{table*}

\section{Experiments}
\label{sec:experiments}

\subsection{Experimental Configurations}
\label{subsec:configurations}
\subsubsection{Datasets}
We adopt the experimental protocol established by the prior research \cite{tang2023demipl}, which includes a diverse range of datasets, including four benchmark datasets and seven real-world datasets. Specifically, the benchmark datasets are MNIST-{\scriptsize{MIPL}}, FMNIST-{\scriptsize{MIPL}}, Birdsong-{\scriptsize{MIPL}}, and SIVAL-{\scriptsize{MIPL}}, which span various domains from image analysis to biological data \cite{lecun1998gradient, han1708, briggs2012rank, SettlesCR07}. 
Additionally, we include CRC-{\scriptsize{MIPL}}, a real-world dataset for colorectal cancer classification. The candidate label sets are curated by professionally trained crowdsourcing workers. In the prior research \cite{tang2023demipl, tang2024elimipl}, four types of multi-instance features have been employed across different sub-datasets: CRC-{\scriptsize{MIPL-Row}} (C-Row), CRC-{\scriptsize{MIPL-SBN}} (C-SBN), CRC-{\scriptsize{MIPL-KMeansSeg}} (C-KMeans), and CRC-{\scriptsize{MIPL-SIFT}} (C-SIFT). These features are generated by four image bag generators \cite{WeiZ16}, including Row, single blob with neighbors (SBN), k-means segmentation (KMeansSeg), and scale-invariant feature transform (SIFT). 

Furthermore, a novel feature extraction strategy is proposed for the CRC-{\scriptsize{MIPL}} dataset by leveraging ResNet-34. Each image is partitioned into $N$ non-overlapping patches, and ResNet-34 is applied to obtain feature representations for each patch. Three configurations are investigated, with $N$ set to $9$, $16$, and $25$, resulting in the new datasets C-R34-9, C-R34-16, and C-R34-25, respectively. Notably, the C-R34-9 dataset is first introduced in this work. Table~\ref{tab:datasets} provides further details on these datasets.  More information on the benchmark datasets and the CRC-{\scriptsize{MIPL}} dataset can be found in the literature \cite{tang2023miplgp} and \cite{tang2023demipl}, respectively.

\subsubsection{Comparative Methods}
To our knowledge, there are six available methods relevant to our MIPL settings, namely {\miplgp} \cite{tang2023miplgp}, {\demipl} \cite{tang2023demipl}, {\elimipl} \cite{tang2024elimipl}, {\miplma} \cite{tang2024miplma}, {\promipl} \cite{yang2024promipl}, and {\fastmipl} \cite{yang2025fastmipl}. All corresponding code implementations for these methods are publicly available. We systematically compare our six methods against the six MIPL methods to evaluate their classification accuracy and calibration performance. The parameters for all compared methods were set following the recommendations from the original literature or were further optimized to enhance performance.

\subsubsection{Implementation Details}
Our algorithm is implemented using PyTorch and executed on an NVIDIA Tesla V100 GPU. We utilize stochastic gradient descent (SGD) with a momentum of $0.9$ and a weight decay of $0.0001$ for optimization. We adopt the same feature extractor $\Psi(\cdot)$ as used in previous MIPL methods \cite{tang2023demipl, tang2024elimipl, tang2024miplma, yang2024promipl, yang2025fastmipl}. For MNIST-{\scriptsize{MIPL}} and FMNIST-{\scriptsize{MIPL}}, we extract instance-level features using a two-layer convolutional neural network followed by a fully connected network. For preprocessed datasets, \ie Birdsong-{\scriptsize{MIPL}} and SIVAL-{\scriptsize{MIPL}}, we use a fully connected network for feature extraction. For CRC-{\scriptsize{MIPL}}, we explore four image bag generators and use ResNet-34 to extract instance-level features. We select the learning rate from $\{0.01, 0.05\}$ and adjust it using cosine annealing. {\damcc} and {\damcn} are trained for $200$ epochs on CRC-{\scriptsize{MIPL}} and $100$ epochs on benchmark datasets. Meanwhile, {\samcc}, {\samcn}, {\mamcc}, and {\mamcn} are trained for $100$ epochs on all datasets. We set the CDL parameter $\gamma$ to $1$ in all experiments. We report the mean and standard deviation (std) of classification accuracy and expected calibration error over ten random train/test splits with a ratio of $\text{7:3}$. In our experiments, the expected calibration error is evaluated with $15$ bins. For each dataset, the highest classification accuracy and the lowest expected calibration error are highlighted in bold to facilitate comparison. In the following tables, $\uparrow$ and $\downarrow$ denote improvements and reductions over {\demipl}~\cite{tang2023demipl}, {\elimipl}~\cite{tang2024elimipl}, or {\miplma}~\cite{tang2024miplma}, respectively.
The code of this work can be found at \url{https://github.com/tangw-seu/MIPLCDL}.

\begin{table*}[!t]
  \centering  
   \caption{Classification accuracy (mean$\pm$std\%) on the benchmark datasets with varying numbers of false-positive labels.}
\label{tab:res_benchmark}
    \begin{tabular}{lcccccccc}
    \hline  \hline
    \multicolumn{1}{l}{} 	& \multicolumn{2}{c}{MNIST-{\tiny{MIPL}}} 					& \multicolumn{2}{c}{FMNIST-{\tiny{MIPL}}} 				& \multicolumn{2}{c}{Birdsong-{\tiny{MIPL}}} 				& \multicolumn{2}{c}{SIVAL-{\tiny{MIPL}}} \\
     \hline	
    \multicolumn{9}{c}{$r=1$} \\
    \hline
    {\miplgp}     		& \multicolumn{2}{c}{94.87$\pm$1.55} 		& \multicolumn{2}{c}{84.67$\pm$2.98}		& \multicolumn{2}{c}{71.62$\pm$2.59}	& \multicolumn{2}{c}{66.87$\pm$1.95} \\
    {\promipl}     		& \multicolumn{2}{c}{99.92$\pm$0.29} 		& \multicolumn{2}{c}{92.20$\pm$2.39}		& \multicolumn{2}{c}{77.63$\pm$1.50}		& \multicolumn{2}{c}{68.20$\pm$3.22} \\
    {\fastmipl}     		& \multicolumn{2}{c}{99.91$\pm$0.23} 		& \multicolumn{2}{c}{90.99$\pm$2.19}		& \multicolumn{2}{c}{79.60$\pm$2.35}		& \multicolumn{2}{c}{\textbf{77.65$\pm$3.02}} \\
    \hdashline
    {\demipl}           & \multicolumn{2}{c}{97.60$\pm$0.80}		& \multicolumn{2}{c}{88.01$\pm$2.08} 	& \multicolumn{2}{c}{74.36$\pm$1.57}		& \multicolumn{2}{c}{63.53$\pm$4.10}	 \\
    {\damcc} (ours)      	& \multicolumn{2}{c}{99.33$\pm$0.52 ($1.73\uparrow$)} 			& \multicolumn{2}{c}{92.07$\pm$2.18 ($4.06\uparrow$)} 			& \multicolumn{2}{c}{79.03$\pm$1.49 ($4.67\uparrow$)}			& \multicolumn{2}{c}{73.87$\pm$2.25 ($10.34\uparrow$)}	 \\
    {\damcn} (ours)      	& \multicolumn{2}{c}{99.40$\pm$0.55 ($1.80\uparrow$)} 			& \multicolumn{2}{c}{91.87$\pm$1.90 ($3.86\uparrow$)} 			& \multicolumn{2}{c}{\textbf{80.38$\pm$2.11} ($6.02\uparrow$)}			& \multicolumn{2}{c}{75.49$\pm$1.99 ($11.96\uparrow$)}	 \\
    \hdashline
    {\elimipl}      	& \multicolumn{2}{c}{99.20$\pm$0.65} 			& \multicolumn{2}{c}{90.27$\pm$1.84} 			& \multicolumn{2}{c}{77.13$\pm$1.80}			& \multicolumn{2}{c}{67.49$\pm$2.18}	 \\
    {\samcc} (ours)      	& \multicolumn{2}{c}{99.80$\pm$0.43 ($0.60\uparrow$)} 			& \multicolumn{2}{c}{90.27$\pm$1.98 ($0.00\uparrow$)} 			& \multicolumn{2}{c}{79.21$\pm$2.26 ($2.08\uparrow$)}			& \multicolumn{2}{c}{71.96$\pm$2.34 ($4.47\uparrow$)}	 \\
    {\samcn} (ours)      	& \multicolumn{2}{c}{99.73$\pm$0.44 ($0.53\uparrow$)} 			& \multicolumn{2}{c}{90.33$\pm$2.09 ($0.06\uparrow$)} 			& \multicolumn{2}{c}{79.54$\pm$1.97 ($2.41\uparrow$)}			& \multicolumn{2}{c}{73.64$\pm$1.75 ($6.15\uparrow$)}	 \\
    \hdashline
    {\miplma}      	& \multicolumn{2}{c}{98.47$\pm$1.03} 			& \multicolumn{2}{c}{91.53$\pm$1.61} 			& \multicolumn{2}{c}{77.56$\pm$2.05}			& \multicolumn{2}{c}{70.33$\pm$2.63}	 \\
    {\mamcc} (ours)      	& \multicolumn{2}{c}{\textbf{99.93$\pm$0.20} ($1.46\uparrow$)} 			& \multicolumn{2}{c}{92.73$\pm$1.21 ($1.20\uparrow$)} 			& \multicolumn{2}{c}{79.54$\pm$1.59 ($1.98\uparrow$)}			& \multicolumn{2}{c}{72.40$\pm$1.86 ($2.07\uparrow$)}	 \\
    {\mamcn} (ours)          	& \multicolumn{2}{c}{\textbf{99.93$\pm$0.20} ($1.46\uparrow$)} 			& \multicolumn{2}{c}{\textbf{93.20$\pm$1.33} ($1.67\uparrow$)} 			& \multicolumn{2}{c}{79.95$\pm$1.23 ($2.39\uparrow$)}			& \multicolumn{2}{c}{74.40$\pm$2.21 ($4.07\uparrow$)}	 \\
    \hline
    \multicolumn{9}{c}{$r=2$} \\
    \hline
    
    {\miplgp}      		& \multicolumn{2}{c}{81.73$\pm$3.01} 		& \multicolumn{2}{c}{79.07$\pm$2.74}		& \multicolumn{2}{c}{67.18$\pm$1.50} 	& \multicolumn{2}{c}{61.31$\pm$2.61} \\
    {\promipl}   		& \multicolumn{2}{c}{99.86$\pm$0.24} 		& \multicolumn{2}{c}{88.83$\pm$2.26}		& \multicolumn{2}{c}{71.84$\pm$1.96} 	& \multicolumn{2}{c}{63.35$\pm$2.32} \\
    {\fastmipl}    		& \multicolumn{2}{c}{99.77$\pm$0.41} 		& \multicolumn{2}{c}{\textbf{90.15$\pm$2.53}}		& \multicolumn{2}{c}{\textbf{78.91$\pm$2.29}} 	& \multicolumn{2}{c}{70.72$\pm$2.67} \\
    \hdashline
    {\demipl}         	& \multicolumn{2}{c}{94.27$\pm$2.72} 		& \multicolumn{2}{c}{82.33$\pm$2.85}	& \multicolumn{2}{c}{70.13$\pm$2.40}		& \multicolumn{2}{c}{55.40$\pm$5.09}	 \\
    {\damcc} (ours)   		& \multicolumn{2}{c}{99.40$\pm$0.81 ($5.13\uparrow$)}				& \multicolumn{2}{c}{89.33$\pm$2.88 ($7.00\uparrow$)} 			& \multicolumn{2}{c}{77.03$\pm$1.64 ($6.90\uparrow$)} 			& \multicolumn{2}{c}{70.24$\pm$2.39 ($14.84\uparrow$)} \\
    {\damcn} (ours)   		& \multicolumn{2}{c}{99.53$\pm$0.60 ($5.26\uparrow$)}				& \multicolumn{2}{c}{89.40$\pm$2.69 ($7.07\uparrow$)} 			& \multicolumn{2}{c}{78.00$\pm$1.54 ($7.87\uparrow$)} 			& \multicolumn{2}{c}{\textbf{71.93$\pm$1.82} ($16.53\uparrow$)} \\
    \hdashline
    {\elimipl}    		& \multicolumn{2}{c}{98.67$\pm$0.99}				& \multicolumn{2}{c}{84.53$\pm$2.56} 			& \multicolumn{2}{c}{74.46$\pm$1.53} 			& \multicolumn{2}{c}{61.58$\pm$2.54} \\			   
    {\samcc} (ours)   		& \multicolumn{2}{c}{98.80$\pm$0.83 ($0.13\uparrow$)}				& \multicolumn{2}{c}{85.33$\pm$3.24 ($0.80\uparrow$)} 			& \multicolumn{2}{c}{77.21$\pm$2.03 ($2.75\uparrow$)} 			& \multicolumn{2}{c}{66.02$\pm$2.79 ($4.44\uparrow$)} \\  
    {\samcn} (ours)   		& \multicolumn{2}{c}{98.93$\pm$0.85 ($0.26\uparrow$)}				& \multicolumn{2}{c}{85.40$\pm$2.88 ($0.87\uparrow$)} 			& \multicolumn{2}{c}{77.54$\pm$2.20 ($3.08\uparrow$)} 			& \multicolumn{2}{c}{66.87$\pm$2.32 ($5.29\uparrow$)} \\
    \hdashline
    {\miplma}    		& \multicolumn{2}{c}{97.87$\pm$1.39}				& \multicolumn{2}{c}{86.67$\pm$2.76} 			& \multicolumn{2}{c}{76.15$\pm$1.55} 			& \multicolumn{2}{c}{66.82$\pm$3.10} \\
    {\mamcc} (ours)   		& \multicolumn{2}{c}{99.80$\pm$0.31 ($1.93\uparrow$)}				& \multicolumn{2}{c}{89.47$\pm$1.86 ($2.80\uparrow$)} 			& \multicolumn{2}{c}{77.36$\pm$2.13 ($1.21\uparrow$)} 			& \multicolumn{2}{c}{69.91$\pm$2.34 ($3.09\uparrow$)} \\
    {\mamcn} (ours)       		& \multicolumn{2}{c}{\textbf{99.87$\pm$0.27} ($2.00\uparrow$)}				& \multicolumn{2}{c}{90.07$\pm$1.50 ($3.40\uparrow$)} 			& \multicolumn{2}{c}{78.46$\pm$1.92 ($2.31\uparrow$)}			& \multicolumn{2}{c}{71.42$\pm$1.75 ($4.60\uparrow$)} \\
    
    \hline
    \multicolumn{9}{c}{$r=3$} \\
    \hline
    {\miplgp}     		& \multicolumn{2}{c}{62.13$\pm$6.39} 		& \multicolumn{2}{c}{67.00$\pm$5.24} 	  & \multicolumn{2}{c}{62.51$\pm$1.47}		& \multicolumn{2}{c}{56.93$\pm$3.21} \\
    {\promipl}   		& \multicolumn{2}{c}{78.32$\pm$11.64} 		& \multicolumn{2}{c}{65.89$\pm$4.14} 	& \multicolumn{2}{c}{69.33$\pm$2.11}		& \multicolumn{2}{c}{53.88$\pm$2.41} \\
    {\fastmipl}	    	& \multicolumn{2}{c}{97.43$\pm$7.37} 		& \multicolumn{2}{c}{\textbf{81.60$\pm$7.01}} 	& \multicolumn{2}{c}{77.30$\pm$2.24}		& \multicolumn{2}{c}{61.49$\pm$3.48} \\
    \hdashline
    {\demipl}           & \multicolumn{2}{c}{70.93$\pm$8.83}		& \multicolumn{2}{c}{65.73$\pm$2.48} 	& \multicolumn{2}{c}{69.62$\pm$2.39}		& \multicolumn{2}{c}{50.31$\pm$1.84}	 \\
    {\damcc} (ours)      	& \multicolumn{2}{c}{85.27$\pm$9.06 ($14.37\uparrow$)} 			& \multicolumn{2}{c}{72.87$\pm$7.20 ($7.14\uparrow$)}				& \multicolumn{2}{c}{76.90$\pm$1.73 ($7.28\uparrow$)} 		& \multicolumn{2}{c}{66.76$\pm$2.40 ($16.45\uparrow$)} \\						
    {\damcn} (ours)      	& \multicolumn{2}{c}{78.73$\pm$15.67 ($7.80\uparrow$)} 			& \multicolumn{2}{c}{74.60$\pm$4.63 ($8.87\uparrow$)}				& \multicolumn{2}{c}{77.69$\pm$1.75 ($8.07\uparrow$)} 		& \multicolumn{2}{c}{68.89$\pm$1.94 ($18.58\uparrow$)} \\
    \hdashline
    {\elimipl}	         & \multicolumn{2}{c}{74.80$\pm$14.41} 			& \multicolumn{2}{c}{70.20$\pm$5.54}				& \multicolumn{2}{c}{71.67$\pm$1.67} 		& \multicolumn{2}{c}{60.02$\pm$2.89} \\ 
    {\samcc} (ours)      	& \multicolumn{2}{c}{88.40$\pm$9.09 ($13.60\uparrow$)} 			& \multicolumn{2}{c}{72.53$\pm$7.86 ($2.33\uparrow$)}				& \multicolumn{2}{c}{75.13$\pm$2.17 ($3.46\uparrow$)} 		& \multicolumn{2}{c}{61.62$\pm$1.93 ($1.60\uparrow$)} \\
    {\samcn} (ours)      	& \multicolumn{2}{c}{86.13$\pm$9.48 ($11.33\uparrow$)} 			& \multicolumn{2}{c}{74.27$\pm$4.09 ($4.07\uparrow$)}				& \multicolumn{2}{c}{77.05$\pm$1.46 ($5.38\uparrow$)} 		& \multicolumn{2}{c}{63.16$\pm$1.94 ($3.14\uparrow$)} \\
    \hdashline
    {\miplma}       & \multicolumn{2}{c}{74.93$\pm$10.32} 			& \multicolumn{2}{c}{65.40$\pm$5.53}				& \multicolumn{2}{c}{74.56$\pm$1.29} 		& \multicolumn{2}{c}{62.73$\pm$2.38} \\				
    {\mamcc} (ours)      	& \multicolumn{2}{c}{95.07$\pm$7.26 ($20.14\uparrow$)} 			& \multicolumn{2}{c}{74.00$\pm$7.57 ($8.60\uparrow$)}				& \multicolumn{2}{c}{77.18$\pm$1.57 ($2.62\uparrow$)} 		& \multicolumn{2}{c}{67.20$\pm$2.73 ($4.47\uparrow$)} \\
    {\mamcn} (ours)          	& \multicolumn{2}{c}{\textbf{98.00$\pm$1.49} ($23.07\uparrow$)} 			& \multicolumn{2}{c}{77.73$\pm$4.95 ($12.33\uparrow$)}				& \multicolumn{2}{c}{\textbf{77.82$\pm$1.32} ($3.26\uparrow$)} 		& \multicolumn{2}{c}{\textbf{69.13$\pm$1.47} ($6.40\uparrow$)} \\
     \hline \hline
 \end{tabular}
\end{table*}

\begin{table*}[!t]
    \centering  
    \caption{Classification accuracy (mean$\pm$std\%) on the real-world datasets. -- indicates computational constraints. \looseness=-1}
    \label{tab:res_crc}
    \begin{tabular}{p{1.2cm} cccccccccccccc}
    \hline  \hline
    \multicolumn{1}{l}{}   & \multicolumn{2}{c}{C-Row}           & \multicolumn{2}{c}{C-SBN}           & \multicolumn{2}{c}{C-KMeans}          & \multicolumn{2}{c}{C-SIFT}      & \multicolumn{2}{c}{C-R34-9}   & \multicolumn{2}{c}{C-R34-16}    & \multicolumn{2}{c}{C-R34-25} \\
    \hline
    \multicolumn{1}{l}{{\miplgp}}   & \multicolumn{2}{c}{43.13$\pm$0.57}      & \multicolumn{2}{c}{33.49$\pm$0.63}      & \multicolumn{2}{c}{32.90$\pm$1.25}        & \multicolumn{2}{c}{--}  & \multicolumn{2}{c}{--}    & \multicolumn{2}{c}{--}    & \multicolumn{2}{c}{--}\\ 
    \multicolumn{1}{l}{{\promipl}}    & \multicolumn{2}{c}{43.51$\pm$0.93}      & \multicolumn{2}{c}{51.56$\pm$1.23}      & \multicolumn{2}{c}{56.52$\pm$1.30}        & \multicolumn{2}{c}{56.25$\pm$1.11}  & \multicolumn{2}{c}{60.62$\pm$1.02}    & \multicolumn{2}{c}{64.63$\pm$0.75}    & \multicolumn{2}{c}{67.24$\pm$0.80} \\
    \multicolumn{1}{l}{{\fastmipl}}   & \multicolumn{2}{c}{48.68$\pm$3.81}      & \multicolumn{2}{c}{57.21$\pm$3.08}      & \multicolumn{2}{c}{57.30$\pm$1.28}        & \multicolumn{2}{c}{52.55$\pm$2.95}  & \multicolumn{2}{c}{56.78$\pm$1.61}    & \multicolumn{2}{c}{61.85$\pm$1.48}    & \multicolumn{2}{c}{64.24$\pm$1.64}\\ 
    \hdashline
    \multicolumn{1}{l}{{\demipl}}   & \multicolumn{2}{c}{40.78$\pm$1.01}      & \multicolumn{2}{c}{48.58$\pm$1.40}        & \multicolumn{2}{c}{52.11$\pm$1.20}      & \multicolumn{2}{c}{53.17$\pm$1.27}  & \multicolumn{2}{c}{58.84$\pm$1.40}    & \multicolumn{2}{c}{62.34$\pm$0.82}    & \multicolumn{2}{c}{65.19$\pm$0.85}\\
    \multicolumn{1}{l}{\multirow{2}[0]{*}{{\damcc} (ours)}} & \multicolumn{2}{c}{48.10$\pm$0.60}      & \multicolumn{2}{c}{56.54$\pm$0.84}      & \multicolumn{2}{c}{62.78$\pm$1.30}      & \multicolumn{2}{c}{53.59$\pm$1.05}  & \multicolumn{2}{c}{62.13$\pm$1.21}    & \multicolumn{2}{c}{63.99$\pm$0.87}    & \multicolumn{2}{c}{65.34$\pm$0.74}\\
        & \multicolumn{2}{c}{($7.32\uparrow$)}      & \multicolumn{2}{c}{($7.96\uparrow$)}      & \multicolumn{2}{c}{($10.67\uparrow$)}     & \multicolumn{2}{c}{($0.42\uparrow$)}  & \multicolumn{2}{c}{($3.29\uparrow$)}    & \multicolumn{2}{c}{($1.65\uparrow$)}    & \multicolumn{2}{c}{($0.15\uparrow$)}  \\
    \multicolumn{1}{l}{\multirow{2}[0]{*}{{\damcn} (ours)}} & \multicolumn{2}{c}{\textbf{49.06$\pm$0.71}}       & \multicolumn{2}{c}{\textbf{57.91$\pm$0.60}}       & \multicolumn{2}{c}{\textbf{64.78$\pm$1.02}}     & \multicolumn{2}{c}{55.34$\pm$1.05}  & \multicolumn{2}{c}{63.74$\pm$0.94}    & \multicolumn{2}{c}{64.97$\pm$1.00}    & \multicolumn{2}{c}{66.64$\pm$0.85}\\
        & \multicolumn{2}{c}{($8.28\uparrow$)}      & \multicolumn{2}{c}{($9.33\uparrow$)}      & \multicolumn{2}{c}{($12.67\uparrow$)}     & \multicolumn{2}{c}{($2.17\uparrow$)}  & \multicolumn{2}{c}{($4.90\uparrow$)}    & \multicolumn{2}{c}{($2.63\uparrow$)}    & \multicolumn{2}{c}{($1.45\uparrow$)}  \\
    \hdashline    
    \multicolumn{1}{l}{{\elimipl}}    & \multicolumn{2}{c}{43.26$\pm$0.82}      & \multicolumn{2}{c}{50.90$\pm$0.79}      & \multicolumn{2}{c}{54.58$\pm$1.18}      & \multicolumn{2}{c}{54.05$\pm$1.02}  & \multicolumn{2}{c}{60.97$\pm$1.22}    & \multicolumn{2}{c}{63.08$\pm$0.66}    & \multicolumn{2}{c}{66.50$\pm$0.67}\\    
    \multicolumn{1}{l}{\multirow{2}[0]{*}{{\samcc} (ours)}} & \multicolumn{2}{c}{46.58$\pm$0.57}      & \multicolumn{2}{c}{55.91$\pm$0.95}      & \multicolumn{2}{c}{62.04$\pm$1.78}      & \multicolumn{2}{c}{\textbf{57.20$\pm$0.94}}   & \multicolumn{2}{c}{63.95$\pm$0.98}    & \multicolumn{2}{c}{67.41$\pm$0.70}    & \multicolumn{2}{c}{69.43$\pm$0.94}  \\
        & \multicolumn{2}{c}{($3.32\uparrow$)}      & \multicolumn{2}{c}{($5.01\uparrow$)}      & \multicolumn{2}{c}{($7.46\uparrow$)}      & \multicolumn{2}{c}{($3.15\uparrow$)}  & \multicolumn{2}{c}{($2.98\uparrow$)}    & \multicolumn{2}{c}{($4.33\uparrow$)}    & \multicolumn{2}{c}{($2.93\uparrow$)}  \\    
    \multicolumn{1}{l}{\multirow{2}[0]{*}{{\samcn} (ours)}} & \multicolumn{2}{c}{47.50$\pm$0.54}      & \multicolumn{2}{c}{57.24$\pm$0.90}      & \multicolumn{2}{c}{63.84$\pm$1.01}      & \multicolumn{2}{c}{57.10$\pm$0.76}    & \multicolumn{2}{c}{64.58$\pm$1.07}    & \multicolumn{2}{c}{67.81$\pm$0.72}    & \multicolumn{2}{c}{69.75$\pm$0.83}  \\
        & \multicolumn{2}{c}{($4.24\uparrow$)}      & \multicolumn{2}{c}{($6.34\uparrow$)}      & \multicolumn{2}{c}{($9.26\uparrow$)}      & \multicolumn{2}{c}{($3.05\uparrow$)}  & \multicolumn{2}{c}{($3.61\uparrow$)}    & \multicolumn{2}{c}{($4.73\uparrow$)}    & \multicolumn{2}{c}{($3.25\uparrow$)}  \\
    \hdashline   
    \multicolumn{1}{l}{{\miplma}}   & \multicolumn{2}{c}{44.37$\pm$0.99}      & \multicolumn{2}{c}{52.46$\pm$0.70}      & \multicolumn{2}{c}{55.73$\pm$1.01}       & \multicolumn{2}{c}{55.29$\pm$0.94}  & \multicolumn{2}{c}{59.38$\pm$1.24}    & \multicolumn{2}{c}{63.12$\pm$0.77}    & \multicolumn{2}{c}{68.70$\pm$1.08}\\
    \multicolumn{1}{l}{\multirow{2}[0]{*}{{\mamcc} (ours)}} & \multicolumn{2}{c}{47.66$\pm$0.57}      & \multicolumn{2}{c}{56.83$\pm$1.06}      & \multicolumn{2}{c}{57.60$\pm$1.28}      & \multicolumn{2}{c}{56.97$\pm$0.97}  & \multicolumn{2}{c}{\textbf{64.74$\pm$0.98}}   & \multicolumn{2}{c}{\textbf{67.88$\pm$0.85}}   & \multicolumn{2}{c}{\textbf{69.83$\pm$0.96}} \\
        & \multicolumn{2}{c}{($3.29\uparrow$)}      & \multicolumn{2}{c}{($4.37\uparrow$)}      & \multicolumn{2}{c}{($1.87\uparrow$)}      & \multicolumn{2}{c}{($1.68\uparrow$)}  & \multicolumn{2}{c}{($5.36\uparrow$)}    & \multicolumn{2}{c}{($4.76\uparrow$)}    & \multicolumn{2}{c}{($1.13\uparrow$)}  \\    
    \multicolumn{1}{l}{\multirow{2}[0]{*}{{\mamcn} (ours)}} & \multicolumn{2}{c}{47.99$\pm$0.50}      & \multicolumn{2}{c}{57.40$\pm$0.94}      & \multicolumn{2}{c}{59.02$\pm$0.96}      & \multicolumn{2}{c}{57.07$\pm$0.86}    & \multicolumn{2}{c}{64.73$\pm$1.02}    & \multicolumn{2}{c}{67.72$\pm$0.78}    & \multicolumn{2}{c}{69.81$\pm$1.00}  \\
        & \multicolumn{2}{c}{($3.62\uparrow$)}      & \multicolumn{2}{c}{($4.94\uparrow$)}      & \multicolumn{2}{c}{($3.29\uparrow$)}      & \multicolumn{2}{c}{($1.78\uparrow$)}  & \multicolumn{2}{c}{($5.35\uparrow$)}    & \multicolumn{2}{c}{($4.60\uparrow$)}    & \multicolumn{2}{c}{($1.11\uparrow$)}  \\
    \hline  \hline  
\end{tabular}
\end{table*}

\subsection{Classification Performance}
\label{subsec:res_classification}
\subsubsection{Accuracy on the Benchmark Datasets}
\label{subsec:res_benchmark}
Table \ref{tab:res_benchmark} demonstrates that our methods outperform the comparative methods in $67$ of the $72$ cases. In $12$ of the $72$ cases, the improvement exceeds $10\%$. {\fastmipl} outperforms our models in $4$ of the $72$ cases. We speculate that {\fastmipl} benefits from training all samples in a single batch, leveraging greater computational resources.

For MNIST-{\scriptsize{MIPL}}, the potential for accuracy improvement is limited for $r=1$ and $2$. However, for $r=3$, {\damcc} and {\samcc} achieve significant accuracy improvements of $14.37\%$ and $13.60\%$, respectively. Notably, {\mamcc} and {\mamcn} surpass {\miplma} by $20.14\%$ and $23.07\%$ on MNIST-{\scriptsize{MIPL}} with $r=3$. On the FMNIST-{\scriptsize{MIPL}} and Birdsong-{\scriptsize{MIPL}} datasets, {\damcn} achieves improvements exceeding $8\%$. Furthermore, on SIVAL-{\scriptsize{MIPL}}, both {\damcc} and {\damcn} achieve gains exceeding $10\%$ for $r \in \{1,2,3\}$, with a peak improvement of $18.58\%$. As $r$ increases from $1$ to $3$, the mean accuracy improvement across benchmark datasets is $3.27\%$, $4.57\%$, and $8.93\%$. These results indicate that CDL effectively handles more challenging scenarios. \looseness=-1

\begin{table*}[!t]
  \centering  
     \caption{Expected calibration error (mean$\pm$std\%) on the benchmark datasets with varying numbers of false-positive labels. \looseness=-1}
\label{tab:calibration_benchmark}
   \begin{tabular}{lcccccccc}
    \hline  \hline
    \multicolumn{1}{l}{}  & \multicolumn{2}{c}{MNIST-{\tiny{MIPL}}}           & \multicolumn{2}{c}{FMNIST-{\tiny{MIPL}}}        & \multicolumn{2}{c}{Birdsong-{\tiny{MIPL}}}        & \multicolumn{2}{c}{SIVAL-{\tiny{MIPL}}} \\
    \hline
    \multicolumn{9}{c}{$r=1$} \\
    \hline 
    {\promipl}          & \multicolumn{2}{c}{59.62$\pm$0.16}    & \multicolumn{2}{c}{53.56$\pm$2.20}      & \multicolumn{2}{c}{60.77$\pm$1.52}    & \multicolumn{2}{c}{59.46$\pm$3.12}   \\
    {\fastmipl}           & \multicolumn{2}{c}{3.47$\pm$0.54}   & \multicolumn{2}{c}{5.53$\pm$1.35}      & \multicolumn{2}{c}{7.15$\pm$1.72}     & \multicolumn{2}{c}{\textbf{10.90$\pm$2.11}}   \\
    \hdashline
    {\demipl}           & \multicolumn{2}{c}{59.40$\pm$0.77}    & \multicolumn{2}{c}{51.50$\pm$1.54}      & \multicolumn{2}{c}{57.86$\pm$1.16}    & \multicolumn{2}{c}{56.89$\pm$3.01}   \\
    {\damcc} (ours)            & \multicolumn{2}{c}{1.08$\pm$0.41 ($58.32\downarrow$)}       & \multicolumn{2}{c}{5.09$\pm$1.05 ($46.41\downarrow$)}      & \multicolumn{2}{c}{4.68$\pm$1.10 ($53.18\downarrow$)}      & \multicolumn{2}{c}{12.74$\pm$1.63 ($44.15\downarrow$)}  \\
    {\damcn} (ours)            & \multicolumn{2}{c}{1.11$\pm$0.38 ($58.29\downarrow$)}       & \multicolumn{2}{c}{5.44$\pm$1.49 ($46.06\downarrow$)}       & \multicolumn{2}{c}{7.53$\pm$1.63 ($50.33\downarrow$)}     & \multicolumn{2}{c}{16.25$\pm$1.71 ($40.64\downarrow$)}   \\
    \hdashline
    {\elimipl}         & \multicolumn{2}{c}{59.26$\pm$0.38}      & \multicolumn{2}{c}{51.55$\pm$1.44}      & \multicolumn{2}{c}{60.16$\pm$1.83}      & \multicolumn{2}{c}{58.83$\pm$2.15}   \\
    {\samcc} (ours)            & \multicolumn{2}{c}{1.45$\pm$0.38 ($57.81\downarrow$)}       & \multicolumn{2}{c}{5.60$\pm$1.64 ($45.95\downarrow$)}       & \multicolumn{2}{c}{5.85$\pm$1.43 ($54.31\downarrow$)}     & \multicolumn{2}{c}{12.85$\pm$1.52 ($45.98\downarrow$)}   \\
    {\samcn} (ours)            & \multicolumn{2}{c}{1.34$\pm$0.35 ($57.92\downarrow$)}       & \multicolumn{2}{c}{5.75$\pm$1.04 ($45.80\downarrow$)}       & \multicolumn{2}{c}{6.32$\pm$2.26 ($53.84\downarrow$)}     & \multicolumn{2}{c}{16.75$\pm$1.49 ($42.08\downarrow$)}   \\
    \hdashline
    {\miplma}        & \multicolumn{2}{c}{58.65$\pm$0.85}      & \multicolumn{2}{c}{52.38$\pm$1.60}      & \multicolumn{2}{c}{60.89$\pm$2.04}      & \multicolumn{2}{c}{61.67$\pm$2.58}   \\
    {\mamcc} (ours)            & \multicolumn{2}{c}{1.00$\pm$0.15 ($57.65\downarrow$)}       & \multicolumn{2}{c}{4.84$\pm$1.43 ($47.54\downarrow$)}       & \multicolumn{2}{c}{\textbf{4.53$\pm$0.68} ($56.36\downarrow$)}     & \multicolumn{2}{c}{12.15$\pm$1.02 ($49.52\downarrow$)}   \\
    {\mamcn} (ours)            & \multicolumn{2}{c}{\textbf{0.94$\pm$0.17} ($57.71\downarrow$)}      & \multicolumn{2}{c}{\textbf{4.18$\pm$1.57} ($48.20\downarrow$)}       & \multicolumn{2}{c}{6.87$\pm$1.75 ($54.02\downarrow$)}     & \multicolumn{2}{c}{15.91$\pm$1.38 ($45.76\downarrow$)}   \\

    \hline
    \multicolumn{9}{c}{$r=2$} \\
    \hline
    {\promipl}        & \multicolumn{2}{c}{59.64$\pm$0.20}  & \multicolumn{2}{c}{50.57$\pm$1.96}        & \multicolumn{2}{c}{55.76$\pm$1.76}    & \multicolumn{2}{c}{54.83$\pm$2.21}   \\
    {\fastmipl}       & \multicolumn{2}{c}{11.05$\pm$1.31}  & \multicolumn{2}{c}{10.02$\pm$1.98}        & \multicolumn{2}{c}{5.34$\pm$1.19}   & \multicolumn{2}{c}{11.395$\pm$2.07}  \\
    \hdashline
    {\demipl}        & \multicolumn{2}{c}{59.62$\pm$1.09}  & \multicolumn{2}{c}{48.84$\pm$2.18}        & \multicolumn{2}{c}{51.65$\pm$1.63}    & \multicolumn{2}{c}{51.55$\pm$3.68}   \\
    {\damcc} (ours)          & \multicolumn{2}{c}{1.20$\pm$0.58 ($58.42\downarrow$)}       & \multicolumn{2}{c}{6.62$\pm$2.04 ($42.22\downarrow$)}       & \multicolumn{2}{c}{5.23$\pm$0.74 ($46.42\downarrow$)}       & \multicolumn{2}{c}{\textbf{10.32$\pm$1.20} ($41.23\downarrow$)} \\
    {\damcn} (ours)          & \multicolumn{2}{c}{\textbf{0.86$\pm$0.42} ($58.76\downarrow$)}        & \multicolumn{2}{c}{6.33$\pm$2.41 ($42.51\downarrow$)}      & \multicolumn{2}{c}{5.54$\pm$1.38 ($46.11\downarrow$)}       & \multicolumn{2}{c}{14.45$\pm$1.63 ($37.10\downarrow$)} \\
    \hdashline
    {\elimipl}       & \multicolumn{2}{c}{60.02$\pm$1.04}        & \multicolumn{2}{c}{48.89$\pm$2.55}      & \multicolumn{2}{c}{57.77$\pm$1.56}      & \multicolumn{2}{c}{53.18$\pm$2.49} \\
    {\samcc} (ours)          & \multicolumn{2}{c}{3.13$\pm$0.51 ($56.89\downarrow$)}       & \multicolumn{2}{c}{7.53$\pm$1.35 ($41.36\downarrow$)}       & \multicolumn{2}{c}{\textbf{4.98$\pm$0.83} ($52.79\downarrow$)}      & \multicolumn{2}{c}{10.66$\pm$1.81 ($42.52\downarrow$)} \\
    {\samcn} (ours)          & \multicolumn{2}{c}{2.72$\pm$0.32 ($57.30\downarrow$)}       & \multicolumn{2}{c}{6.93$\pm$1.74 ($41.96\downarrow$)}       & \multicolumn{2}{c}{6.70$\pm$1.21 ($51.07\downarrow$)}       & \multicolumn{2}{c}{16.53$\pm$1.97 ($36.65\downarrow$)} \\
    \hdashline
    {\miplma}       & \multicolumn{2}{c}{59.31$\pm$1.11}        & \multicolumn{2}{c}{48.54$\pm$3.00}      & \multicolumn{2}{c}{59.50$\pm$1.61}      & \multicolumn{2}{c}{58.31$\pm$3.07} \\
    {\mamcc} (ours)          & \multicolumn{2}{c}{1.20$\pm$0.26 ($58.11\downarrow$)}       & \multicolumn{2}{c}{\textbf{6.26$\pm$1.30} ($42.28\downarrow$)}       & \multicolumn{2}{c}{5.01$\pm$0.73 ($54.49\downarrow$)}      & \multicolumn{2}{c}{10.60$\pm$1.54 ($47.71\downarrow$)} \\
    {\mamcn} (ours)          & \multicolumn{2}{c}{1.16$\pm$0.26 ($58.15\downarrow$)}       & \multicolumn{2}{c}{6.46$\pm$1.71 ($42.08\downarrow$)}       & \multicolumn{2}{c}{5.71$\pm$0.97 ($53.79\downarrow$)}       & \multicolumn{2}{c}{14.38$\pm$1.59 ($43.93\downarrow$)} \\

    \hline
    \multicolumn{9}{c}{$r=3$} \\
    \hline
     {\promipl}         & \multicolumn{2}{c}{46.41$\pm$11.20}   & \multicolumn{2}{c}{35.26$\pm$4.25}      & \multicolumn{2}{c}{54.40$\pm$2.22}      & \multicolumn{2}{c}{45.89$\pm$2.26}   \\
    {\fastmipl}         & \multicolumn{2}{c}{21.46$\pm$3.29}    & \multicolumn{2}{c}{\textbf{16.92$\pm$4.35}}      & \multicolumn{2}{c}{6.77$\pm$1.24}     & \multicolumn{2}{c}{13.91$\pm$2.48}   \\    
    \hdashline
    {\demipl}       & \multicolumn{2}{c}{44.55$\pm$12.76}   & \multicolumn{2}{c}{40.04$\pm$3.27}      & \multicolumn{2}{c}{53.42$\pm$1.19}      & \multicolumn{2}{c}{46.26$\pm$2.70}   \\  
    {\damcc} (ours)            & \multicolumn{2}{c}{12.99$\pm$7.43 ($31.56\downarrow$)}      & \multicolumn{2}{c}{23.28$\pm$6.11 ($16.76\downarrow$)}        & \multicolumn{2}{c}{5.63$\pm$0.88 ($47.79\downarrow$)}     & \multicolumn{2}{c}{9.48$\pm$1.60 ($36.78\downarrow$)} \\       
    {\damcn} (ours)            & \multicolumn{2}{c}{17.99$\pm$13.08 ($26.56\downarrow$)}       & \multicolumn{2}{c}{22.16$\pm$4.02 ($17.88\downarrow$)}        & \multicolumn{2}{c}{\textbf{4.52$\pm$0.87} ($48.90\downarrow$)}    & \multicolumn{2}{c}{15.09$\pm$2.06 ($31.17\downarrow$)} \\   
    \hdashline
    {\elimipl}      & \multicolumn{2}{c}{43.59$\pm$15.00}       & \multicolumn{2}{c}{40.65$\pm$5.38}        & \multicolumn{2}{c}{55.31$\pm$1.64}    & \multicolumn{2}{c}{51.63$\pm$2.85} \\
    {\samcc} (ours)            & \multicolumn{2}{c}{9.92$\pm$7.18 ($33.67\downarrow$)}       & \multicolumn{2}{c}{22.70$\pm$7.34 ($17.95\downarrow$)}        & \multicolumn{2}{c}{5.03$\pm$0.89 ($50.28\downarrow$)}     & \multicolumn{2}{c}{9.05$\pm$1.45 ($42.58\downarrow$)} \\
    {\samcn} (ours)            & \multicolumn{2}{c}{11.24$\pm$7.52 ($32.35\downarrow$)}      & \multicolumn{2}{c}{20.89$\pm$4.20 ($19.76\downarrow$)}       & \multicolumn{2}{c}{4.87$\pm$1.04 ($50.44\downarrow$)}     & \multicolumn{2}{c}{15.35$\pm$2.51 ($36.28\downarrow$)} \\   
    \hdashline
    {\miplma}       & \multicolumn{2}{c}{44.90$\pm$10.56}       & \multicolumn{2}{c}{38.04$\pm$5.32}        & \multicolumn{2}{c}{58.02$\pm$1.22}    & \multicolumn{2}{c}{54.35$\pm$2.33} \\   
    {\mamcc} (ours)            & \multicolumn{2}{c}{4.55$\pm$5.62 ($40.35\downarrow$)}       & \multicolumn{2}{c}{21.33$\pm$7.25 ($16.71\downarrow$)}        & \multicolumn{2}{c}{5.55$\pm$1.12 ($52.47\downarrow$)}     & \multicolumn{2}{c}{\textbf{8.88$\pm$1.45} ($45.47\downarrow$)} \\
    {\mamcn} (ours)            & \multicolumn{2}{c}{\textbf{1.72$\pm$0.92} ($43.18\downarrow$)}      & \multicolumn{2}{c}{18.90$\pm$4.59 ($19.14\downarrow$)}       & \multicolumn{2}{c}{5.34$\pm$1.32 ($52.68\downarrow$)}     & \multicolumn{2}{c}{14.02$\pm$1.42 ($40.33\downarrow$)} \\
     \hline \hline
    \end{tabular}
\end{table*}

\begin{figure*}[!t]
    \centering
    \begin{overpic}[width=18cm]{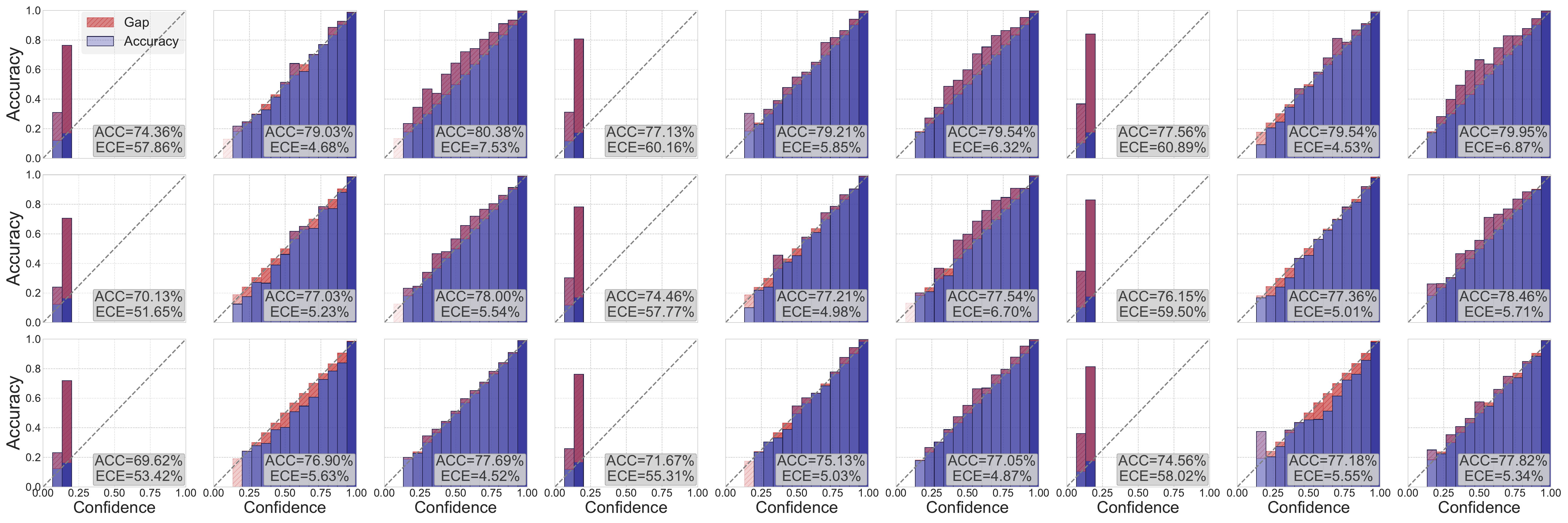}
        \put(-10, 51) {\rotatebox{90}{\small $r=3$}} 
	\put(-10, 154) {\rotatebox{90}{\small $r=2$}} 
	\put(-10, 260) {\rotatebox{90}{\small $r=1$}}
	\put(40, -10){\small {\demipl}}
	\put(150, -10){\small {\damcc}}
	\put(260, -10){\small {\damcn}}
	\put(370, -10){\small {\elimipl}}
	\put(480, -10){\small {\samcc}}
	\put(590, -10){\small {\samcn}}
        \put(700, -10){\small {\miplma}}
	\put(810, -10){\small {\mamcc}}
	\put(920, -10){\small {\mamcn}}
    \end{overpic}
    \caption{Reliability diagrams of {\demipl} \cite{tang2023demipl}, {\elimipl} \cite{tang2024elimipl}, and {\miplma} \cite{tang2024miplma}, and our methods on the Birdsong-{\scriptsize{MIPL}} dataset with varying numbers of false-positive labels. The bar color intensity indicates that more samples are assigned to the corresponding confidence intervals.}
 \label{fig:ece_birdsong}
\end{figure*}

\subsubsection{Accuracy on the Real-World Datasets}
\label{subsec:res_real}
The experimental results in Table \ref{tab:res_crc} demonstrate that our methods consistently achieve superior classification accuracy compared to all baseline methods. Some {\miplgp} results are marked as ``--" due to computational limitations, which hinder evaluation on certain datasets. This limitation indicates that {\miplgp} may struggle to integrate with deep learning-based feature representations.

The integration of CDL with {\dam}, {\sam}, or {\mam} consistently improves classification performance over {\demipl}, {\elimipl}, and {\miplma}. On the C-KMeans dataset, {\damcc} and {\damcn} achieve significant accuracy improvements of $10.67\%$ and $12.67\%$, respectively. For image bag generators such as Row, SBN, and KMeans, {\dam} outperforms {\sam} and {\mam} in classification accuracy. In contrast, for instance-level feature learning with SIFT or ResNet-34, {\sam} and {\mam} outperform {\dam}. 
Furthermore, in the three CRC-{\scriptsize{MIPL}} datasets utilizing ResNet-34, all MIPL approaches exhibit improved performance as the number of partitioned patches increases. This trend indicates that in CRC-{\scriptsize{MIPL}} datasets, leveraging deep learning-based features contributes to enhancing classification performance. 

Additionally, {\dam} performs well on simple feature representations in the CRC-{\scriptsize{MIPL}} dataset but struggles with complex ones. In contrast, {\sam} and {\mam} excel at complex feature representations in the CRC-{\scriptsize{MIPL}} dataset but are not the best with simpler ones. Therefore, the choice of attention mechanism should align with the dataset's feature complexity. For simple feature representations, {\dam} is optimal, whereas complex representations, such as those derived from deep learning-based feature extractors, benefit from the advanced capabilities of {\sam} and {\mam}.

\begin{table*}[!t]
\centering  
 \caption{Expected calibration error (mean$\pm$std\%) on the real-world datasets.}
\label{tab:calibration_crc}
\begin{tabular}{p{1.2cm} cccccccccccccc}
\hline  \hline
\multicolumn{1}{l}{}  	& \multicolumn{2}{c}{C-Row}   				& \multicolumn{2}{c}{C-SBN}   				& \multicolumn{2}{c}{C-KMeans}  				& \multicolumn{2}{c}{C-SIFT}  		& \multicolumn{2}{c}{C-R34-9}		& \multicolumn{2}{c}{C-R34-16}		& \multicolumn{2}{c}{C-R34-25} \\
\hline
\multicolumn{1}{l}{{\promipl}}  	& \multicolumn{2}{c}{17.99$\pm$1.64} 			& \multicolumn{2}{c}{25.48$\pm$0.65} 			& \multicolumn{2}{c}{31.52$\pm$1.11}  			& \multicolumn{2}{c}{31.42$\pm$1.12} 	& \multicolumn{2}{c}{34.30$\pm$0.99}		& \multicolumn{2}{c}{37.72$\pm$0.72}		& \multicolumn{2}{c}{41.31$\pm$0.85} \\
\multicolumn{1}{l}{{\fastmipl}}  	& \multicolumn{2}{c}{50.54$\pm$3.78} 			& \multicolumn{2}{c}{42.17$\pm$3.07} 			& \multicolumn{2}{c}{41.62$\pm$1.27}  			& \multicolumn{2}{c}{43.48$\pm$2.75} 	& \multicolumn{2}{c}{41.01$\pm$1.60}		& \multicolumn{2}{c}{36.19$\pm$1.46}		& \multicolumn{2}{c}{33.97$\pm$1.62}\\ 
\hdashline
\multicolumn{1}{l}{{\demipl}}  	& \multicolumn{2}{c}{16.40$\pm$1.06} 			& \multicolumn{2}{c}{22.59$\pm$1.10}  			& \multicolumn{2}{c}{27.82$\pm$1.09}			& \multicolumn{2}{c}{28.17$\pm$0.99} 	& \multicolumn{2}{c}{32.89$\pm$1.37}		& \multicolumn{2}{c}{36.37$\pm$0.80}		& \multicolumn{2}{c}{39.23$\pm$0.85}\\	
\multicolumn{1}{l}{\multirow{2}[0]{*}{{\damcc} (ours)}}	& \multicolumn{2}{c}{16.10$\pm$0.57} 			& \multicolumn{2}{c}{15.13$\pm$0.61} 			& \multicolumn{2}{c}{10.39$\pm$0.71}			& \multicolumn{2}{c}{23.42$\pm$1.21} 	& \multicolumn{2}{c}{20.90$\pm$1.37}		& \multicolumn{2}{c}{20.42$\pm$0.84}		& \multicolumn{2}{c}{19.83$\pm$0.98}\\
& \multicolumn{2}{c}{($0.30\downarrow$)} 			& \multicolumn{2}{c}{($7.46\downarrow$)} 			& \multicolumn{2}{c}{($17.43\downarrow$)}			& \multicolumn{2}{c}{($4.75\downarrow$)} 	& \multicolumn{2}{c}{($11.99\downarrow$)}		& \multicolumn{2}{c}{($15.95\downarrow$)}		& \multicolumn{2}{c}{($19.40\downarrow$)}	\\
\multicolumn{1}{l}{\multirow{2}[0]{*}{{\damcn} (ours)}}	& \multicolumn{2}{c}{10.54$\pm$0.86} 			& \multicolumn{2}{c}{12.12$\pm$0.57} 			& \multicolumn{2}{c}{\textbf{9.67$\pm$0.44}}			& \multicolumn{2}{c}{21.38$\pm$1.08} 	& \multicolumn{2}{c}{19.96$\pm$0.78}		& \multicolumn{2}{c}{20.12$\pm$0.77}		& \multicolumn{2}{c}{19.59$\pm$1.01}\\
& \multicolumn{2}{c}{($5.86\downarrow$)} 			& \multicolumn{2}{c}{($10.47\downarrow$)} 			& \multicolumn{2}{c}{($18.15\downarrow$)}			& \multicolumn{2}{c}{($6.79\downarrow$)} 	& \multicolumn{2}{c}{($12.93\downarrow$)}		& \multicolumn{2}{c}{($16.25\downarrow$)}		& \multicolumn{2}{c}{($19.64\downarrow$)}	\\
\hdashline

\multicolumn{1}{l}{{\elimipl}}  	& \multicolumn{2}{c}{18.21$\pm$0.83} 			& \multicolumn{2}{c}{24.91$\pm$0.78} 			& \multicolumn{2}{c}{28.99$\pm$1.18} 			& \multicolumn{2}{c}{28.76$\pm$1.04} 	& \multicolumn{2}{c}{34.47$\pm$1.17}		& \multicolumn{2}{c}{36.70$\pm$0.67}		& \multicolumn{2}{c}{40.24$\pm$0.66}\\		
\multicolumn{1}{l}{\multirow{2}[0]{*}{{\samcc} (ours)}}	& \multicolumn{2}{c}{15.52$\pm$0.51} 			& \multicolumn{2}{c}{14.58$\pm$0.93} 			& \multicolumn{2}{c}{10.83$\pm$0.69}			& \multicolumn{2}{c}{17.73$\pm$0.86} 	& \multicolumn{2}{c}{18.38$\pm$1.03}		& \multicolumn{2}{c}{16.88$\pm$0.60}		& \multicolumn{2}{c}{16.04$\pm$0.87}	\\
& \multicolumn{2}{c}{($2.69\downarrow$)} 			& \multicolumn{2}{c}{($10.33\downarrow$)} 			& \multicolumn{2}{c}{($18.16\downarrow$)}			& \multicolumn{2}{c}{($11.03\downarrow$)} 	& \multicolumn{2}{c}{($16.09\downarrow$)}		& \multicolumn{2}{c}{($19.82\downarrow$)}		& \multicolumn{2}{c}{($24.2\downarrow$)}	\\		
\multicolumn{1}{l}{\multirow{2}[0]{*}{{\samcn} (ours)}}	& \multicolumn{2}{c}{9.82$\pm$0.86} 			& \multicolumn{2}{c}{10.66$\pm$0.60} 			& \multicolumn{2}{c}{10.95$\pm$1.27}			& \multicolumn{2}{c}{18.91$\pm$0.68} 		& \multicolumn{2}{c}{20.45$\pm$0.93}		& \multicolumn{2}{c}{18.53$\pm$0.71}		& \multicolumn{2}{c}{16.11$\pm$0.84}  \\
& \multicolumn{2}{c}{($8.39\downarrow$)} 			& \multicolumn{2}{c}{($14.25\downarrow$)} 			& \multicolumn{2}{c}{($18.04\downarrow$)}			& \multicolumn{2}{c}{($9.85\downarrow$)} 	& \multicolumn{2}{c}{($14.02\downarrow$)}		& \multicolumn{2}{c}{($18.17\downarrow$)}		& \multicolumn{2}{c}{($24.13\downarrow$)}	\\
\hdashline

\multicolumn{1}{l}{{\miplma}}  	& \multicolumn{2}{c}{18.40$\pm$0.99} 			& \multicolumn{2}{c}{25.24$\pm$0.68} 			& \multicolumn{2}{c}{30.74$\pm$0.90} 			& \multicolumn{2}{c}{29.48$\pm$0.92} 	& \multicolumn{2}{c}{33.30$\pm$1.25}		& \multicolumn{2}{c}{37.05$\pm$0.73}		& \multicolumn{2}{c}{41.58$\pm$1.05}\\		
\multicolumn{1}{l}{\multirow{2}[0]{*}{{\mamcc} (ours)}}	& \multicolumn{2}{c}{\textbf{6.01$\pm$0.85}} 			& \multicolumn{2}{c}{\textbf{6.23$\pm$1.14}} 			& \multicolumn{2}{c}{10.28$\pm$0.74}			& \multicolumn{2}{c}{6.80$\pm$0.43} 	& \multicolumn{2}{c}{\textbf{9.46$\pm$0.81}}		& \multicolumn{2}{c}{\textbf{8.77$\pm$0.68}}		& \multicolumn{2}{c}{\textbf{8.26$\pm$0.59}}	\\
& \multicolumn{2}{c}{($12.39\downarrow$)} 			& \multicolumn{2}{c}{($19.01\downarrow$)} 			& \multicolumn{2}{c}{($20.46\downarrow$)}			& \multicolumn{2}{c}{($22.68\downarrow$)} 	& \multicolumn{2}{c}{($23.84\downarrow$)}		& \multicolumn{2}{c}{($28.28\downarrow$)}		& \multicolumn{2}{c}{($33.32\downarrow$)}	\\		
\multicolumn{1}{l}{\multirow{2}[0]{*}{{\mamcn} (ours)}}	& \multicolumn{2}{c}{10.76$\pm$0.65} 			& \multicolumn{2}{c}{8.97$\pm$1.16} 			& \multicolumn{2}{c}{11.96$\pm$1.03}			& \multicolumn{2}{c}{\textbf{6.68$\pm$0.34}} 		& \multicolumn{2}{c}{10.25$\pm$0.73}		& \multicolumn{2}{c}{9.25$\pm$0.68}		& \multicolumn{2}{c}{8.50$\pm$0.76}  \\
& \multicolumn{2}{c}{($7.64\downarrow$)} 			& \multicolumn{2}{c}{($16.27\downarrow$)} 			& \multicolumn{2}{c}{($18.78\downarrow$)}			& \multicolumn{2}{c}{($22.8\downarrow$)} 	& \multicolumn{2}{c}{($23.05\downarrow$)}		& \multicolumn{2}{c}{($27.8\downarrow$)}		& \multicolumn{2}{c}{($33.08\downarrow$)}	\\
  \hline  \hline  
\end{tabular}
\end{table*}

\subsection{Calibration Performance}
\label{subsec:calibration}
\subsubsection{ECE on the Benchmark Datasets}
Table \ref{tab:calibration_benchmark} displays the expected calibration error for our methods with five comparative MIPL methods on the benchmark datasets. Notably, {\miplgp} follows the instance-space paradigm, where bag-level predicted labels are aggregated from the instance-level predictions. This method fits the predicted probability and label to the most prominent instance rather than the entire multi-instance bag. Consequently, {\miplgp} generally underperforms compared to MIPL approaches based on the embedded-space paradigm. Therefore, we exclude the calibration results for {\miplgp}. 

The calibration results demonstrate that our CDL significantly reduces the ECE on the MIPL datasets. On the benchmark datasets, CDL reduces the ECE by over $50\%$ in $26$ out of $60$ cases, with the minimum reduction being $16.76\%$, the maximum reduction being $58.76\%$, and the mean reduction being $44.76\%$. Additionally, {\mam} outperforms {\dam} and {\sam} in most cases, which is consistent with the accuracy observed on the benchmark datasets. 

To evaluate the calibration performance of our methods relative to baseline methods, we present reliability diagrams on the test set of the Birdsong-{\scriptsize{MIPL}} dataset. Fig. \ref{fig:ece_birdsong} presents the reliability diagrams on the Birdsong-MIPL dataset. The diagrams display the classification accuracies and ECE based on ten random runs. The baseline methods, {\demipl}, {\elimipl}, and {\miplma}, consistently produce predicted probabilities below $0.25$. Although these methods achieve over $70\%$ classification accuracy, their calibration performance is notably subpar. In contrast, our methods exhibit significant improvements in both calibration performance and classification performance.

\subsubsection{ECE on the Real-World Datasets}
Table~\ref{tab:calibration_crc} presents ECE across the real-world datasets, comparing our proposed methods against state-of-the-art methods in MIPL. 
Our proposed methods consistently achieve lower calibration error than all the comparative methods, demonstrating their effectiveness in improving model calibration. Notably, {\mam} yields the best overall performance, with {\mamcc} attaining the lowest ECE across most datasets. 
The most significant improvement occurs on C-R34-25, where {\mamcc} reduces the ECE by $33.32\%$ compared to {\miplma}. Our methods achieve over a $10\%$ reduction in ECE in $33$ out of $42$ cases. These results indicate that our CDL effectively aligns predicted confidences with classification accuracies, thereby enhancing model calibration.

Our methods exhibit particularly strong performance gains on the C-R34-9, C-R34-16, and C-R34-25 datasets compared to previous MIPL methods such as {\demipl}, {\elimipl}, and {\miplma}. Specifically, when combined with {\dam}, {\sam}, or {\mam}, our CDL achieves up to $19\%$, $24\%$, and $33\%$ lower ECE on the C-R34-25 dataset. These findings highlight our framework's ability to leverage fine-grained image representations for improved calibration, demonstrating its effectiveness on challenging datasets.

\subsection{Discussion}
\label{subsec:discussion}
Appendices B and C provide supplementary empirical evidence for CDL. Appendix B analyzes its feature aggregation and label disambiguation mechanisms, examines sensitivity to $\gamma$, compares CDL with FL/IFL and PLL baselines, and further evaluates robustness, scalability, computational cost, and statistical significance. Appendix C presents a CRC-{\scriptsize{MIPL}} engineering study on patch granularity and confidence-based triage for practical deployment.

Our evaluation confirms that CDL achieves state-of-the-art performance across the benchmark and real-world MIPL datasets. The experimental results highlight three key insights:  
(1) CDL significantly improves classification accuracy, particularly in high-ambiguity scenarios ($r=3$), achieving a mean improvement of $8.93\%$ on benchmark datasets.  
(2) CDL simultaneously enhances calibration, reducing ECE by an average of $44.76\%$ on benchmark datasets due to its dynamic sample weighting mechanism.  
(3) Visualization analyses and probability distribution studies confirm that these improvements result from CDL’s ability to encourage more separable feature distributions while adaptively prioritizing ambiguous labels during training.

The two instantiations of CDL consistently enhance both classification and calibration compared to the baseline models. In terms of classification accuracy, the second instantiation $\mathcal{L}_{\text{CDL-CN}}$ outperforms $\mathcal{L}_{\text{CDL-CC}}$ in most cases. Conversely, $\mathcal{L}_{\text{CDL-CC}}$ generally achieves lower expected calibration error than $\mathcal{L}_{\text{CDL-CN}}$. These observations lead to several insights: First, incorporating non-candidate label confidences can facilitate improved learning in MIPL models. Second, $\mathcal{L}_{\text{CDL-CC}}$ appears more effective when the candidate label set contains semantically similar labels. Third, $\mathcal{L}_{\text{CDL-CN}}$ tends to perform better in scenarios with a high degree of label randomness, as commonly seen in benchmark datasets.
Moreover, performance differences among the three attention mechanisms on real-world datasets suggest that architectural complexity should be matched to the characteristics of the feature space. For example, {\dam} is more efficient for relatively simple features, while {\sam} and {\mam} are better suited for complex representations. No single CDL instantiation or attention mechanism is optimal for both classification and calibration, highlighting the importance of selecting the appropriate combination based on specific task requirements.

\section{Conclusion}
\label{sec:con}
This paper investigates the calibration performance of multi-instance partial-label learning. We propose a calibratable disambiguation loss (CDL), a plug-and-play top-vs-competitor margin-modulated disambiguation loss. 
Theoretically, we show that CDL can be viewed as a margin-modulated MDL objective with an adaptive regularization effect. We further relate top-label calibration to weight alignment and analyze how the additional margin-dependent gradient term and momentum-based weight updates propagate margin information.
Extensive experiments demonstrate CDL’s effectiveness, achieving superior classification performance in $105$ out of $110$ cases and superior calibration performance in $93$ out of $95$ cases compared to state-of-the-art MIPL approaches. These results highlight CDL’s robustness in handling high-ambiguity scenarios and improving the alignment between model confidence and accuracy.

While our approach demonstrates superior classification and calibration performance on challenging datasets like the C-R34-25 dataset, it does not achieve perfect calibration, where the expected calibration error reaches zero. Moreover, CDL introduces a focusing factor $\gamma$ to control the strength of adaptive down-weighting. We use $\gamma=1$ as a robust default in all main experiments, while tuning within the valid range can bring additional calibration gains. Finally, we follow the standard MIPL assumption that the candidate label set contains the true label. When this assumption is violated (missing-true-label cases), learning becomes substantially more difficult, as reflected by our stress test (Appendix~B.6). Future work will investigate adaptive strategies for selecting $\gamma$ and extend calibration as well as disambiguation to more challenging MIPL settings with missing true labels and improved uncertainty-aware learning frameworks \cite{MoghaddamH16}.

\bibliographystyle{IEEEtran}
\bibliography{myref}

\clearpage
\appendices
\renewcommand*{\thefigure}{A\arabic{figure}}
\renewcommand*{\thetable}{A\arabic{table}}
\renewcommand*{\theequation}{A\arabic{equation}}
\setcounter{table}{0}
\setcounter{figure}{0}
\setcounter{equation}{0}
\setcounter{theorem}{0}

\section{Proofs for Theoretical Analysis}
\label{app:cdl_theory}
This section provides the complete proofs for the theoretical results presented in the main text, including the lower-bound, calibration, gradient, logit-space effect, and pseudo-label momentum analyses of CDL. Unless otherwise stated, we use the notation of the main text.

\subsection{Proof of Theorem \ref{thm:cdl_linear_lower_bound}}
\begin{theorem}[Linear lower bound of CDL]
\label{thm:cdl_linear_lower_bound}
Let $\gamma\ge 1$. For the $i$-th training bag, assume that $\hat{\boldsymbol p}_i\in\Delta^{k-1}$, $\hat p_{i,c}>0$ for all $c\in\mathcal Y$, and $\boldsymbol w_i\in\Delta^{k-1}$ with $w_{i,c}=0$ for all $c\notin\mathcal S_i$. Let $u_i=\arg\max_{c\in\mathcal S_i}\hat p_{i,c}$, $q_i=\hat p_{i,u_i}=\max_{c\in\mathcal S_i}\hat p_{i,c}$. Let
\begin{equation}
\label{eq:app_thm1_mdl_cdl_def}
\ell_i^{\mathrm{MDL}}=-\sum_{c\in\mathcal S_i}w_{i,c}\log \hat p_{i,c},\qquad \ell_i^{\mathrm{CDL}}=\lambda_i\ell_i^{\mathrm{MDL}} .
\end{equation}
Then, for each training bag,
\begin{equation}
\label{eq:app_thm1_per_bag_bound}
\ell_i^{\mathrm{CDL}}\ge (1-\gamma\beta_i)\ell_i^{\mathrm{MDL}}=(1-\gamma\beta_i)\left[\mathrm{KL}\!\left(\boldsymbol w_i\middle\|\hat{\boldsymbol p}_i\right)+\mathbb H(\boldsymbol w_i)\right].
\end{equation}
Consequently,
\begin{equation}
\label{eq:thm1_empirical_weighted_bound}
\mathcal L_{\mathrm{CDL}}\ge \frac{1}{m}\sum_{i=1}^{m}(1-\gamma\beta_i)\ell_i^{\mathrm{MDL}} .
\end{equation}
Moreover, if $\beta_{\max}=\max_{1\le i\le m}\beta_i$, then
\begin{equation}
\label{eq:app_thm1_dataset_bound}
\mathcal L_{\mathrm{CDL}}\ge (1-\gamma\beta_{\max})\mathcal L_{\mathrm{MDL}}.
\end{equation}
\end{theorem}

\begin{proof}
Fix an arbitrary training bag indexed by $i$. Since $\hat{\boldsymbol p}_i\in\Delta^{k-1}$, we have $0\le \hat p_{i,c}\le 1$ for every $c\in\mathcal Y$, and therefore $q_i\in[0,1]$. By assumption, the competitor probability $\phi_i$ is well-defined and satisfies $\phi_i\in[0,1]$. Hence
\begin{equation}
\label{eq:app_thm1_beta_upper}
\beta_i=q_i-\phi_i\le 1 .
\end{equation}
It follows from Eq. \eqref{eq:app_thm1_beta_upper} that $1-\beta_i\ge 0$, so the quantity $(1-\beta_i)^\gamma$ is well-defined for every real exponent $\gamma\ge 1$. Consider the function $f(t)=t^\gamma$ on $[0,\infty)$. Since $\gamma\ge 1$, $f$ is convex on $[0,\infty)$, and the first-order supporting hyperplane inequality at $t=1$ gives
\begin{equation}
\label{eq:app_thm1_convex_support}
t^\gamma\ge f(1)+f'(1)(t-1)=1+\gamma(t-1),\qquad t\ge 0 .
\end{equation}
Taking $t=1-\beta_i$ in Eq. \eqref{eq:app_thm1_convex_support} yields the Bernoulli-type bound
\begin{equation}
\label{eq:app_thm1_bernoulli}
(1-\beta_i)^\gamma\ge 1-\gamma\beta_i .
\end{equation}
Next, since $\hat p_{i,c}>0$ and $\hat p_{i,c}\le 1$ for all $c\in\mathcal Y$, we have $\log \hat p_{i,c}\le 0$, and since $w_{i,c}\ge 0$, every summand in the MDL loss is nonnegative. Therefore
\begin{equation}
\label{eq:app_thm1_mdl_nonnegative}
\ell_i^{\mathrm{MDL}}=-\sum_{c\in\mathcal S_i}w_{i,c}\log \hat p_{i,c}\ge 0 .
\end{equation}
Multiplying both sides of Eq. \eqref{eq:app_thm1_bernoulli} by the nonnegative scalar and using the definition $\ell_i^{\mathrm{CDL}}=(1-\beta_i)^\gamma\ell_i^{\mathrm{MDL}}$, we obtain
\begin{equation}
\label{eq:app_thm1_first_lower}
\ell_i^{\mathrm{CDL}}=(1-\beta_i)^\gamma\ell_i^{\mathrm{MDL}}\ge (1-\gamma\beta_i)\ell_i^{\mathrm{MDL}} .
\end{equation}
We now verify the KL-entropy decomposition of $\ell_i^{\mathrm{MDL}}$. Since $\boldsymbol w_i\in\Delta^{k-1}$, $w_{i,c}=0$ for $c\notin\mathcal S_i$, and $\hat p_{i,c}>0$ for all $c\in\mathcal Y$, the KL divergence is finite and can be written as
\begin{equation}
\label{eq:app_thm1_kl_expand}
\begin{aligned}
\mathrm{KL}\!\left(\boldsymbol w_i\middle\|\hat{\boldsymbol p}_i\right) &= \sum_{c\in\mathcal Y}w_{i,c}\log\frac{w_{i,c}}{\hat p_{i,c}} \\
&= \sum_{c\in\mathcal Y}w_{i,c}\log w_{i,c}-\sum_{c\in\mathcal Y}w_{i,c}\log \hat p_{i,c},
\end{aligned}
\end{equation}
where the convention $0\log 0=0$ is used. By the definition of entropy,
\begin{equation}
\label{eq:app_thm1_entropy_def}
\mathbb H(\boldsymbol w_i)=-\sum_{c\in\mathcal S_i}w_{i,c}\log w_{i,c}.
\end{equation}
Combining Eqs. \eqref{eq:app_thm1_kl_expand} and \eqref{eq:app_thm1_entropy_def} gives
\begin{equation}
\label{eq:app_thm1_kl_entropy_sum}
\mathrm{KL}\!\left(\boldsymbol w_i\middle\|\hat{\boldsymbol p}_i\right)+\mathbb H(\boldsymbol w_i)=-\sum_{c\in\mathcal S_i}w_{i,c}\log \hat p_{i,c}.
\end{equation}
Because $w_{i,c}=0$ for every $c\notin\mathcal S_i$, the right-hand side of Eq. \eqref{eq:app_thm1_kl_entropy_sum} reduces to the MDL loss:
\begin{equation}
\label{eq:app_thm1_support_reduction}
-\sum_{c\in\mathcal S_i}w_{i,c}\log \hat p_{i,c}=\ell_i^{\mathrm{MDL}} .
\end{equation}
Hence
\begin{equation}
\label{eq:app_thm1_mdl_kl_entropy}
\ell_i^{\mathrm{MDL}}=\mathrm{KL}\!\left(\boldsymbol w_i\middle\|\hat{\boldsymbol p}_i\right)+\mathbb H(\boldsymbol w_i).
\end{equation}
Substituting Eq. \eqref{eq:app_thm1_mdl_kl_entropy} into Eq. \eqref{eq:app_thm1_first_lower} proves the per-bag lower bound Eq. \eqref{eq:app_thm1_per_bag_bound}. Averaging Eq. \eqref{eq:app_thm1_first_lower} over all $m$ training bags gives
\begin{equation}
\label{eq:app_thm1_average}
\mathcal L_{\mathrm{CDL}}=\frac{1}{m}\sum_{i=1}^{m}\ell_i^{\mathrm{CDL}}\ge \frac{1}{m}\sum_{i=1}^{m}(1-\gamma\beta_i)\ell_i^{\mathrm{MDL}},
\end{equation}
which proves Eq. \eqref{eq:thm1_empirical_weighted_bound}. Finally, by the definition $\beta_{\max}=\max_{1\le i\le m}\beta_i$, we have $\beta_i\le\beta_{\max}$ for all $i$, and since $\gamma>0$,
\begin{equation}
\label{eq:app_thm1_factor_order}
1-\gamma\beta_i\ge 1-\gamma\beta_{\max}.
\end{equation}
Using $\ell_i^{\mathrm{MDL}}\ge 0$ again, Eq. \eqref{eq:app_thm1_factor_order} implies
\begin{equation}
\label{eq:app_thm1_dataset_step}
\frac{1}{m}\sum_{i=1}^{m}(1-\gamma\beta_i)\ell_i^{\mathrm{MDL}}\ge (1-\gamma\beta_{\max})\frac{1}{m}\sum_{i=1}^{m}\ell_i^{\mathrm{MDL}}.
\end{equation}
Since $\mathcal L_{\mathrm{MDL}}=\frac{1}{m}\sum_{i=1}^{m}\ell_i^{\mathrm{MDL}}$, combining Eqs. \eqref{eq:app_thm1_average} and \eqref{eq:app_thm1_dataset_step} gives
\begin{equation}
\label{eq:app_thm1_final_dataset}
\mathcal L_{\mathrm{CDL}}\ge (1-\gamma\beta_{\max})\mathcal L_{\mathrm{MDL}}.
\end{equation}
The factor $1-\gamma\beta_{\max}$ is positive exactly when $1-\gamma\beta_{\max}>0$, equivalently $\gamma\beta_{\max}<1$. This completes the proof.
\end{proof}

\subsection{Proof of Proposition \ref{prop:calibration_confidence_pseudo}}
\label{subsec:proof_calibration_target}
\begin{proposition}[Calibration is controlled by pseudo-label confidence error]
\label{prop:calibration_confidence_pseudo}
Fix the current training state. Assume that $\hat{\boldsymbol p}_i$ and $\boldsymbol w_i$ are measurable with respect to $\mathcal G_i=\sigma(\boldsymbol X_i, \mathcal S_i)$. Then
\begin{equation}
\label{eq:prop_indexed_main_bound}
E_{\mathrm{cal}} \le E_{\mathrm{conf}} \le E_{\mathrm{pconf}}+\delta_w^{\mathrm{TV}}.
\end{equation}
\end{proposition}

\begin{proof}
Since $\hat{\boldsymbol p}_i$ is $\mathcal G_i$-measurable and the tie-breaking rule is fixed, both $\hat y_i$ and $C_i$ are $\mathcal G_i$-measurable. Moreover,
\begin{equation}
\label{eq:prop_indexed_conditional_correctness}
\mathbb E \left[ \mathbb I\{Y_i=\hat y_i\}\mid \mathcal G_i \right] = \eta_{i,\hat y_i}.
\end{equation}
Using the tower property and the fact that $C_i$ is $\mathcal G_i$-measurable gives
\begin{equation}
\label{eq:prop_indexed_tower}
\mathbb E \left[ \mathbb I\{Y_i=\hat y_i\}\mid C_i \right] = \mathbb E \left[ \eta_{i,\hat y_i}\mid C_i \right].
\end{equation}
Therefore, Jensen's inequality yields
\begin{equation}
\label{eq:prop_indexed_first_bound}
E_{\mathrm{cal}} = \mathbb E \left[ \left| \mathbb E \left[ \eta_{i,\hat y_i}-C_i\mid C_i \right] \right| \right] \le \mathbb E \left[ \left|\eta_{i,\hat y_i}-C_i\right| \right] = E_{\mathrm{conf}}.
\end{equation}
For the second inequality, the triangle inequality gives
\begin{equation}
\label{eq:prop_indexed_triangle}
\left|\eta_{i,\hat y_i}-C_i\right| \le \left|w_{i,\hat y_i}-C_i\right| + \left|\eta_{i,\hat y_i}-w_{i,\hat y_i}\right|.
\end{equation}
Since total variation distance dominates the discrepancy of every singleton event,
\begin{equation}
\label{eq:prop_indexed_singleton_tv}
\left|\eta_{i,\hat y_i}-w_{i,\hat y_i}\right| \le d_{\mathrm{TV}}(\boldsymbol\eta_i,\boldsymbol w_i).
\end{equation}
Combining Eqs.~\eqref{eq:prop_indexed_triangle} and \eqref{eq:prop_indexed_singleton_tv}, and then taking expectations, gives
\begin{equation}
\label{eq:prop_indexed_second_bound}
E_{\mathrm{conf}} \le E_{\mathrm{pconf}}+\delta_w^{\mathrm{TV}}.
\end{equation}
The proof follows by combining Eqs.~\eqref{eq:prop_indexed_first_bound} and \eqref{eq:prop_indexed_second_bound}.
\end{proof}

\subsection{Proof of Theorem \ref{thm:pseudo_label_confidence_alignment}}
\label{subsec:proof_confidence_bound}
\begin{theorem}[Pseudo-label confidence alignment bound]
\label{thm:pseudo_label_confidence_alignment}
Assume that $\hat{\boldsymbol p}_i$ is induced by finite logits, so that $\hat p_{i,c}>0$ for all $c\in\mathcal Y$. Assume further that the relevant competitor set in CDL is nonempty and that the CDL modulation satisfies $\lambda_i\ge\lambda_0>0$. Then
\begin{equation}
\label{eq:thm_indexed_pconf_bound}
E_{\mathrm{pconf}}\le\sqrt{2\left(\lambda_0^{-1}R_{\mathrm{CDL}}^w-\mathcal H_w\right)}.
\end{equation}
Consequently,
\begin{equation}
\label{eq:thm_indexed_calibration_bound}
E_{\mathrm{cal}}\le E_{\mathrm{conf}}\le\sqrt{2\left(\lambda_0^{-1}R_{\mathrm{CDL}}^w-\mathcal H_w\right)}+\delta_w^{\mathrm{TV}}.
\end{equation}
\end{theorem}

\begin{proof}
Since $C_i=\hat p_{i,\hat y_i}$, we have
\begin{equation}
\label{eq:indexed_top_coordinate}
\left|w_{i,\hat y_i}-C_i\right|=\left|w_{i,\hat y_i}-\hat p_{i,\hat y_i}\right|.
\end{equation}
A single-coordinate discrepancy is bounded by the full $\ell_1$ distance:
\begin{equation}
\label{eq:indexed_l1_bound}
\left|w_{i,\hat y_i}-\hat p_{i,\hat y_i}\right|\le\sum_{c\in\mathcal Y}\left|w_{i,c}-\hat p_{i,c}\right|=\|\boldsymbol w_i-\hat{\boldsymbol p}_i\|_1.
\end{equation}
By Pinsker's inequality,
\begin{equation}
\label{eq:indexed_pinsker}
\|\boldsymbol w_i-\hat{\boldsymbol p}_i\|_1\le\sqrt{2\,\mathrm{KL}(\boldsymbol w_i\|\hat{\boldsymbol p}_i)}.
\end{equation}
Combining Eqs.~\eqref{eq:indexed_top_coordinate}, \eqref{eq:indexed_l1_bound}, and \eqref{eq:indexed_pinsker} gives
\begin{equation}
\label{eq:indexed_pointwise_pconf}
\left|w_{i,\hat y_i}-C_i\right|\le\sqrt{2\,\mathrm{KL}(\boldsymbol w_i\|\hat{\boldsymbol p}_i)}.
\end{equation}
Taking expectations and applying Jensen's inequality to the concave square-root function yield
\begin{equation}
\label{eq:indexed_expected_kl}
E_{\mathrm{pconf}}\le\sqrt{2\,\mathbb E\left[\mathrm{KL}(\boldsymbol w_i\|\hat{\boldsymbol p}_i)\right]}.
\end{equation}
It remains to upper bound the expected KL divergence by the CDL risk. Since $w_{i,c}=0$ for $c\notin\mathcal S_i$ and $\hat p_{i,c}>0$, the KL-entropy decomposition gives
\begin{equation}
\label{eq:indexed_kl_decomposition}
\mathrm{KL}(\boldsymbol w_i\|\hat{\boldsymbol p}_i)=\ell_i^{\mathrm{MDL}}-\mathbb H(\boldsymbol w_i).
\end{equation}
Taking expectations gives
\begin{equation}
\label{eq:indexed_expected_kl_decomposition}
\mathbb E\left[\mathrm{KL}(\boldsymbol w_i\|\hat{\boldsymbol p}_i)\right]=R_{\mathrm{MDL}}^w-\mathcal H_w.
\end{equation}
Because $\ell_i^{\mathrm{MDL}}\ge 0$ and $\lambda_i\ge\lambda_0>0$ almost surely,
\begin{equation}
\label{eq:indexed_risk_comparison}
R_{\mathrm{MDL}}^w=\mathbb E\left[\ell_i^{\mathrm{MDL}}\right]\le\lambda_0^{-1}\mathbb E\left[\lambda_i\ell_i^{\mathrm{MDL}}\right]=\lambda_0^{-1}R_{\mathrm{CDL}}^w.
\end{equation}
Combining Eqs.~\eqref{eq:indexed_expected_kl}, \eqref{eq:indexed_expected_kl_decomposition}, and \eqref{eq:indexed_risk_comparison} proves \eqref{eq:thm_indexed_pconf_bound}. Finally, Eq.~\eqref{eq:thm_indexed_calibration_bound} follows directly from Proposition~\ref{prop:calibration_confidence_pseudo}. The proof is complete.
\end{proof}

\subsection{Proof of Proposition \ref{prop:gradient_decomposition}}
\label{subsec:proof_optimization_dynamics}
\begin{proposition}[Gradient decomposition of CDL on differentiable regions]
\label{prop:gradient_decomposition}
Fix a training bag $(\boldsymbol X_i,\mathcal S_i)$, and let $\theta$ denote the model parameters. Suppose that there is an open parameter region $\mathcal U$ on which each $\hat p_{i,c}(\theta)$ is differentiable and strictly positive. Assume that the top candidate label $u_i=\arg\max_{c\in\mathcal S_i}\hat p_{i,c}(\theta)$ is uniquely attained and remains unchanged on $\mathcal U$. For CDL-CC, let $\mathcal C_i=\mathcal S_i\setminus\{u_i\}$; for CDL-CN, let $\mathcal C_i=\bar{\mathcal S}_i$. Assume that $\mathcal C_i\neq\varnothing$ and that the competitor $v_i=\arg\max_{c\in\mathcal C_i}\hat p_{i,c}(\theta)$ is also uniquely attained and remains unchanged on $\mathcal U$. Define
\begin{equation}
\label{eq:theory_active_qr_beta_lambda_1}
q_i(\theta)=\hat p_{i,u_i}(\theta),\qquad \phi_i(\theta)=\hat p_{i,v_i}(\theta),
\end{equation}
\begin{equation}
\label{eq:theory_active_qr_beta_lambda_2}
\beta_i(\theta)=q_i(\theta)-\phi_i(\theta),\qquad \lambda_i(\theta)=\big(1-\beta_i(\theta)\big)^\gamma.
\end{equation}
During the current gradient computation, regard the pseudo-label weights $\boldsymbol w_i$ as fixed, and write
\begin{equation}
\label{eq:theory_mdl_cdl_theta}
\ell_i^{\mathrm{MDL}}(\theta)=-\sum_{c\in\mathcal S_i}w_{i,c}\log \hat p_{i,c}(\theta),~~~
\ell_i^{\mathrm{CDL}}(\theta)=\lambda_i(\theta)\ell_i^{\mathrm{MDL}}(\theta).
\end{equation}
Then, for every $\theta\in\mathcal U$,
\begin{equation}
\label{eq:theory_gradient_decomposition}
\begin{aligned}
\nabla_{\theta}\ell_i^{\mathrm{CDL}}(\theta)
&= \lambda_i(\theta)\nabla_{\theta}\ell_i^{\mathrm{MDL}}(\theta) \\
&\quad-\gamma\big(1-\beta_i(\theta)\big)^{\gamma-1}\ell_i^{\mathrm{MDL}}(\theta)\nabla_{\theta}\beta_i(\theta).
\end{aligned}
\end{equation}
\end{proposition}

\begin{proof}
Fix an arbitrary $\theta\in\mathcal U$. Since the active top candidate and the active competitor are fixed on $\mathcal U$, the two max operations reduce locally to ordinary coordinate projections:
\begin{equation}
\label{eq:theory_local_coordinate_projection}
\max_{c\in\mathcal S_i}\hat p_{i,c}(\theta)=\hat p_{i,u_i}(\theta),\qquad
\max_{c\in\mathcal C_i}\hat p_{i,c}(\theta)=\hat p_{i,v_i}(\theta).
\end{equation}
Consequently, the local margin is an ordinary differentiable scalar function:
\begin{equation}
\label{eq:theory_beta_local_coordinate}
\beta_i(\theta)=\hat p_{i,u_i}(\theta)-\hat p_{i,v_i}(\theta).
\end{equation}
Taking the gradient of Eq. \eqref{eq:theory_beta_local_coordinate} gives
\begin{equation}
\label{eq:theory_beta_gradient_coordinate}
\nabla_{\theta}\beta_i(\theta)=\nabla_{\theta}\hat p_{i,u_i}(\theta)-\nabla_{\theta}\hat p_{i,v_i}(\theta).
\end{equation}
The positivity of the predictive probabilities ensures that the logarithms in $\ell_i^{\mathrm{MDL}}$ are well defined. Moreover, since $u_i$ and $v_i$ are distinct labels and the predictive vector is a strictly positive probability distribution, we have
\begin{equation}
\label{eq:theory_positive_modulation_base}
1-\beta_i(\theta)=1-\hat p_{i,u_i}(\theta)+\hat p_{i,v_i}(\theta)>0.
\end{equation}
Hence $\lambda_i(\theta)=\big(1-\beta_i(\theta)\big)^\gamma$ is differentiable on $\mathcal U$. By the definition of CDL in the margin-modulated form,
\begin{equation}
\label{eq:theory_cdl_product_form}
\ell_i^{\mathrm{CDL}}(\theta)=\lambda_i(\theta)\ell_i^{\mathrm{MDL}}(\theta).
\end{equation}
Applying the product rule to Eq. \eqref{eq:theory_cdl_product_form} yields
\begin{equation}
\label{eq:theory_product_rule}
\nabla_{\theta}\ell_i^{\mathrm{CDL}}(\theta)
=
\lambda_i(\theta)\nabla_{\theta}\ell_i^{\mathrm{MDL}}(\theta)
+
\ell_i^{\mathrm{MDL}}(\theta)\nabla_{\theta}\lambda_i(\theta).
\end{equation}
It remains to compute $\nabla_{\theta}\lambda_i(\theta)$. By the chain rule,
\begin{equation}
\label{eq:theory_lambda_chain_rule}
\nabla_{\theta}\lambda_i(\theta)
=
\nabla_{\theta}\big(1-\beta_i(\theta)\big)^\gamma
=
-\gamma\big(1-\beta_i(\theta)\big)^{\gamma-1}\nabla_{\theta}\beta_i(\theta).
\end{equation}
Substituting Eq. \eqref{eq:theory_lambda_chain_rule} into Eq. \eqref{eq:theory_product_rule} gives
\begin{equation}
\label{eq:theory_gradient_decomposition_proof}
\begin{aligned}
\nabla_{\theta}\ell_i^{\mathrm{CDL}}(\theta)
&= \lambda_i(\theta)\nabla_{\theta}\ell_i^{\mathrm{MDL}}(\theta) \\
&\quad-\gamma\big(1-\beta_i(\theta)\big)^{\gamma-1}
\ell_i^{\mathrm{MDL}}(\theta)\nabla_{\theta}\beta_i(\theta).
\end{aligned}
\end{equation}
This is Eq. \eqref{eq:theory_gradient_decomposition}. When $\boldsymbol w_i$ is detached during the current optimization step, the MDL gradient appearing above is
\begin{equation}
\label{eq:theory_mdl_gradient_detached_weight}
\begin{aligned}
\nabla_{\theta}\ell_i^{\mathrm{MDL}}(\theta)
&= -\sum_{c\in\mathcal S_i}w_{i,c}\nabla_{\theta}\log \hat p_{i,c}(\theta) \\
&= -\sum_{c\in\mathcal S_i}\frac{w_{i,c}}{\hat p_{i,c}(\theta)}
\nabla_{\theta}\hat p_{i,c}(\theta).
\end{aligned}
\end{equation}
If one differentiates through the pseudo-label update itself, the same product-rule identity still holds, but $\nabla_{\theta}\ell_i^{\mathrm{MDL}}(\theta)$ must then be interpreted as the full derivative, including the derivatives of $w_{i,c}(\theta)$. The ordinary-gradient statement above is local: at exact ties of the top candidate or the competitor, the max-based margin is generally not differentiable and must instead be handled with subdifferentials.
\end{proof}

\subsection{Proof of Corollary \ref{cor:logit_margin_direction}}
\begin{corollary}[Logit-space effect of margin shaping]
\label{cor:logit_margin_direction}
Fix a training bag $i$. Let $s_{i,c}\in\mathbb R$ be the logit of class $c$, and let
\begin{equation}
\label{eq:theory_cor3_softmax}
\hat p_{i,c}
=
\frac{\exp(s_{i,c})}{\sum_{a\in\mathcal Y}\exp(s_{i,a})},
\qquad c\in\mathcal Y .
\end{equation}
Assume that, in a neighborhood of the current logits, the top candidate $u$ and the competitor $v$ are unique, distinct, and fixed. Then $\beta_i=\hat p_{i,u}-\hat p_{i,v}$ is differentiable in this neighborhood, and
\begin{align}
\label{eq:theory_cor3_top}
\frac{\partial \beta_i}{\partial s_{i,u}}
&=\hat p_{i,u}(1-\hat p_{i,u}+\hat p_{i,v})>0,\\
\label{eq:theory_cor3_competitor}
\frac{\partial \beta_i}{\partial s_{i,v}}
&=-\hat p_{i,v}(1+\hat p_{i,u}-\hat p_{i,v})<0,\\
\label{eq:theory_cor3_other}
\frac{\partial \beta_i}{\partial s_{i,c}}
&=\hat p_{i,c}(\hat p_{i,v}-\hat p_{i,u}),~~~~ c\notin\{u,v\}.
\end{align}
Thus a positive step along $\nabla_{s_i}\beta_i$ locally increases the top-candidate logit and decreases the active-competitor logit.
\end{corollary}

\begin{proof}
The local uniqueness assumption fixes the active indices $u$ and $v$, so no derivative of the max operator is involved. For the softmax map,
\begin{equation}
\label{eq:theory_cor3_jacobian}
\frac{\partial \hat p_{i,a}}{\partial s_{i,b}} = \hat p_{i,a}\big(\mathbb I\{a=b\}-\hat p_{i,b}\big),
~~~~ a,b\in\mathcal Y.
\end{equation}
Using $\beta_i=\hat p_{i,u}-\hat p_{i,v}$ and $u\ne v$, we obtain
\begin{align}
\frac{\partial\beta_i}{\partial s_{i,u}}
&=\hat p_{i,u}(1-\hat p_{i,u})+\hat p_{i,u}\hat p_{i,v}
=\hat p_{i,u}(1-\hat p_{i,u}+\hat p_{i,v}),\\
\frac{\partial\beta_i}{\partial s_{i,v}}
&=-\hat p_{i,u}\hat p_{i,v}-\hat p_{i,v}(1-\hat p_{i,v})
=-\hat p_{i,v}(1+\hat p_{i,u}-\hat p_{i,v}),\\
\frac{\partial\beta_i}{\partial s_{i,c}}
&=-\hat p_{i,u}\hat p_{i,c}+\hat p_{i,v}\hat p_{i,c}
=\hat p_{i,c}(\hat p_{i,v}-\hat p_{i,u}),
~~~~ c\notin\{u,v\}.
\end{align}
Since the logits are finite, all softmax probabilities are strictly positive. Moreover,
\begin{equation}
1-\hat p_{i,u}+\hat p_{i,v}\ge 2\hat p_{i,v}>0,
~~~~
1+\hat p_{i,u}-\hat p_{i,v}\ge 2\hat p_{i,u}>0 .
\end{equation}
The strict signs in Eqs. \eqref{eq:theory_cor3_top} and \eqref{eq:theory_cor3_competitor} follow immediately.
\end{proof}

\subsection{Proof of Lemma \ref{lem:pseudo_label_margin}}
\begin{lemma}[Momentum recursion for candidate pseudo-label margins]
\label{lem:pseudo_label_margin}
Fix a training bag $(\boldsymbol X_i,\mathcal S_i)$ and two candidate labels $u,v\in\mathcal S_i$. For $t=2,\ldots,T$, define
\begin{equation}
\label{eq:theory_margin_dynamic_def}
\tilde p_{i,c}^{(t)}
=
\frac{\hat p_{i,c}^{(t)}}{\sum_{a\in\mathcal S_i}\hat p_{i,a}^{(t)}},
~~~
M_{i,uv}^{(t)}=w_{i,u}^{(t)}-w_{i,v}^{(t)},
~~~
\Delta_{i,uv}^{(t)}=\tilde p_{i,u}^{(t)}-\tilde p_{i,v}^{(t)} .
\end{equation}
Assume that the pseudo-label weights are updated by
\begin{equation}
\label{eq:theory_weight_dynamic_update}
w_{i,c}^{(t)}
=
\alpha^{(t)}w_{i,c}^{(t-1)}
+
\big(1-\alpha^{(t)}\big)\tilde p_{i,c}^{(t)},
~~~
c\in\mathcal S_i .
\end{equation}
Then, for every $t=2,\ldots,T$,
\begin{equation}
\label{eq:theory_margin_dynamic_one_step}
M_{i,uv}^{(t)}
=
\alpha^{(t)}M_{i,uv}^{(t-1)}
+
\big(1-\alpha^{(t)}\big)\Delta_{i,uv}^{(t)} .
\end{equation}
Equivalently, with $A_{a:b}=\prod_{\tau=a}^{b}\alpha^{(\tau)}$ and the empty product defined as one,
\begin{equation}
\label{eq:theory_margin_dynamic_closed_form}
M_{i,uv}^{(t)}
=
A_{2:t}M_{i,uv}^{(1)}
+
\sum_{s=2}^{t}
\big(1-\alpha^{(s)}\big)A_{s+1:t}\Delta_{i,uv}^{(s)} .
\end{equation}
Moreover,
\begin{equation}
\label{eq:theory_delta_raw_prediction_margin}
\Delta_{i,uv}^{(s)}
=
\frac{\hat p_{i,u}^{(s)}-\hat p_{i,v}^{(s)}}
{\sum_{a\in\mathcal S_i}\hat p_{i,a}^{(s)}} .
\end{equation}
Thus past candidate prediction margins enter the current pseudo-label margin through the momentum coefficients. If $\alpha^{(s)}\in[0,1]$ for all $s$, then $M_{i,uv}^{(t)}$ is a convex combination of $M_{i,uv}^{(1)}$ and $\Delta_{i,uv}^{(2)},\ldots,\Delta_{i,uv}^{(t)}$.
\end{lemma}

\begin{proof}
Applying Eq. \eqref{eq:theory_weight_dynamic_update} to $u$ and $v$, respectively, and subtracting the two identities, we obtain
\begin{equation}
\label{eq:theory_margin_dynamic_one_step_proof}
\begin{aligned}
w_{i,u}^{(t)}-w_{i,v}^{(t)}
&=
\alpha^{(t)}\left(w_{i,u}^{(t-1)}-w_{i,v}^{(t-1)}\right) \\
&\quad+
\big(1-\alpha^{(t)}\big)\left(\tilde p_{i,u}^{(t)}-\tilde p_{i,v}^{(t)}\right).
\end{aligned}
\end{equation}
By the definitions of $M_{i,uv}^{(t)}$ and $\Delta_{i,uv}^{(t)}$, this proves Eq. \eqref{eq:theory_margin_dynamic_one_step}. Iterating Eq. \eqref{eq:theory_margin_dynamic_one_step} from epoch $2$ to epoch $t$ gives
\begin{equation}
\label{eq:theory_margin_dynamic_iteration}
\begin{aligned}
M_{i,uv}^{(t)}
&=
\left(\prod_{\tau=2}^{t}\alpha^{(\tau)}\right)M_{i,uv}^{(1)} \\
&\quad+
\sum_{s=2}^{t}
\left[
\big(1-\alpha^{(s)}\big)
\left(\prod_{\tau=s+1}^{t}\alpha^{(\tau)}\right)
\Delta_{i,uv}^{(s)}
\right],
\end{aligned}
\end{equation}
which is Eq. \eqref{eq:theory_margin_dynamic_closed_form}. Moreover, the identity Eq. \eqref{eq:theory_delta_raw_prediction_margin} follows from Eq. \eqref{eq:theory_margin_dynamic_def}:
\begin{equation}
\label{eq:theory_delta_raw_prediction_margin_proof}
\Delta_{i,uv}^{(s)}
=
\tilde p_{i,u}^{(s)}-\tilde p_{i,v}^{(s)}
=
\frac{\hat p_{i,u}^{(s)}-\hat p_{i,v}^{(s)}}
{\sum_{a\in\mathcal S_i}\hat p_{i,a}^{(s)}}.
\end{equation}
Finally, if $\alpha^{(s)}\in[0,1]$, all coefficients in \eqref{eq:theory_margin_dynamic_closed_form} are nonnegative. Their sum is
\begin{equation}
\label{eq:theory_margin_dynamic_weight_sum}
\left(\prod_{\tau=2}^{t}\alpha^{(\tau)}\right)
+
\sum_{s=2}^{t}
\left[
\big(1-\alpha^{(s)}\big)
\left(\prod_{\tau=s+1}^{t}\alpha^{(\tau)}\right)
\right]
=
1.
\end{equation}
Hence \eqref{eq:theory_margin_dynamic_closed_form} is a convex combination, and the proof is complete. \looseness=-1
\end{proof}

\section{Further Experimental Analysis}
\label{app:analysis}

\begin{figure*}[!t]
    \centering
    \begin{overpic}[width=18cm]{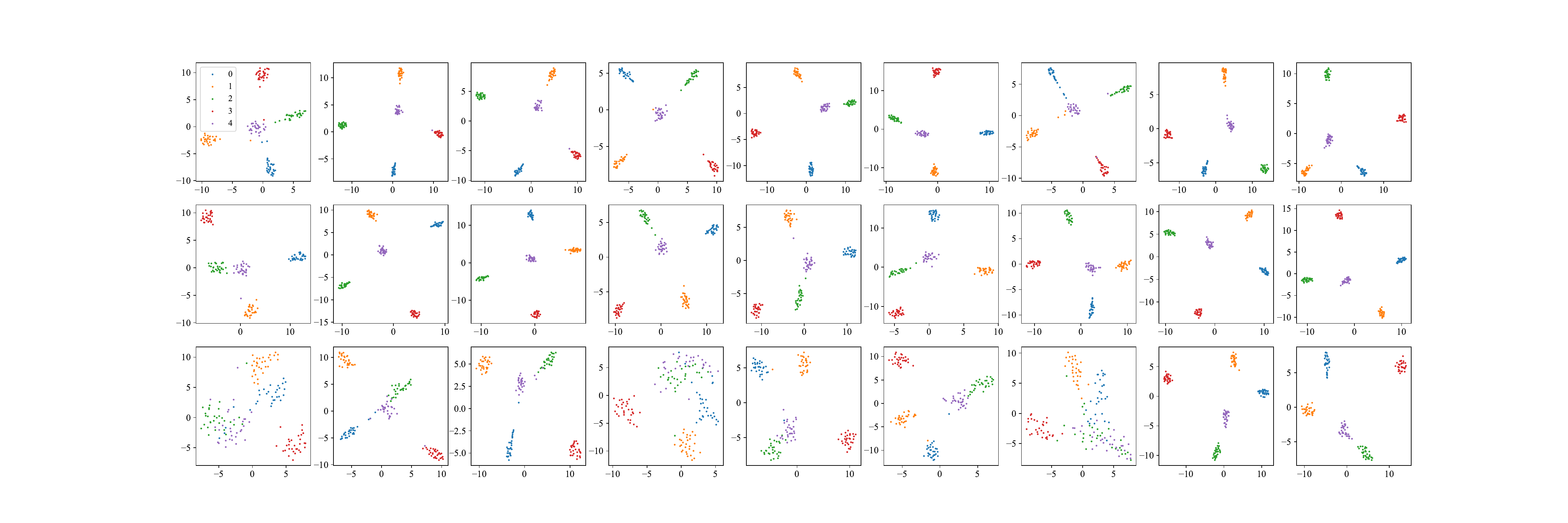}
	\put(0, 50) {\rotatebox{90}{\small $r=3$}} 
	\put(0, 160) {\rotatebox{90}{\small $r=2$}} 
	\put(0, 270) {\rotatebox{90}{\small $r=1$}} 
	\put(40, -5){\small {\demipl}}
	\put(150, -5){\small {\damcc}}
	\put(260, -5){\small {\damcn}}
	\put(370, -5){\small {\elimipl}}
	\put(480, -5){\small {\samcc}}
	\put(590, -5){\small {\samcn}}
        \put(700, -5){\small {\miplma}}
	\put(810, -5){\small {\mamcc}}
	\put(920, -5){\small {\mamcn}}
    \end{overpic}
    \caption{t-SNE visualization of aggregated bag-level feature representations produced by the attention mechanisms in {\demipl} \cite{tang2023demipl}, {\elimipl} \cite{tang2024elimipl}, {\miplma} \cite{tang2024miplma}, and ours on the test set of the MNIST-{\scriptsize{MIPL}} dataset, which comprises five classes. \looseness=-1}

 \label{fig:features_mnist_mipl}
\end{figure*}

\begin{figure}[!t]
    \centering
    \begin{overpic}[width=8.9cm, height=3.6cm]{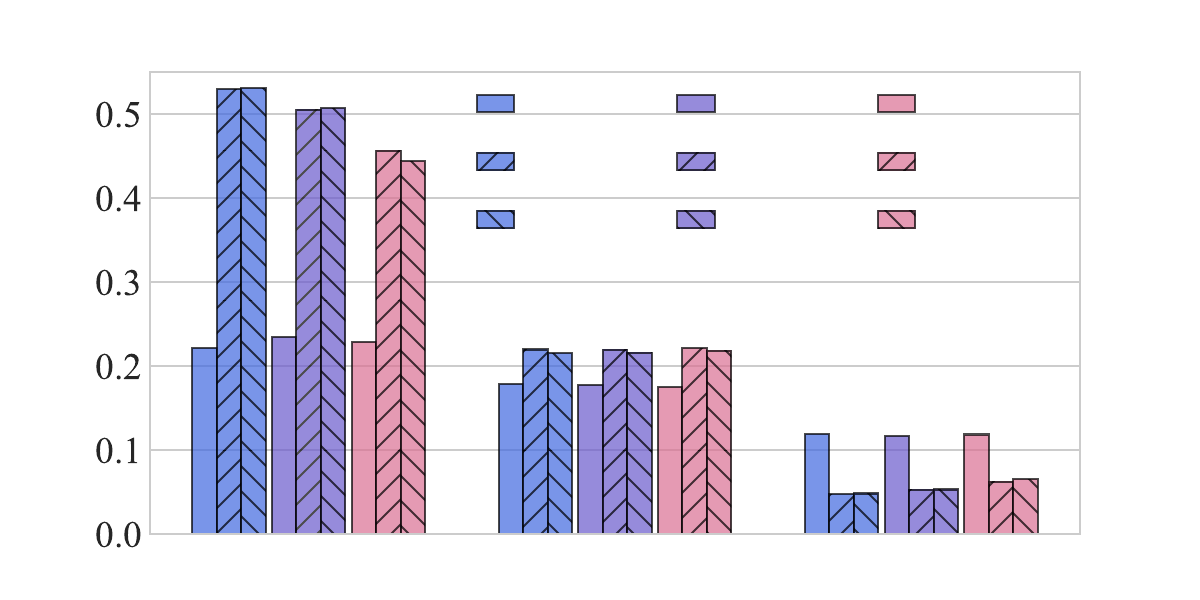}
	\put(455, 361){\small {\demipl}}
	\put(455, 315){\small {\damcc}}
	\put(455, 268){\small {\damcn}}
	\put(650, 361){\small {\elimipl}}
	\put(650, 315){\small {\samcc}}
	\put(650, 268){\small {\samcn}}
	\put(845, 361){\small {\miplma}}
	\put(845, 315){\small {\mamcc}}
	\put(845, 268){\small {\mamcn}}
	\put(190, -10){\small T-labels}
	\put(480, -10){\small FP-labels}
	\put(765, -10){\small NC-labels}
	\put(-5, 90) {\rotatebox{90}{Probability}} 
    \end{overpic}
    \caption{Probabilities at the last epoch on the training set of the C-KMeans dataset. T-labels, FP-labels, and NC-labels denote the average probabilities for true, false positive, and non-candidate labels, respectively.}
 \label{fig:prob_ckmeans}
\end{figure}

\begin{figure*}[!t]
\setlength{\abovecaptionskip}{0.cm} 
    \centering
    \begin{overpic}[width=17cm]{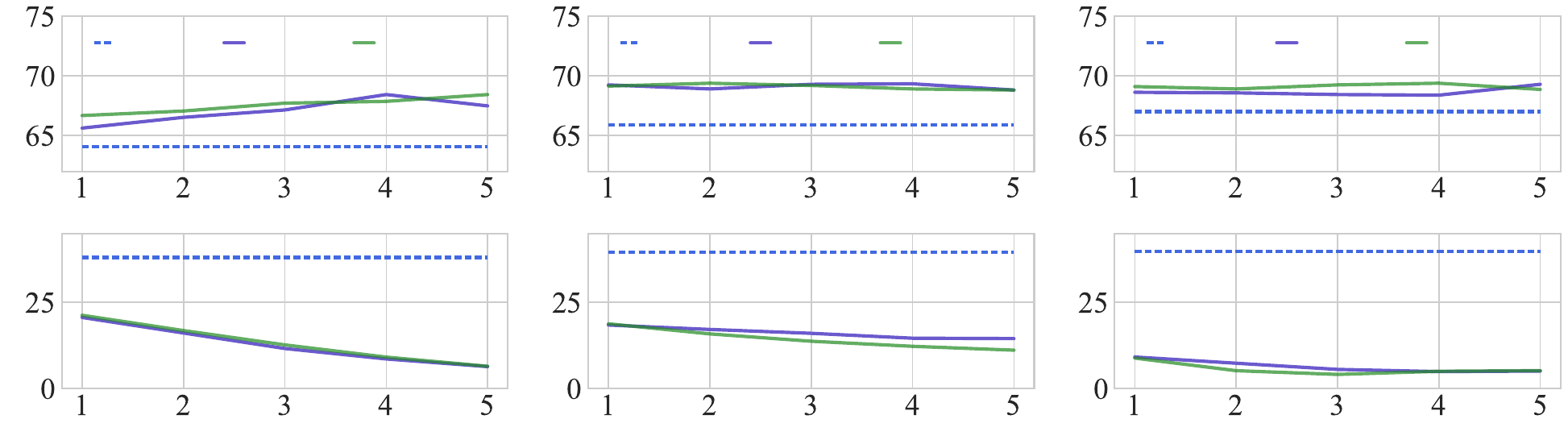}
    \put(74, 248){\scriptsize {\demipl}}
	\put(160, 248){\scriptsize {\damcc}}
	\put(242, 248){\scriptsize {\damcn}}
    \put(412, 248){\scriptsize {\elimipl}}
	\put(498, 248){\scriptsize {\samcc}}
	\put(580, 248){\scriptsize {\samcn}}
    \put(747, 248){\scriptsize {\miplma}}
	\put(833, 248){\scriptsize {\mamcc}}
	\put(915, 248){\scriptsize {\mamcn}}
    \put(-5, 190) {\small \rotatebox{90}{Accuracy}} 
    \put(-5, 68) {\small \rotatebox{90}{ECE}}
    \put(173, 140){\scriptsize (a)}
    \put(173, 1){\scriptsize (b)}
    \put(510, 140){\scriptsize (c)}
    \put(510, 1){\scriptsize (d)}
    \put(845, 140){\scriptsize (e)}
    \put(845, 1){\scriptsize (f)}
    \end{overpic}
    \caption{Classification accuracy and expected calibration error of {\demipl} (a, b), {\elimipl} (c, d), {\miplma} (e, f), and our six methods on the C-R34-25 dataset for $\gamma \in \{1, 2, 3, 4, 5\}$. The horizontal axis represents $\gamma$.}
\label{fig:varying_gamma_on_C_R25}
\end{figure*}

\subsection{Mechanisms Underlying the Effectiveness of CDL}
\label{subsec:res_cdl}
The calibratable disambiguation loss (CDL) significantly improves both classification and calibration performance compared to baseline methods {\demipl}, {\elimipl}, and {\miplma}. To elucidate the reasons behind its effectiveness, we analyze its impact on two critical components of MIPL approaches based on the embedded-space paradigm: feature aggregation and label disambiguation.

\subsubsection{Enhancing Feature Aggregation}
To evaluate the impact on feature aggregation, we visualize the bag-level feature representations, \ie $z_i$ from Eq. (13), for the baseline methods {\demipl}, {\elimipl}, {\miplma}, and our proposed methods on the MNIST-{\scriptsize{MIPL}} dataset. Unlike the standard MNIST dataset, which comprises $10$ target classes, the MNIST-{\scriptsize{MIPL}} dataset focuses on $5$ target classes, with negative instances sourced from the remaining $5$ classes \cite{tang2023miplgp}. For $r=1$ and $r=2$, {\demipl}, {\elimipl}, and {\miplma} produce reasonably good results. However, when $r=3$, the feature representations aggregated by the three methods become increasingly disordered, with features from different classes becoming mixed. In contrast, our methods consistently produce well-separated feature representations for $r \in \{1, 2, 3\}$. Specifically, for $r=1$ and $r=2$, the class clusters are compact and distinctly separable. Although the clusters are somewhat dispersed at $r=3$, our methods remain notably more distinct and separable compared to those generated by the baseline methods {\demipl}, {\elimipl}, and {\miplma}.

Therefore, the visualizations in Fig. \ref{fig:features_mnist_mipl} demonstrate that our proposed CDL significantly improves the aggregation of bag-level feature representations, resulting in more compact and distinguishable clusters. This enhancement directly contributes to superior classification performance.

\subsubsection{Optimizing Label Disambiguation and Calibration}
The disambiguation performance and calibration performance of a model are directly influenced by its predicted probabilities. As shown in Fig. 2, the baseline methods {\demipl}, {\elimipl}, and {\miplma} generate low predicted probabilities for true labels on the training set, leading to poor calibration performance. In contrast, our methods yield significantly higher predicted probabilities for true labels compared to these baseline methods. To further investigate the effect of CDL on predicted probabilities, we present the average predicted probabilities for true labels (T-labels), false-positive labels (FP-labels), and non-candidate labels (NC-labels) on the training set of the C-KMeans dataset.

As illustrated in Fig. \ref{fig:prob_ckmeans}, for {\demipl}, {\elimipl}, and {\miplma}, the probabilities for true labels are comparable to those for false-positive labels. In contrast, our methods achieve markedly higher probabilities for true labels compared to false-positive labels. Although our methods also show higher probabilities for false-positive labels relative to {\demipl}, {\elimipl}, and {\miplma}, the proportion of probabilities for false-positive labels relative to candidate labels is substantially lower. Additionally, our methods achieve significantly lower predicted probabilities for non-candidate labels compared to the baseline methods.

According to the definition of CDL, there exists an inverse relationship between the maximum predicted probabilities on candidate labels and the outcome of $\Phi(\cdot)$. For samples with high maximum predicted probabilities on candidate labels, the outcome of $\Phi(\cdot)$ is relatively small, resulting in lower loss values for these samples. Samples with low maximum predicted probabilities produce a larger outcome of $\Phi(\cdot)$, leading to higher loss values. 
As a result, the model focuses more on these high-loss, low-margin cases and increases their separability, which alleviates under-confidence. When a bag becomes well separated, the modulation decreases and automatically limits further sharpening, which helps mitigate over-confidence.

\subsection{Parameter Sensitivity}
In our proposed CDL, the only hyperparameter is the exponential factor $\gamma$. Consistent with focal loss \cite{LinGGHD20, WangFZ21}, we treat $\gamma$ as a constant throughout our experiments. To evaluate the sensitivity of our methods to $\gamma$, Fig. \ref{fig:varying_gamma_on_C_R25} presents the accuracy and expected calibration error of {\demipl}, {\elimipl}, {\miplma}, and our six methods with $\gamma$ ranging over $\{1, 2, 3, 4, 5\}$ on the C-R34-25 dataset.

Across all $\gamma$ settings, our methods consistently outperform {\demipl}, {\elimipl}, and {\miplma} in both classification and calibration performance. As $\gamma$ increases from $1$ to $5$, we observe a general improvement in calibration performance, though the gains diminish at higher values. Notably, {\elimipl} and {\miplma} exhibit greater robustness to variations in $\gamma$ compared to {\demipl}, likely due to their more sophisticated attention mechanisms, which provide better resistance to hyperparameter fluctuations. Furthermore, the two CDL instantiations yield comparable results, indicating that CDL's effectiveness is largely independent of the specific instantiation.  \looseness=-1

\subsection{Comparison with FL and IFL}
\label{subsec:fl_ilf}
To evaluate the effectiveness of CDL compared to focal loss (FL) and inverse focal loss (IFL), we integrate FL and IFL into {\mam}, resulting in two variants: {\mamfl} and {\mamifl}. The corresponding loss functions are defined in Eqs. (5) and (6), respectively, where the weights $w_{i,c}^{(t)}$ of candidate labels are initialized as $w_{i,c}^{(1)} = \frac{1}{\left|\mathcal{S}_i\right|}$ at $t=1$ and updated using Eq. (15). Therefore, the only distinction between {\mamfl}/{\mamifl} and our proposed {\mamcc}/{\mamcn} lies in the exponential term of the loss functions. 

\begin{figure}[!t]
\setlength{\abovecaptionskip}{0.1cm}  
    \centering
    \begin{overpic}[width=8.8cm]{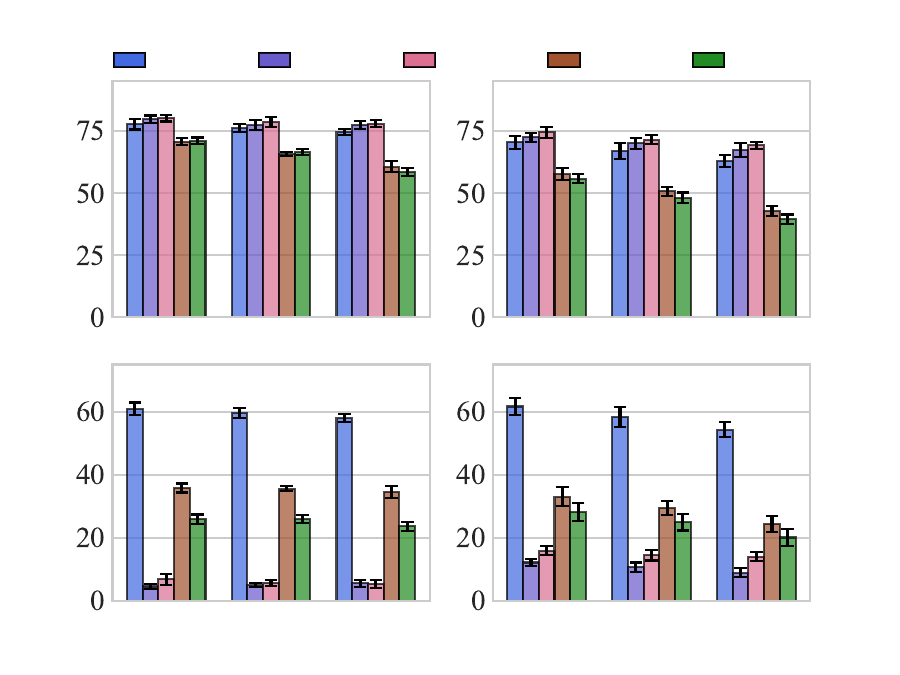}
    	\put(135, 730){\scriptsize {\miplma}}
        \put(325, 730){\scriptsize {\mamcc}}
        \put(515, 730){\scriptsize {\mamcn}}
        \put(700, 730){\scriptsize {\mamfl}}
        \put(890, 730){\scriptsize {\mamifl}}
        \put(110, 370){\small $r=1$}
        \put(250, 370){\small $r=2$}
        \put(390, 370){\small $r=3$}
        \put(610, 370){\small $r=1$}
        \put(740, 370){\small $r=2$}
        \put(880, 370){\small $r=3$}
        \put(110, 5){\small $r=1$}
        \put(250, 5){\small $r=2$}
        \put(390, 5){\small $r=3$}
        \put(610, 5){\small $r=1$}
        \put(740, 5){\small $r=2$}
        \put(880, 5){\small $r=3$}   
        \put(0, 480) {\small \rotatebox{90}{Accuracy}} 
        \put(0, 150) {\small \rotatebox{90}{ECE}}
        \put(100, 680){\small (a)}
        \put(100, 310){\small (b)}
        \put(590, 680){\small (c)}
        \put(590, 310){\small (d)}
    \end{overpic}
    \caption{Classification accuracy and expected calibration error (mean and std) of {\miplma}, {\mamcc}, {\mamcn}, {\mamfl}, and {\mamifl} on the Birdsong-{\scriptsize{MIPL}} (a, b) and SIVAL-{\scriptsize{MIPL}} (c, d) datasets. \looseness=-1}
\label{fig:fl_ifl_on_birdsong_sival}
\end{figure}

Fig. \ref{fig:fl_ifl_on_birdsong_sival} presents the classification accuracy and expected calibration error of {\miplma}, {\mamcc}, {\mamcn}, {\mamfl}, and {\mamifl} on the Birdsong-{\scriptsize{MIPL}} and SIVAL-{\scriptsize{MIPL}} datasets, with the number of false-positive labels $r \in \{1,2,3\}$.  As shown in Fig. \ref{fig:fl_ifl_on_birdsong_sival}, {\mamfl} and {\mamifl} improve the calibration of {\miplma} by reducing ECE. However, this improvement comes at the expense of classification accuracy, which declines more significantly as the number of false-positive labels increases. Furthermore, both {\mamfl} and {\mamifl} underperform compared to our proposed CDL in both classification and calibration. These findings indicate that FL and IFL are insufficient for handling classification and calibration challenges in MIPL.

\begin{figure}[!t]
    \centering
    \begin{overpic}[width=8.8cm, height=9cm]{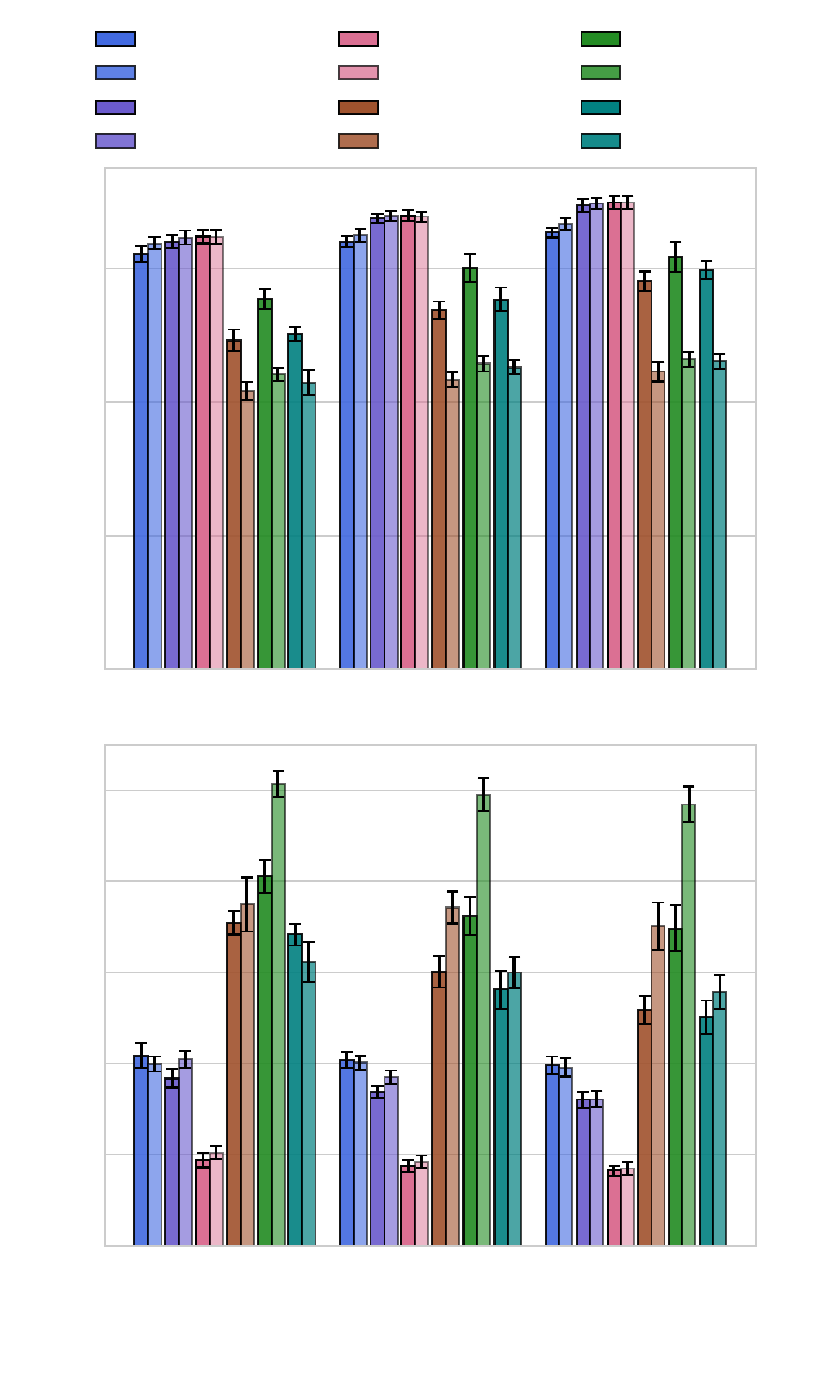}
	\put(130, 980){\scriptsize {\damcc} (ours)}
	\put(130, 953){\scriptsize {\damcn} (ours)}
	\put(130, 925){\scriptsize {\samcc} (ours)}
	\put(130, 896){\scriptsize {\samcn} (ours)}
	\put(450, 980){\scriptsize {\mamcc} (ours)}
	\put(450, 953){\scriptsize {\mamcn} (ours)}
	\put(450, 925){\scriptsize {\proden} (Mean)}
	\put(450, 896){\scriptsize {\proden} (MaxMin)}
        \put(770, 980){\scriptsize {\lws} (Mean)}
	\put(770, 953){\scriptsize {\lws} (MaxMin)}
	\put(770, 925){\scriptsize {\pop} (Mean)}
	\put(770, 896){\scriptsize {\pop} (MaxMin)}
        \put(30, 795){\small 60}
        \put(30, 685){\small 40}
        \put(30, 580){\small 20}
        \put(40, 475){\small 0}
        \put(170, 450){\small C-R34-9}
        \put(435, 450){\small C-R34-16}
        \put(705, 450){\small C-R34-25}
        \put(30, 378){\small 50}
        \put(30, 302){\small 40}
        \put(30, 232){\small 30}
        \put(30, 160){\small 20}
        \put(30, 88){\small 10}
        \put(40, 10){\small 0}
        \put(170, -10){\small C-R34-9}
        \put(435, -10){\small C-R34-16}
        \put(705, -10){\small C-R34-25}
        \put(-10, 630) {\small \rotatebox{90}{Accuracy}} 
        \put(-10, 190) {\small \rotatebox{90}{ECE}}
        \put(90, 850){\small (a)}
        \put(90, 385){\small (b)}
    \end{overpic}
    \caption{Classification accuracy and expected calibration error (mean and std) of our six methods and three PLL methods on the CRC-{\scriptsize{MIPL}} datasets with ResNet-34 as the feature extractor.}
\label{fig:pll_crc_resnet}
\end{figure}

\subsection{Comparison with PLL Methods}
We compare our methods with three popular PLL methods: {\proden} \cite{LvXF0GS20}, {\lws} \cite{WenCHL0L21}, and {\pop} \cite{XuLLQG23}. {\proden} is a classical PLL method based on deep learning that employs progressive disambiguation loss to identify true labels from candidate label sets. {\lws} introduces a weighted disambiguation loss to refine the true labels. {\pop} is an instance-dependent PLL method that trains the classifier using progressive purification of candidate labels. To adapt MIPL data for PLL methods, we employ two strategies from the existing MIPL literature \cite{tang2023miplgp} to convert MIPL data into PLL data. 1) Mean: For each bag, we compute the average value of instances in each feature dimension as the representative feature value for that dimension. This strategy generates a holistic feature vector that retains the same dimensionality as the original instances. 2) MaxMin: We derive the maximum and minimum values across all instances for each feature dimension and concatenate these values. This strategy produces a holistic feature vector with a combined dimensionality of $2d$.  \looseness=-1

Fig. \ref{fig:pll_crc_resnet} presents the classification accuracy and expected calibration error of our six methods and the three PLL methods on the CRC-{\scriptsize{MIPL}} dataset using ResNet-34 features. Our methods consistently achieve the highest classification accuracy and the lowest expected calibration error across all three datasets. The results demonstrate that our methods substantially outperform the comparative PLL methods.

\begin{table*}[!t]
\setlength{\abovecaptionskip}{0.cm} 
\centering  
\caption{Classification accuracy and expected calibration error (mean$\pm$std\%) of {\pop}, {\pop}-{\cc}, and {\pop}-{\cn} on the CRC-{\scriptsize{MIPL}} datasets using ResNet-34.}
\label{tab:pop_cdl}
\begin{tabular}{p{1.2cm} cccccc}
\hline  \hline
\multicolumn{1}{l}{\multirow{2}[1]{*}{}} 	& \multicolumn{2}{c}{C-R34-9}		& \multicolumn{2}{c}{C-R34-16}		& \multicolumn{2}{c}{C-R34-25} \\
\cdashline{2-7}
&      Mean  & MaxMin & Mean  & MaxMin & Mean  & MaxMin  \\ 
\hline
\multicolumn{7}{c}{Accuracy} \\  \cdashline{1-7}
\multicolumn{1}{l}{{\pop}} 	& 49.83$\pm$1.28	& 43.23$\pm$1.08		& 54.86$\pm$1.37	& 44.96$\pm$0.93			& 59.06$\pm$1.41	& 46.16$\pm$1.49 	 \\
\multicolumn{1}{l}{{\pop}-{\cc}} 	& 61.85$\pm$1.67	& \textbf{53.74$\pm$1.61}		& 66.36$\pm$1.43	& \textbf{56.11$\pm$0.78}			& 69.17$\pm$1.07	& 57.51$\pm$1.21 	 \\
\multicolumn{1}{l}{{\pop}-{\cn}} 	& \textbf{62.25$\pm$1.55}	& 53.15$\pm$1.66		& \textbf{66.75$\pm$1.05}	& 56.00$\pm$1.25			& \textbf{69.18$\pm$1.01}	& \textbf{57.74$\pm$1.04}	 \\
\cdashline{1-7}
\multicolumn{7}{c}{ECE} \\  \cdashline{1-7}
\multicolumn{1}{l}{{\pop}} 	& 25.01$\pm$1.43	& 29.74$\pm$2.01		& 22.00$\pm$1.22	& 28.87$\pm$1.23			& 19.49$\pm$1.52	& 27.26$\pm$1.41 	 \\
\multicolumn{1}{l}{{\pop}-{\cc}} 	& 16.91$\pm$2.14	& \textbf{18.50$\pm$2.21}		& 13.90$\pm$1.38	& \textbf{16.98$\pm$1.54}			& \textbf{12.94$\pm$1.27}	& \textbf{16.29$\pm$1.53} 	 \\
\multicolumn{1}{l}{{\pop}-{\cn}} 	& \textbf{16.66$\pm$1.45}	& 19.53$\pm$1.83		& \textbf{13.70$\pm$1.31}	& 17.13$\pm$1.71			& 13.08$\pm$1.05	& 16.35$\pm$1.60	 \\
  \hline  \hline  
\end{tabular}
\end{table*}

\subsection{CDL for PLL Methods}
The proposed CDL is a plug-and-play loss function that can be seamlessly incorporated into both MIPL and PLL frameworks. To adapt CDL for PLL, we propose two variants for {\pop} \cite{XuLLQG23}, namely {\pop}-{\cc} and {\pop}-{\cn}. The {\pop}-{\cc} variant integrates the first instantiation $\mathcal{L}_{\text{CDL-CC}}$ into {\pop}, whereas {\pop}-{\cn} incorporates the second instantiation $\mathcal{L}_{\text{CDL-CN}}$.

Table \ref{tab:pop_cdl} reports the classification accuracy and ECE of these three methods on the CRC-{\scriptsize{MIPL}} datasets, using features extracted by ResNet-34. Both variants significantly outperform the vanilla {\pop}, achieving higher classification accuracy and lower expected calibration error, highlighting the effectiveness of CDL in improving both classification and calibration. Notably, the accuracy and expected calibration error of {\pop}-{\cc} and {\pop}-{\cn} are highly similar, indicating that both CDL instantiations are equally effective in enhancing {\pop}'s classification and calibration capability.

\subsection{Additional Robustness Analysis}
\label{app:robustness}

\begin{figure*}[!t]
\setlength{\abovecaptionskip}{0.1cm} 
    \centering
	\begin{overpic}[width=17.0cm]{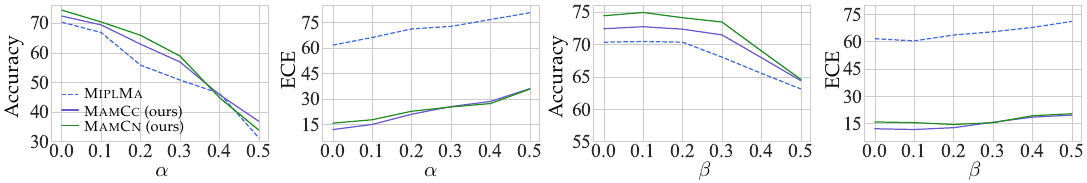}
    \end{overpic}
    \caption{Classification accuracy and expected calibration error of {\miplma}, {\mamcc}, and {\mamcn} on the SIVAL-{\small MIPL} dataset ($r=1$) with varying class noise and attribute noise.}
 \label{fig:varying_noise}
\end{figure*}
\begin{figure*}[!h]
\setlength{\abovecaptionskip}{0.cm} 
    \centering
	\begin{overpic}[width=17.0cm]{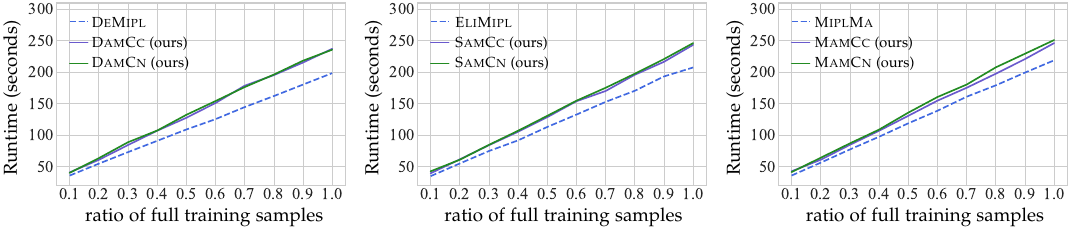}
    \end{overpic}
    \caption{Runtime of training and testing on the MNIST-{\small MIPL} dataset ($r=1$) versus the fraction of full training samples.}
 \label{fig:times}
\end{figure*}

Following the taxonomy of \cite{zhu2004class}, we explicitly distinguish \emph{class noise} and \emph{attribute noise} in MIPL, and we include both a discussion and controlled experiments to substantiate the robustness of our method.

MIPL naturally contains noise in both spaces: the candidate label set is ambiguous and contains false positives; and each bag contains many irrelevant/background instances, and the informative instances are unknown and can be sparse, which introduces instance-level outliers. To quantitatively evaluate robustness under different noise sources, we conduct experiments on the SIVAL-{\small MIPL} dataset with $r=1$ and consider two perturbations: \textbf{(i) Class noise.} We randomly select an $\alpha$ proportion of bags and remove the ground-truth label from their candidate label sets, so that these bags are associated with only false-positive labels. We vary $\alpha \in \{0, 0.1, 0.2, 0.3, 0.4, 0.5\}$. Note that this is a stringent test that violates the standard MIPL assumption that the true label is included in the candidate label set, and it simulates severe label noise.
\textbf{(ii) Attribute noise.} We randomly drop a $\beta$ proportion of instances in each bag, which may remove informative instances (including positive ones) and thus increases the difficulty of the bag. We vary $\beta \in \{0, 0.1, 0.2, 0.3, 0.4, 0.5\}$.

\cref{fig:varying_noise} reports the accuracy and ECE of {\miplma}, {\mamcc}, and {\mamcn}. Under both class noise and attribute noise, our methods consistently achieve higher or comparable accuracy and much lower ECE than the baselines. For example, under severe class noise ($\alpha=0.5$), all methods degrade in accuracy, but {\mamcc} and {\mamcn} remain noticeably better calibrated (ECE $\approx 35\%$ vs.\ $\approx 80\%$ for {\miplma}) and retain slightly higher accuracy.
Under severe attribute noise ($\beta=0.5$), {\mamcc} and {\mamcn} maintain accuracy around $65\%$ while keeping ECE around $20\%$, whereas {\miplma} exhibits substantially worse calibration (ECE $\approx 70\%$). Interestingly, removing a small number of instances may improve performance or leave it unchanged. We speculate that this is because randomly removing some uninformative instances allows the model to focus more on learning from informative ones. Since informative instances are sparse in MIPL, increasing the removal ratio degrades the model’s performance.
Overall, these results support our robustness claim: CDL improves not only classification performance but also, importantly, calibration when the data contain class noise and attribute noise.
This observation is also consistent with our theoretical analysis that CDL acts as an adaptive regularizer and alleviates poor calibration arising from ambiguous supervision.

\begin{table*}[!h]
  \centering   \footnotesize  
   \caption{The computational costs on the MNIST-{\small MIPL} ($r=1$) with full training set.}
      \label{tab:complexity}
    \begin{tabular}{p{2.cm}  ccccc p{0.9cm}  p{0.9cm}  p{0.9cm}  p{0.9cm}}
    \hline  \hline
    \multicolumn{1}{l|}{Algorithm}  	& FLOPs (M)	&   Params (M)  	 & MM (MiB) 	& Runtime (s)  & Acc\\
    \hline
    \multicolumn{1}{l|}{{\demipl}}  	& 95.58		  & 0.13	& 1822	& 198.47 		& 97.60$\pm$0.80\%\\
    \multicolumn{1}{l|}{{\damcc} (ours)}  	    & 95.58 		& 0.13 	& 1858 	      & 237.11 		& 99.33$\pm$0.52\%\\
    \multicolumn{1}{l|}{{\damcn} (ours)}  	    & 95.58 		& 0.13 	& 1858 	      & 235.65 		& 99.40$\pm$0.55\%\\
    \cdashline{1-6}
    \multicolumn{1}{l|}{{\elimipl}}  	& 95.58			& 0.13	  & 1824	& 207.47		& 99.20$\pm$0.65\%\\
    \multicolumn{1}{l|}{{\samcc} (ours)}  	    & 95.58 		& 0.13    & 1863  	& 243.40 		& 99.80$\pm$0.43\%\\
    \multicolumn{1}{l|}{{\samcn} (ours)}  	    & 95.58 		& 0.13    & 1863 	& 246.16 & 99.73$\pm$0.44\%\\
    \cdashline{1-6}
    \multicolumn{1}{l|}{{\miplma}}  	& 95.58		& 0.13	   & 1848	& 218.86 		& 98.47$\pm$1.03\%\\
    \multicolumn{1}{l|}{{\mamcc} (ours)}  	    & 95.58		& 0.13  	& 1870  	& 246.16 		& 99.93$\pm$0.20\%\\
    \multicolumn{1}{l|}{{\mamcn} (ours)}  	    & 95.58		& 0.13  	& 1870 	& 251.02 		& 99.93$\pm$0.20\%\\
    \cdashline{1-6}
    \hline  \hline
    \end{tabular}
\end{table*}

\subsection{Scalability Experiments}
\label{app:Scalability_Experiments}
We conduct additional scalability experiments by progressively increasing the training set size from $10\%$ to $100\%$ of the full training samples. As shown in \cref{fig:times}, the runtime of all methods increases in an approximately linear manner with respect to the proportion of training samples. Moreover, the curves of our methods remain nearly parallel to those of their corresponding baselines, suggesting that our methods preserve the same asymptotic scaling trend with respect to data size as the comparative methods.
In addition to the scalability curves, we summarize the computational and resource costs in \cref{tab:complexity}, including floating point operations (FLOPs), number of parameters (Params), maximum GPU memory usage (MM), total runtime (Runtime) for training and testing, and the average accuracy (Acc) over $10$ runs. Notably, all methods have identical FLOPs (95.58M) and parameter counts (0.13M), indicating that our methods do not increase the model size or theoretical computational complexity. In practice, our methods incur only modest overhead relative to the corresponding baselines, the maximum GPU memory increases by approximately $1.2\%$--$2.1\%$, and the total runtime increases by about $12.5\%$--$19.5\%$ when using the full training set. Overall, these results demonstrate that our method scales favorably with increasing training data and exhibits good scalability in terms of both runtime growth and resource consumption.

\begin{table}[t]
    \centering
    \caption{Summary of the Friedman statistics $F_F\,(l = 11, N = 19)$ and the critical value
    in terms of each evaluation metric ($l$: \# comparing methods; $N$: \# datasets).}
    \label{tab:friedman}
    \begin{tabular}{lcc}
    \hline\hline
    Evaluation metric & $F_F$ & critical value ($\alpha = 0.05$) \\
    \hline
    \emph{Accuracy}        &  116.6938   & \multirow{2}{*}{1.8836} \\
    \emph{Expected calibration error}           & 50.3444   &                           \\
    \hline\hline
    \end{tabular}
\end{table}

\subsection{Statistical Comparison}
\label{app:Statistical_Comparison}
To systematically compare the performance of our methods and the compared methods, we employ the Friedman test~\cite{Demsar06}, a widely used nonparametric procedure for comparing multiple methods across multiple datasets. Given $l$ methods and $N$ datasets, let $r_{i,j}$ denote the rank of the $j$-th method on the $i$-th dataset (mean ranks are assigned in case of ties). The average rank of method $j$ is
$R_j = \frac{1}{N} \sum_{i=1}^N r_{i,j}$. Under the null hypothesis that all methods have equivalent performance, the Friedman statistic $F_F$ is approximately $F$-distributed with $(l-1)$ and $(l-1)(N-1)$ degrees of freedom in the numerator and denominator, respectively:
\begin{equation}
  F_F = \frac{(N-1)\chi_F^{2}}{N(l-1)-\chi_F^{2}},
  \label{eq:ff}
\end{equation}
where
\begin{equation}
  \chi_F^{2}
  = \frac{12N}{l(l+1)}
    \left[
      \sum_{j=1}^{l} R_j^{2}
      - \frac{l(l+1)^{2}}{4}
    \right].
    \label{eq:chif2}
\end{equation}

\begin{figure}[!t]
    \setlength{\abovecaptionskip}{-0.1cm}
    \centering
    \begin{subfigure}{0.99\linewidth}
        \centering
        \begin{overpic}[width=8.8cm]{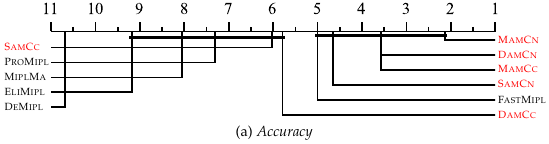}
        \end{overpic}
    \end{subfigure}
    \vspace{0.4cm} 
    \begin{subfigure}{0.99\linewidth}
        \centering
        \begin{overpic}[width=8.8cm]{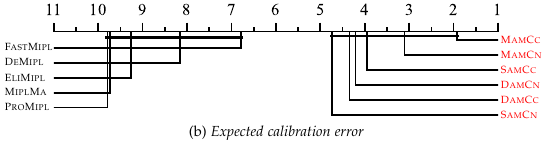}
        \end{overpic}
    \end{subfigure}
    \caption{Critical difference (CD) diagrams comparing the proposed methods (in red) with the baseline methods using the Nemenyi test. Methods that are not connected in the CD diagram are regarded as having significantly different performance at significance level $\alpha = 0.05$.}
    \label{fig:cd}
\end{figure}

\begin{figure*}[!t]
    \centering
	\begin{overpic}[width=18.0cm]{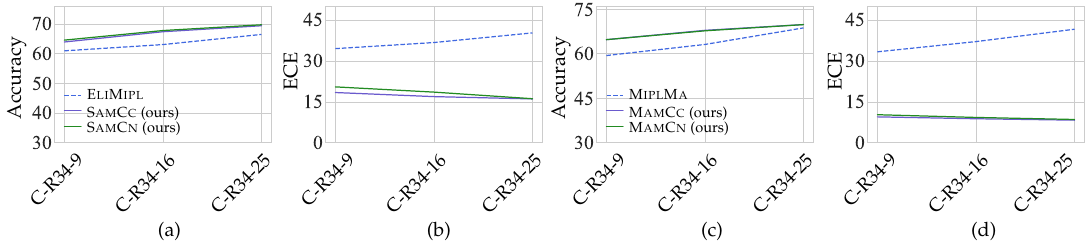}
    \end{overpic}
    \caption{Effect of patch granularity. (a, b) Accuracy/ECE of {\elimipl} and our {\samcc}/{\samcn}, (c, d) Accuracy/ECE of {\miplma} and our {\mamcc}/{\mamcn}.}
\label{fig:crc_patch_granularity_engineering}
\end{figure*}

\Cref{tab:friedman} reports the Friedman statistics $F_F$ and the corresponding critical values for both accuracy and expected calibration error. We exclude {\miplgp} from the statistical significance analysis because computational constraints prevent its evaluation across multiple datasets. At significance level $\alpha = 0.05$, the null hypothesis of equal performance among the competing methods is rejected for all metrics. This implies that at least one method performs differently from the others, and a post-hoc analysis is required to characterize the relative performance of individual methods. \looseness=-1

To this end, we apply the Nemenyi test~\cite{nemenyi1963}, which compares all pairs of methods based on their average ranks (smaller ranks correspond to better performance). In this test, the absolute difference between the average ranks of two methods is compared with the critical difference (CD):
\begin{equation}
  \mathrm{CD}
  = q_{\alpha} \sqrt{\frac{l(l+1)}{6N}},
  \label{eq:cd}
\end{equation}
where $q_{\alpha}$ is the critical value of the Studentized range statistic for significance level $\alpha$.

In our experimental setting, we have $q_{\alpha} = 3.219$ at significance level $\alpha = 0.05$, yielding $\mathrm{CD} = 3.4638$ (with $l = 11$ and $N = 19$). Consequently, the performance of two methods is deemed significantly different if their average ranks over all datasets differ by at least one CD. To visualize the relative performance of the proposed and baseline methods, \Cref{fig:cd} shows the CD diagrams for both accuracy and expected calibration error. Each method is positioned on the axis according to its average rank. Any pair of methods whose average ranks differ by less than one CD is connected by a thick horizontal line, indicating that their performances are not significantly different at level $\alpha = 0.05$. Conversely, methods that are not connected are regarded as having significantly different performance.

From the CD diagrams and the accompanying statistics, we make the following observations: 1) The proposed methods obtain the best accuracy in $ 78.95\%$ of the cases. Most proposed variants appear on the right side of the CD diagram and form a single connected group, which indicates top average ranks with no significant differences within the group. Most baselines lie to the left and are not connected to this group, implying significantly worse accuracy. An exception is {\fastmipl}, whose rank is comparable to the top group. In contrast, {\samcc} and {\damcc} do not significantly outperform several baselines. These patterns show that accuracy gains are not uniform across all variants.
2) The proposed methods achieve the lowest ECE in $94.74\%$ of the cases. We can observe a clearer separation in the CD diagram. All proposed methods are grouped on the far right and are disconnected from the baseline cluster on the left. This pattern indicates statistically significant improvements in calibration for every proposed method relative to all baselines, whereas differences within the proposed methods are not statistically significant.

Overall, the CD diagrams show that our methods deliver consistently better calibration and competitive, often superior, accuracy compared with the baselines on the benchmark and real-world datasets.

\section{Engineering Case Study}
\label{app:engineering_case_study}

\subsection{Engineering Impact of Patch Granularity on the CRC-{\small MIPL}}
\label{app:crc_patch_granularity_engineering}

In practical pathology pipelines, a common engineering decision is the patch granularity used to represent a slide as a multi-instance bag: using more patches can capture finer tissue heterogeneity, but also increases the bag size and may change the confidence behavior of the model.
To examine the performance on the CRC-{\scriptsize{MIPL}} dataset, we adopt the ResNet-34 to learn patch-based features and vary the number of non-overlapping patches per image as $N\in\{9,16,25\}$, yielding C-R34-9, C-R34-16, and C-R34-25.

Fig.~\ref{fig:crc_patch_granularity_engineering} summarizes how both classification and calibration performance evolve as the patch granularity increases.
From an engineering perspective, increasing $N$ consistently improves classification accuracy across methods, indicating that finer patch representations provide more informative bag evidence for colorectal cancer (CRC) classification.
However, the same increase in granularity can make disambiguation-only baselines less reliable in terms of calibration. As shown in Fig.~\ref{fig:crc_patch_granularity_engineering}~(b,d), while their accuracies improve, their ECE grows notably as $N$ increases, suggesting that their confidence scores become less suitable for downstream confidence-based decisions.
In contrast, our CDL-based variants maintain low and stable ECE across all $N$ while achieving higher accuracy. This behavior is desirable in engineering deployments, where calibrated confidence is needed for confidence-aware operation modes such as triaging low-confidence cases to human review, setting safe decision thresholds, or prioritizing uncertain samples for further inspection.
Overall, this study shows that CDL enables practitioners to benefit from finer patch granularity without sacrificing confidence reliability, strengthening its applicability to real CRC pathology pipelines. 

We recommend selecting $N$ based on a three-way trade-off among (i) accuracy gain, (ii) calibration reliability (ECE), and (iii) computational budget.
A practical rule is to increase $N$ until accuracy improvements begin to saturate while ECE remains acceptable for the intended operating mode. In our CRC pipeline, larger $N$ provides consistent accuracy gains; with CDL, calibration remains stable, so $N$ can be chosen primarily by the desired throughput. On a single NVIDIA RTX 3090 GPU, our training time is approximately $39.97$, $41.34$, and $43.31$ minutes for $N=9,16,25$, respectively, and CDL incurs only about $\sim$10\% runtime overhead over the corresponding disambiguation-only baselines. Therefore, when resources allow, $N=25$ is a strong default to maximize accuracy; $N=16$ offers a favorable trade-off between performance and cost; and $N=9$ can be used when throughput is the dominant constraint.

\begin{figure}[!t]
    \centering
	\begin{overpic}[width=9.0cm]{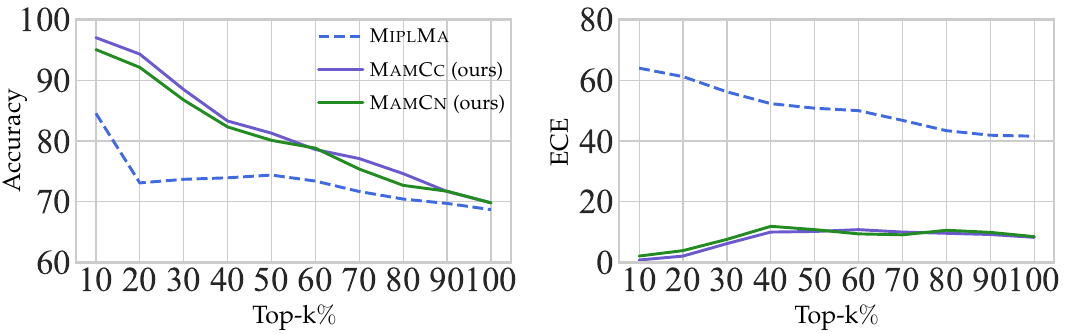}
    \end{overpic}
    \caption{Classification accuracy and expected calibration error of {\miplma}, {\mamcc}, and {\mamcn} on the C-R34-25 dataset with varying top-k$\%$ predictions.}
 \label{fig:crc_topk_curve}
\end{figure}

\subsection{Confidence-Based Triage on the C-R34-25 via Top-$k\%$ Predictions}
\label{app:crc_topk_confidence}

In engineering deployments of pathology-style MIPL systems, a common operating mode is confidence-aware decision making: the model automatically reports only the most confident predictions, while low-confidence cases are deferred for further review. To assess whether a model's confidence is practically actionable, we analyze classification and calibration on the top-$k\%$ most confident predictions.

For each test bag in the C-R34-25 dataset, we compute the predictive confidence
$s=\max_{c}\hat{p}_c$ (maximum predicted class probability), and rank all test bags by $s$ in descending order.
For $k\in\{10,20,\ldots,100\}$, we retain the top-$k\%$ predictions and evaluate:
(i) \emph{Accuracy@Top-$k\%$} and (ii) \emph{ECE@Top-$k\%$} on the retained subset.
This evaluation directly reflects how reliable the model is on uncertain cases.

Fig.~\ref{fig:crc_topk_curve} compares {\miplma} with our CDL-enhanced variants {\mamcc} and {\mamcn}. Two observations are most relevant for triage. First, {\mamcc} and {\mamcn} achieve markedly higher Accuracy@Top-$k\%$ in the low-coverage regime (e.g., top-$10\%$ and top-$20\%$) and then decrease smoothly as $k$ increases, suggesting that CDL yields a more informative confidence ranking: the most confident predictions are substantially more likely to be correct. Second, {\mamcc} and {\mamcn} maintain consistently low ECE across $k$, indicating that their probability estimates remain numerically reliable for thresholding and downstream decision policies. In contrast, {\miplma} exhibits both lower Accuracy@Top-$k\%$ and much larger ECE, implying that its confidence scores are less effective for filtering and are poorly aligned with empirical correctness. This trend aligns with the under-confident behavior observed for MIPL baselines in the reliability analysis (as shown in Fig.~2).

In deployment, one can choose an operating point $k$ or a confidence threshold $\tau$ according to the desired safety-throughput trade-off:
(i) use Accuracy@Top-$k\%$ to estimate the expected correctness of auto-reported cases ($\text{risk} \approx 1-\text{Accuracy@Top-}k\%$), and (ii) use ECE@Top-$k\%$ to assess whether the reported confidence scores can be trusted for thresholding and risk communication.
Then, the system auto-reports the top-$k\%$ predictions or those with $s\ge\tau$ and routes the remaining low-confidence cases to human review or further testing. Because {\mamcc} and {\mamcn} achieve high accuracy at low coverage and low ECE across operating points, they enable safer and more reliable confidence-based triage on CRC pathology data than {\miplma}.

\end{document}